\documentclass[lettersize,journal]{IEEEtran}
\usepackage[utf8]{inputenc} 
\usepackage[T1]{fontenc}
\usepackage{url}              
\usepackage{cite}             

\usepackage[cmex10]{amsmath}	  
\usepackage{amsfonts}
\usepackage{mleftright}       
\usepackage{cgitIEEE}		  
\mleftright

\usepackage{graphicx}
\usepackage{hyperref}
\hypersetup{ 
    colorlinks=true, 
    linkcolor=black, 
    citecolor=black, 
    urlcolor=black, 
    pdftitle={Your Title}, 
    pdfauthor={Your Name}, 
    bookmarksnumbered=true,
    bookmarksopen=true,
}

\usepackage{orcidlink}
\usepackage{balance}

\begin{document}

\title{Asymmetry of the Relative Entropy in the Regularization of Empirical Risk Minimization}

\author{Francisco Daunas{\orcidlink{0009-0009-2038-9985}},
I{\~n}aki Esnaola{\orcidlink{0000-0001-5597-1718}},
Samir M. Perlaza{\orcidlink{0000-0002-1887-9215}}, and 
H.~Vincent Poor{\orcidlink{0000-0002-2062-131X}}
\thanks{
This work was presented in part at the International Symposium on Information Theory (ISIT) 2023 in \cite{perlazaISIT2023a}. This work is supported in part by a University of Sheffield ACSE PGR scholarship; the European Commission through the H2020-MSCA-RISE-2019 project 872172; the French National Agency for Research (ANR) through the Project ANR-21-CE25-0013 and the project ANR-22-PEFT-0010 of the France 2030 program PEPR R\'eseaux du Futur; the Agence de l'innovation de d\'efense (AID) through the project UK-FR 2024352; and in part by the U.S National Science Foundation under Grant ECCS-2335876

F. Daunas is with the School of Electrical and Electronic Engineering, University of Sheffield, Sheffield S1 3JD, U.K.; and also with INRIA, Centre Inria d'Universit\'e C\^ote d'Azur, 06902 Sophia Antipolis, France (e-mail: jdaunastorres1@sheffield.ac.uk).

I. Esnaola is with the School of Electrical and Electronic Engineering, University of Sheffield, Sheffield S1 3JD, U.K.; and also with the Department of Electrical and Computer Engineering, Princeton University, Princeton, NJ 08544 USA (e-mail: esnaola@sheffield.ac.uk).

Samir M. Perlaza is with INRIA, Centre Inria d'Universit\'e C\^ote d'Azur, 06902 Sophia Antipolis, France; also with the Department of Electrical and Computer Engineering, Princeton University, Princeton, NJ 08544 USA; and also with the GAATI Mathematics Laboratory, University of French Polynesia, 98702 Faaa, French Polynesia (e-mail: samir.perlaza@inria.fr).

H. Vincent Poor is with the Department of Electrical and Computer Engineering, Princeton University, Princeton, NJ 08544 USA (e-mail: poor@princeton.edu).

}
}


\maketitle

\begin{abstract}
The effect of relative entropy asymmetry is analyzed in the context of empirical risk minimization (ERM) with relative entropy regularization (ERM-RER).
Two regularizations are considered: $(a)$ the relative entropy of the measure to be optimized with respect to a reference measure (\mbox{Type-I} ERM-RER); and $(b)$ the relative entropy of the reference measure with respect to the measure to be optimized (\mbox{Type-II} ERM-RER).
The main result is the characterization of the solution to the \mbox{Type-II} ERM-RER problem and its key properties. By comparing the well-understood \mbox{Type-I} ERM-RER with \mbox{Type-II} ERM-RER, the effects of entropy asymmetry are highlighted.
The analysis shows that in both cases, regularization by relative entropy forces the support of the solution to collapse into the support of the reference measure, introducing a strong inductive bias that negates the evidence provided by the training data.
Finally, it is shown that \mbox{Type-II} regularization is equivalent to \mbox{Type-I} regularization with an appropriate transformation of the empirical risk function.
\end{abstract}


\section{Introduction}
\IEEEPARstart{E}{mpirical} risk minimization (ERM) is a central tool in supervised machine learning. Among other uses, it enables the characterization of sample complexity and probably approximately correct (PAC) learning in a wide range of settings~\cite{vapnik1992principles}.
The application of ERM in the study of theoretical guarantees spans related disciplines such as machine learning \cite{vapnik1964perceptron}, information theory \cite{rodrigues2021information,mezard2009information} and statistics \cite{wainwright2019high, vershynin2018high}.
Classical problems such as classification\cite{blumer1989learnability,guyon1991structural}, pattern recognition \cite{lugosi1995nonparametric,bartlett1998sample}, regression \cite{vapnik1993local, cherkassky1999model}, and density estimation \cite{lugosi1995nonparametric,vapnik1999overview} can be posed as special cases of the ERM problem \cite{vapnik1999overview,bottou2018optimization}.
Unfortunately, ERM is prone to training data memorization, a phenomenon also known as overfitting \cite{krzyzak1996nonparametric,deng2009regularized,arpit2017Memorization}. For that reason, ERM is often regularized in order to provide generalization guarantees \cite{hellstrom2023generalization, bousquet2002stability,vapnik2015uniform,aminian2021exact}.
Regularization establishes a preference over the models by encoding features of interest that conform to prior knowledge.
In different statistical learning frameworks, such as Bayesian learning \cite{robert2007bayesian,mcallester1998pacBayesian} and PAC learning \cite{valiant1984theory,shawe1997pac,cullina2018pac}, the prior knowledge over the set of models can be described by a reference probability measure.
More general references can be adopted as proved in \cite{perlazaISIT2022, perlaza2024ERMRER} for the case of $\sigma$-finite measures.
In either case, the solution to the regularized ERM problem can be cast as a probability distribution over the set of models.
Prior knowledge of the set of datasets can also be represented by probability measures, e.g., the worst-case data-generating probability measure introduced in \cite{zou2024WorstCase}.

\subsection{Motivation}
A common regularizer of the ERM problem is the relative entropy of the optimization probability measure with respect to a given reference measure over the set of models \cite{vapnik1999overview,raginsky2016information,russo2019much,zou2009generalization}.
The resulting problem formulation, termed ERM with relative entropy regularization (ERM-RER) has been extensively studied for both the case in which the reference measure is a probability measure \cite{raginsky2016information,russo2019much,zou2009generalization, aminian2023information} and the case in which it is a $\sigma$-finite measure \cite{perlazaISIT2022, perlaza2024ERMRER, perlazaISIT2023b}. While in both cases the solution is unique and corresponds to a Gibbs probability measure, the existence of the solution is guaranteed only in the case in which the reference measure is a probability measure~\cite{perlaza2024ERMRER}. 
Despite the many merits of the ERM-RER formulation, it has some significant limitations.
Firstly, the absolute continuity of the optimization measure with respect to the reference measure is required for the existence of the corresponding Radon-Nikodym derivative, which is used by the relative entropy regularization. This absolute continuity sets an insurmountable barrier to the exploration of models outside the support of the reference measure. More specifically, models outside the support of the reference measure exhibit zero probability with respect to the Gibbs probability measure solution to ERM-RER, regardless of the evidence provided by the training dataset. Furthermore, selecting priors with full support often leads to computationally expensive partition functions \cite{russo2019much,raginsky2016information,perlaza2024ERMRER} in high-dimensional spaces. While priors with large supports ensure the inclusion of high-performing models, they also assign non-zero probability to models with poor empirical risk.
Secondly, the choice of relative entropy over alternative divergences often follows arguments based on the simplicity of obtaining generalization guarantees in the form of bounds \cite{hellstrom2023generalization} or even closed form expressions, see \cite{perlaza2024ERMRER} and \cite{perlaza2024HAL}. Nonetheless, such bounds and closed form expressions are often hard to calculate and are not always informative when evaluated in practical settings \cite{perlazaISIT2023a, wang2004enhancing,lin2015accelerated,yang2022estimation, zou2024WorstCase, perlaza2024ERMRER, Wu2024OnGeneralization, borjas2024thesis}.
The problem of ERM with a general $f$-divergence regularization has been explored in \cite{teboulle1992entropic} and \cite{beck2003mirror} in the case of a finite countable set of models, and recently extended to uncountable sets of models in \cite{alquier2021non} and \cite{esposito2022GenfDiv}. 
\IEEEpubidadjcol
The authors in \cite{teboulle1992entropic, beck2003mirror, alquier2021non, esposito2022GenfDiv} constrain the optimization domains to sets of measures that are mutually absolutely continuous with respect to the reference probability measure. 
In view of these, exploring the asymmetry of relative entropy is of particular interest to advancing the understanding of entropy regularization in the context of ERM and its role in generalization. Additionally, examining the asymmetry opens novel pathways to overcome some of the constraints imposed by relative entropy regularization, such as, the ability to select models outside the support of the prior.

The use of the relative entropy of the optimization measure with respect to the reference measure as a regularizer in the ERM-RER is termed \mbox{Type-I} ERM-RER. Alternatively, the use of the relative entropy of the reference measure with respect to the optimization measure is termed \mbox{Type-II} ERM-RER.
Interestingly, the results in \cite{teboulle1992entropic, beck2003mirror, alquier2021non}, which lead to special cases of the \mbox{Type-I} and \mbox{Type-II} ERM-RER problems by assuming that $f(x) = -x \log(x)$ and $f(x) = -\log(x)$,  respectively, do not study the impact of the asymmetry of relative entropy.
Another observation that motivates studying the asymmetry of relative entropy in ERM-RER is that numerical analyses of the \mbox{Type-II} ERM-RER, presented in Section~\ref{sec:logERM_RER}, suggest that \mbox{Type-II} regularization exhibits a markedly different relationship between test error and training error when compared to that of \mbox{Type-I} regularization. While, the generalization capabilities of \mbox{Type-I} are better in the simulations carried out for this work, the performance of the \mbox{Type-II} regularization is comparable and displays promising properties that warrant further research.

\subsection{Contributions}
This paper presents the solution to \mbox{Type-II} ERM-RER optimization problem using a new method of proof. In particular,  mutual absolute continuity between the measures involved is not imposed. Surprisingly, mutual absolute continuity is exhibited by the solution as a consequence of the structure of the problem. The key properties of the solution are highlighted, and an equivalence between the \mbox{Type-I} and \mbox{Type-II} ERM-RER problems is presented. This equivalence is achieved by replacing the empirical risk in the \mbox{Type-I} ERM-RER problem with another function, which can be interpreted as a tunable loss function, as described in \cite{liao2018tunable, sypherd2019tunable, kurri2021realizing}.
The remainder of the paper is organized as follows. Section~\ref{sec:ERMproblem} presents the ERM-RER problem and its  two variations: \mbox{Type-I} and \mbox{Type-II}. 
%
%
The main contribution of this paper, which is the solution to the \mbox{Type-II} ERM-RER problem, is presented in Section~\ref{sec:SolutionType2}. This section also presents key properties of the solution. Section~\ref{sec:ExpectedER} uses these properties to characterize the expected empirical risk.
Section~\ref{sec:logERM_RER} studies the equivalence between \mbox{Type-I}  and \mbox{Type-II} ERM-RER problems.
This work is concluded by Section~\ref{sec:FinalRemarks}, with some final remarks.


%
%
\section{Empirical Risk Minimization}
\label{sec:ERMproblem}
Let~$\set{M}$,~$\set{X}$ and~$\set{Y}$, with~$\set{M} \subseteq \reals^{d}$ and~$d \in \ints$, be sets of \emph{models}, \emph{patterns}, and \emph{labels}, respectively.  
A pair~$(x,y) \in \mathcal{X} \times \mathcal{Y}$ is referred to as a \emph{labeled pattern} or as a \emph{data point}.
Given~$n$ data points, with~$n \in \ints$,  denoted by~$\left(x_1, y_1 \right)$, $\left( x_2, y_2\right)$, $\ldots$, $\left( x_n, y_n \right)$, the corresponding dataset is represented by the tuple
\begin{equation}\label{EqTheDataSet}
\vect{z} \triangleq \big(\left(x_1, y_1 \right), \left(x_2, y_2 \right), \ldots, \left(x_n, y_n \right)\big)  \in \left( \set{X} \times \set{Y} \right)^n.
\end{equation}  

Let the function~$f: \set{M} \times \mathcal{X} \rightarrow \mathcal{Y}$ be such that the label assigned to the pattern $x$ according to the model $\vect{\theta} \in \set{M}$ is $f(\vect{\theta}, x)$.
Let also the function 
\begin{equation}\label{EqEll}
\ell: \set{Y} \times \set{Y} \rightarrow [0, \infty)
\end{equation} 
be such that given a data point~$(x, y) \in \set{X} \times \set{Y}$, the  risk induced by a model~$\vect{\theta} \in \set{M}$ is~$\ell\left( f(\vect{\theta}, x), y \right)$.  
In the following, the risk function~$\ell$ is assumed to be nonnegative and  for all~$y \in \set{Y}$, $\ell\left( y , y\right) = 0$.

The \emph{empirical risk} induced by the model~$\vect{\theta}$, with respect to the dataset $\vect{z}$ in~\eqref{EqTheDataSet} is determined by the  function~$\mathsf{L}_{\vect{z}}: \set{M} \rightarrow [0, \infty )$, which satisfies  
\begin{IEEEeqnarray}{rCl}
\label{EqLxy}
\mathsf{L}_{\vect{z}} \left(\vect{\theta} \right)  & = & 
\frac{1}{n}\sum_{i=1}^{n}  \ell\left( f(\vect{\theta}, x_i), y_i\right).
\end{IEEEeqnarray}
Using this notation, the ERM consists of the following optimization problem:
\begin{equation}\label{EqOriginalOP}
\min_{\vect{\theta} \in \set{M}} \mathsf{L}_{\vect{z}} \left(\vect{\theta} \right).
\end{equation}
Let the set of solutions to the ERM problem in~\eqref{EqOriginalOP} be denoted by
\begin{equation}\label{EqHatTheta}
\set{T}\left( \vect{z} \right) \triangleq \arg\min_{\vect{\theta} \in \set{M}}    \mathsf{L}_{\vect{z}} \left(\vect{\theta} \right).
\end{equation}
Note that if the set $\set{M}$ is finite, the ERM problem in~\eqref{EqOriginalOP} always possesses a solution, and thus, $\abs{\set{T}\left( \vect{z} \right)} > 0$. Nonetheless, in general, the ERM problem does not necessarily possess a solution, \ie, it might happen that $\abs{\set{T}\left( \vect{z} \right) } = 0$.

%
The PAC and Bayesian frameworks, as discussed in \cite{mcallester1998pacBayesian} and \cite{shawe1997pac},  address the problem in~\eqref{EqOriginalOP} by constructing probability measures, conditioned on the dataset $\dset{z}$, from which models are randomly sampled.
In this context, the focus is on probability measures that assign high probability to minimizers of the ERM problem in~\eqref{EqOriginalOP}. Such probability measures are defined on the measurable space~$\msblspc{\set{M}}$, which is denoted by $\bigtriangleup(\set{M})$.
From this perspective, the underlying assumption in the remainder of this work is that the functions $f$ and $\ell$  in~\eqref{EqLxy} are such that for all $(x,y) \in \set{X} \times \set{Y}$,  the function $g_{x,y}:\set{M}\rightarrow[0, \infty)$, such that $g_{x,y}(\thetav)=\ell(f(\thetav,x),y)$, is measurable with respect to the Borel measurable spaces $(\set{M}, \field{F})$ and $\left( \reals, \BorSigma{\reals} \right)$, where $\field{F}$ and $\BorSigma{\reals}$ are, respectively, Borel $\sigma$-fields on $\set{M}$ and $\reals$. 
Under these assumptions, a common metric is the expected empirical risk.
\begin{definition}[Expected Empirical Risk]
\label{DefEmpiricalRisk}
Given the dataset \mbox{$\dset{z} \in  ( \set{X} \times \set{Y} )^n$} in~\eqref{EqTheDataSet},  let  the functional $\mathsf{R}_{\dset{z}}: \bigtriangleup(\set{M}) \rightarrow  [0, \infty  )$~be such that
\begin{equation}
\label{EqRxy}
\foo{R}_{\dset{z}}( P ) = \int \foo{L}_{ \dset{z} } ( \thetav )  \diff P(\thetav),
\end{equation}
where the function $\foo{L}_{\dset{z}}$ is defined in~\eqref{EqLxy}.
\end{definition}

In the following section, the \mbox{Type-I} relative entropy regularization is reviewed as it serves as the basis for the analysis of the regularization asymmetry.

%
%
\subsection{The \mbox{Type-I} ERM-RER Problem}
\label{sec:Type1ERM_RER}
The \mbox{Type-I} ERM-RER problem is parameterized by a probability measure $Q \in \bigtriangleup(\set{M})$ and a real $\lambda \in \left( 0, \infty\right)$. The measure $Q$ is referred to as the \emph{reference measure} and $\lambda$ as the  \emph{regularization factor}. 
The {Type-I} ERM-RER problem, with parameters~$Q$ and~$\lambda$, is given by the following optimization problem:
%
\begin{IEEEeqnarray}{rCl}
\label{EqOpType1ERMRERNormal}
    \min_{P \in \bigtriangleup_{Q}(\set{M}) } &\ & \foo{R}_{\dset{z}} ( P )  + \lambda \KL{P}{Q},
\end{IEEEeqnarray}
%
where the functional~$\foo{R}_{\dset{z}}$ is defined in~\eqref{EqRxy}, and the optimization domain~is 
\begin{IEEEeqnarray}{rCl}
\label{DefSetTriangUp}
		\bigtriangleup_{Q}(\set{M})   & \triangleq & \{P\in \bigtriangleup(\set{M}): P\ll Q \},
\end{IEEEeqnarray}
with the notation~$P\ll Q$ standing for~$P$ being absolutely continuous with respect to~$Q$.

%
The solution to the \mbox{Type-I} ERM-RER problem in~\eqref{EqOpType1ERMRERNormal} is the Gibbs probability measure reported in \cite{perlazaISIT2022, raginsky2016information} and \cite{russo2019much}. In order to introduce such a measure, consider the function $K_{Q, \dset{z}}: \reals \rightarrow \reals$ that satisfies for all $t \in \reals$,
\begin{equation}
\label{EqDefKfunction}
		K_{Q, \dset{z}}( t ) = \log (\int \exp(t\foo{L}_{\dset{z}}(\thetav))\diff Q(\thetav)),
\end{equation}
with $\foo{L}_{\dset{z}}$ in~\eqref{EqLxy}.
Using this notation, the solution to the  \mbox{Type-I} ERM-RER problem in~\eqref{EqOpType1ERMRERNormal} is  presented by the following lemma.

%
\begin{lemma}[\text{\cite[Theorem $3$]{perlaza2024ERMRER}}]
\label{lemm_OptimalModelType1}
The solution to the optimization problem in~\eqref{EqOpType1ERMRERNormal} is a unique probability measure, denoted by $\Pgibbs{P}{Q}$, which satisfies for all $\thetav \in \supp Q$,
\begin{equation}
\label{EqGenpdfType1}
\frac{\diff \Pgibbs{P}{Q}}{\diff Q} ( \thetav ) = \exp(-K_{Q, \dset{z}}(-\frac{1}{\lambda})-\frac{1}{\lambda}\foo{L}_{\dset{z}}(\thetav)),
\end{equation}
where the function $\foo{L}_{\dset{z}}$ is defined in~\eqref{EqLxy} and the function $K_{Q, \dset{z}}$ is defined in~\eqref{EqDefKfunction}.
\end{lemma} 

%
%
\subsection{The \mbox{Type-II} ERM-RER Problem}
\label{sec:Type2ERM_RER}

The \mbox{Type-II} ERM-RER problem is parameterized by a probability measure $Q \in \bigtriangleup(\set{M})$ and a real $\lambda \in (0, \infty)$.
As in the \mbox{Type-I} ERM-RER problem, the measure $Q$ is referred to as the \emph{reference measure} and $\lambda$ as the \emph{regularization factor}.
Given the dataset~$\dset{z} \in (\set{X} \times \set{Y})^n$ in~\eqref{EqTheDataSet}, the \mbox{Type-II} ERM-RER problem, with parameters~$Q$ and~$\lambda$, consists of the following optimization problem:
\begin{IEEEeqnarray}{rcl}
\label{EqOpType2ERMRERNormal}
    \min_{P \in \bigtriangledown_{Q}(\set{M}) } &\ & \foo{R}_{\dset{z}} ( P )  + \lambda \KL{Q}{P},
\end{IEEEeqnarray}
where the functional~$\foo{R}_{\dset{z}}$ is defined in~\eqref{EqRxy}, and the optimization domain~is 
\begin{IEEEeqnarray}{rCl}
\label{DefSetTriangDown}
		\bigtriangledown_{Q}(\set{M}) & \triangleq & \{P\in \bigtriangleup(\set{M}): Q\ll P \}.
\end{IEEEeqnarray}
The difference between \mbox{Type-I} and \mbox{Type-II} ERM-RER problems lies on the regularization. While the former uses the relative entropy $\KL{P}{Q}$, the latter uses $\KL{Q}{P}$. 
This translates into different optimization domains due to the asymmetry of the relative entropy.
More specifically, in the \mbox{Type-I} ERM-RER problem, the optimization domain is the set of probability measures on the Borel measurable space $\msblspc{\set{M}}$ that are absolutely continuous with the reference measure $Q$. That is, the set $\bigtriangleup_{Q}(\set{M})$ in~\eqref{DefSetTriangUp}.
Alternatively, in the \mbox{Type-II} ERM-RER problem, the optimization domain consists of probability measures defined on the Borel measurable space $\msblspc{\set{M}}$, with the additional condition that the reference measure $Q$ must be absolutely continuous with respect to them. This corresponds to the set denoted as $\bigtriangledown_{Q}(\set{M})$ in~\eqref{DefSetTriangDown}.
From this perspective, the techniques used in \cite{perlaza2024ERMRER} for solving the \mbox{Type-I} ERM-RER no longer hold. In the next section, a new technique is developed for solving the \mbox{Type-II} ERM-RER.

The problems in~\eqref{EqOpType1ERMRERNormal} and~\eqref{EqOpType2ERMRERNormal} exhibit trivial solutions when the functional $\foo{R}_{\dset{z}}$ is such that for all $P \in \bigtriangleup_{Q}(\set{M})$ or $P \in \bigtriangledown_{Q}(\set{M})$, respectively, it holds that $\foo{R}_{\dset{z}}(P) = c$, for some $c \in [0, \infty)$. In such a case, the solution is unique and equal to the probability measure $Q$, independently of the parameter $\lambda$. In order to avoid this trivial case, the notion of separability of the empirical risk function with respect to the measure $Q$ is borrowed from \cite{perlaza2024ERMRER}. A separable empirical risk function with respect to a given probability measure $P $ is defined as follows.

\begin{definition}[Definition 5 in \cite{perlaza2024ERMRER}]\label{DefSeparableLxy}
The empirical risk function~$\mathsf{L}_{\vect{z}}$ in~\eqref{EqLxy} is said to be separable with respect to the probability measure~$P \in \bigtriangleup(\set{M})$, if there exist a positive real~$c > 0$  and two subsets~$\set{A}$ and~$\set{B}$ of~$\set{M}$  that are nonnegligible with respect to~$P$, and for all~$(\vect{\theta}_1, \vect{\theta}_2) \in \set{A} \times \set{B}$, 
\begin{IEEEeqnarray}{rcl}
\label{EqTwoNonnegligibleSets}
 \mathsf{L}_{\vect{z}} \left( \vect{\theta}_1 \right) &< c <& \mathsf{L}_{\vect{z}}\left(\vect{\theta}_2\right) < \infty .
\end{IEEEeqnarray}
\end{definition}
A nonseparable empirical risk function~$\mathsf{L}_{\vect{z}}$ in~\eqref{EqLxy} with respect to a measure $P$ is a constant almost surely with respect to the measure~$P$. More specifically, there exists a real~$a \geq 0$, such that
\begin{equation}
P\left( \left\lbrace \vect{\theta} \in \set{M}: \mathsf{L}_{\vect{z}}\left(\vect{\theta}\right)  = a \right\rbrace\right) = 1.
\end{equation}
When the empirical risk function~$\mathsf{L}_{\vect{z}}$ in~\eqref{EqLxy} is nonseparable with respect to all measures in $P \in \bigtriangledown_{Q}(\set{M})$, the trivial case described above is observed. The notion of separable empirical risk functions would play a central role in the study of the optimization problem in~\eqref{EqOpType2ERMRERNormal}. 

%
%
\section{The Solution to the \mbox{Type-II} ERM-RER Problem}
\label{sec:SolutionType2}
 
The solution to the \mbox{Type-II} ERM-RER problem in~\eqref{EqOpType2ERMRERNormal} is presented in the following theorem.
\begin{theorem}
\label{Theo_ERMType2RadNikMutualAbs}
If there exists a real $\beta$ such that
\begin{subequations}
\label{EqType2KrescConstrainAll}
\begin{equation}
\label{EqType2KrescConstrain1}
	\beta \in \{t\in \reals: \forall \thetav \in \supp Q, 0 < {t + \foo{L}_{\dset{z}}(\thetav)}\},
\end{equation}
and
\begin{equation}
\label{EqType2KrescConstrain2}
	\int \frac{\lambda}{\beta + \foo{L}_{\dset{z}}(\thetav)} \diff Q(\thetav) = 1,
\end{equation}
\end{subequations}
with the function $\foo{L}_{\dset{z}}$ defined in~\eqref{EqLxy}, and $\lambda$ and $Q$ the parameters of the optimization problem in~\eqref{EqOpType2ERMRERNormal},
then, the solution to such a problem, denoted by $\Pgibbs{\bar{P}}{Q}\!\in\! \bigtriangleup(\set{M})$, is unique and for all $\thetav \in \supp Q$, it satisfies  
\begin{equation}
\label{EqGenpdfType2}
\frac{\diff \Pgibbs{\bar{P}}{Q}}{\diff Q} ( \thetav ) =  \frac{\lambda}{\beta + \foo{L}_{\dset{z}}(\thetav)}.
\end{equation}
\end{theorem}

Before introducing the proof of Theorem~\ref{Theo_ERMType2RadNikMutualAbs}, two important results are presented. The first result provides the solution to the optimization problem in~\eqref{EqOpType2ERMRERNormal}  when the optimization domain is restricted to 
\begin{equation}
	\label{Eq_ProofThT2_setofmeasures}
		\bigcirc_Q(\set{M}) \triangleq \bigtriangledown_{Q}(\set{M}) \cap \bigtriangleup_{Q}(\set{M}),
	\end{equation}
	where the sets~$\bigtriangleup_{Q}(\set{M})$ and~$\bigtriangledown_{Q}(\set{M})$ are defined in~\eqref{DefSetTriangUp} and~\eqref{DefSetTriangDown}, respectively.
This ancillary problem can be formulated as follows:
\begin{IEEEeqnarray}{rcl}
	\label{EqOpType2ERMRERancillary}
	\min_{P \in \bigcirc_{Q}(\set{M}) } &\ & \foo{R}_{\dset{z}} ( P )  + \lambda \KL{Q}{P},
	\end{IEEEeqnarray}
where the functional~$\foo{R}_{\dset{z}}$ is defined in~\eqref{EqRxy}.	
The solution to the problem in~\eqref{EqOpType2ERMRERancillary} is described by the following lemma. 
	\begin{lemma}
	\label{lemm_Type2RNDbigcirc}
The solution to the optimization problem in~\eqref{EqOpType2ERMRERancillary} 
is unique and identical to the probability measure~$\Pgibbs{\bar{P}}{Q}$ in~\eqref{EqGenpdfType2}.
	\end{lemma}
	\begin{IEEEproof}
The proof is presented in Appendix~\ref{app_lemm_Type2RNDbigcirc}.
	\end{IEEEproof}

The second result consists of comparing the optimal values resulting from the optimization problems in~\eqref{EqOpType2ERMRERNormal} and~\eqref{EqOpType2ERMRERancillary}, as shown hereunder.

	\begin{lemma}
	\label{lemm_RadNikDevMutualIneq}
	The optimization problems in~\eqref{EqOpType2ERMRERNormal}   and~\eqref{EqOpType2ERMRERancillary} satisfy
	\begin{equation}
	\label{Eq_ProofT2IneOp2}
		\min_{P \in \bigtriangledown_{Q}} \foo{R}_{\dset{z}}(P) + \lambda \KL{Q}{P}
		\geq \min_{P \in \bigcirc_{Q}} \foo{R}_{\dset{z}}(P) + \lambda \KL{Q}{P}.
	\end{equation}
	\end{lemma}
	\begin{IEEEproof}
		The proof is presented in Appendix~\ref{AppProof_lemm_RadNikDevMutualIneq}.
	\end{IEEEproof}
Lemma~\ref{lemm_RadNikDevMutualIneq} unveils the fact that the objective function in~\eqref{EqOpType2ERMRERNormal} when evaluated at measures whose support extends beyond the support of $Q$ is larger than such an objective function evaluated at measures whose support is identical to the reference measure.
This includes the case in which the set $\set{T}(\dset{z})$ in~\eqref{EqHatTheta} lies outside the support of $Q$. 
Using these results, the proof of  Theorem~\ref{Theo_ERMType2RadNikMutualAbs} is as follows.

\begin{IEEEproof}[Proof of  Theorem~\ref{Theo_ERMType2RadNikMutualAbs}]
The proof follows by observing that from~\eqref{Eq_ProofThT2_setofmeasures},  it holds that
	\begin{equation}
	\label{Eq_ProofT2supset}
		\bigcirc_{Q}(\set{M}) \subseteq \ \bigtriangledown_{Q}(\set{M}).
	\end{equation}
	Hence, from~\eqref{Eq_ProofT2supset}, it follows that
	\begin{equation}
	\min_{P \in \bigtriangledown_{Q}}
	\foo{R}_{\dset{z}}(P) + \lambda \KL{Q}{P} \leq \min_{P \in \bigcirc_{Q}} \foo{R}_{\dset{z}}(P) + \lambda \KL{Q}{P}.
	\label{Eq_ProofT2IneOp}
	\end{equation}
From the inequalities in~\eqref{Eq_ProofT2IneOp2} and~\eqref{Eq_ProofT2IneOp}, it also follows that
	\begin{equation}
	\min_{P \in \bigtriangledown_{Q}}
	\foo{R}_{\dset{z}}(P) + \lambda \KL{Q}{P} = \min_{P \in \bigcirc_{Q}} \foo{R}_{\dset{z}}(P) + \lambda \KL{Q}{P}.
	\label{Eq_ProofT2IneOpSophiaMakesItBig}
	\end{equation}
	Thus, the measure $\Pgibbs{\bar{P}}{Q}$ in~\eqref{EqGenpdfType2} is the solution of the optimization problem in~\eqref{EqOpType2ERMRERNormal}, which completes the proof of Theorem~\ref{Theo_ERMType2RadNikMutualAbs}. 
\end{IEEEproof}

Lemma~\ref{lemm_RadNikDevMutualIneq} implies that the solution to the optimization problem in~\eqref{EqOpType2ERMRERNormal} is in the set~$\bigcirc_{Q}(\set{M})$ in~\eqref{EqGenpdfType2}.
A consequence of this observation is the following corollary.

\begin{corollary}
\label{coro_mutuallyAbsCont}
The probability measures~$Q$ and~$\Pgibbs{\bar{P}}{Q}$ in~\eqref{EqGenpdfType2} are mutually absolutely continuous.
\end{corollary}
%
Corollary~\ref{coro_mutuallyAbsCont} also follows from Theorem~\ref{Theo_ERMType2RadNikMutualAbs} by observing that the solution to the \mbox{Type-II} ERM-RER  problem in~\eqref{EqOpType2ERMRERNormal} is expressed in terms of its \RadonNikodym derivative with respect to $Q$, which implies the absolute continuity of $\Pgibbs{\bar{P}}{Q}$ with respect to $Q$. The absolute continuity of the measure $Q$ with respect to $\Pgibbs{\bar{P}}{Q}$ follows from the optimization domain of the \mbox{Type-II} ERM-RER  problem.
From this perspective, Corollary~\ref{coro_mutuallyAbsCont} conveys the fact that there does not exist a dataset that can overcome the inductive bias induced by the reference measure $Q$. That is, sets of models outside the support of $Q$ exhibit zero probability measure with respect to the measure $\Pgibbs{\bar{P}}{Q}$.

This observation is important as, at first glance, the \mbox{Type-II} relative entropy regularization for the ERM problem in~\eqref{EqOpType2ERMRERNormal} does not restrict the solution to be absolutely continuous with respect to the reference measure $Q$.
However, Theorem~\ref{Theo_ERMType2RadNikMutualAbs} shows that the support of the probability measure $\Pgibbs{\bar{P}}{Q}$ in~\eqref{EqGenpdfType2} collapses into the support of the reference.
A parallel can be established between \mbox{Type-I} and \mbox{Type-II} cases, as in both cases, the support of the solution is the support of the reference measure.
In a nutshell, the use of relative entropy regularization inadvertently forces the solution to coincide with the support of the reference regardless of the training data.

%
%
\section{The Normalization Function}

Let the set $\set{A}_{Q, \dset{z}} \subseteq \left( 0, \infty)$ and $\set{C}_{Q, \dset{z}} \subset \reals$, with $Q$ and $\dset{z}$ in~\eqref{EqOpType2ERMRERNormal}, be such that if $\lambda \in \set{A}_{Q, \dset{z}}$, then there exists a $\beta \in \set{C}_{Q, \dset{z}}$ that satisfies the inclusion in~\eqref{EqType2KrescConstrain1} and~\eqref{EqType2KrescConstrain2}.  
From Theorem~\ref{Theo_ERMType2RadNikMutualAbs}, specifically from the uniqueness of the solution to~\eqref{EqOpType2ERMRERNormal},  it follows that for all $(\lambda,\beta) \in \set{A}_{Q, \dset{z}} \times \set{C}_{Q, \dset{z}}$ and for all $\alpha \in \reals$, with $\alpha \neq \beta$, it holds that  $(\lambda,\alpha) \notin \set{A}_{Q, \dset{z}} \times \set{C}_{Q, \dset{z}}$. 
This observation allows establishing a bijection between these two sets. Let such a bijection be represented by the function
\begin{subequations}
\label{EqDefNormFunction}
\begin{equation}
\label{EqDefMapNormFunction}
	\bar{K}_{Q, \dset{z}}: \set{A}_{Q, \dset{z}} \rightarrow \set{C}_{Q, \dset{z}},
\end{equation}
which satisfies
\begin{equation}
\label{EqType2Krescaling}
	\bar{K}_{Q, \dset{z}}(\lambda) = \beta,
\end{equation}
\end{subequations}
with $\lambda$ and $\beta$ satisfying \eqref{EqType2KrescConstrainAll}.
The function $\bar{K}_{Q, \dset{z}}$ in~\eqref{EqDefNormFunction} is referred to as the \emph{normalization function}.  This is due to the observation that  the \RadonNikodym derivative $\frac{\diff \Pgibbs{\bar{P}}{Q}}{\diff Q}$  in~\eqref{EqGenpdfType2} can be re-written for all~\mbox{$\thetav \in \supp Q$}, as 
\begin{equation}
\label{EqGenpdfType2WithK}
\frac{\diff \Pgibbs{\bar{P}}{Q}}{\diff Q} ( \thetav ) =  \frac{\lambda}{\bar{K}_{Q, \dset{z}}(\lambda) + \foo{L}_{\dset{z}}(\thetav)}, 
\end{equation}
which together with~\eqref{EqType2KrescConstrain2}, implies that the function $\bar{K}_{Q, \dset{z}}$ ensures that $\Pgibbs{\bar{P}}{Q}$ in~\eqref{EqGenpdfType2} is a probability measure. 

The analysis of the normalization function $\bar{K}_{Q, \dset{z}}$ in~\eqref{EqDefNormFunction} relies on the analysis of its functional inverse, denoted by $\bar{K}^{-1}_{Q, \dset{z}}:\set{C}_{Q, \dset{z}}\rightarrow \set{A}_{Q, \dset{z}}$, which can be defined by noticing that 
\begin{subequations}
\label{Eq_ProofLambdaIsTheInvOfKbar}
\begin{IEEEeqnarray}{rCl}
	1 
	& = & \int \frac{\diff \Pgibbs[\dset{z}]{\bar{P}}{Q}}{\diff Q}(\thetav) \diff Q(\thetav)
	\label{Eq_ProofLambdaIsTheInvOfKbar_s1}\\
	& = & \int \frac{\lambda}{\foo{L}_{\dset{z}}(\thetav) + \beta}\diff Q(\thetav),
	\label{Eq_ProofLambdaIsTheInvOfKbar_s2}
\end{IEEEeqnarray}
\end{subequations}
with the function $\foo{L}_{\dset{z}}$ defined in~\eqref{EqLxy}, and $\lambda$ and $Q$ the parameters of the optimization problem in~\eqref{EqOpType2ERMRERNormal}.
More specifically, from~\eqref{EqType2Krescaling}, it follows that $\lambda = \bar{K}^{-1}_{Q, \dset{z}}\left( \beta \right)$; and from~\eqref{Eq_ProofLambdaIsTheInvOfKbar_s2}, it follows that
\begin{IEEEeqnarray}{rCl}
\label{Eq_InvKbarEq}
	\bar{K}^{-1}_{Q, \dset{z}}(\beta) 
	& = & \frac{1}{\int \frac{1}{\foo{L}_{\dset{z}}(\thetav) + \beta}\diff Q(\thetav)}.
\end{IEEEeqnarray}
Note that while the function $\bar{K}_{Q, \dset{z}}$ in~\eqref{EqDefNormFunction} is defined implicitly, its functional inverse $\bar{K}^{-1}_{Q, \dset{z}}$ is defined explicitly  in~\eqref{Eq_InvKbarEq}.
The existence of the function and inverese follows from the fact that $\bar{K}_{Q, \dset{z}}$ is a bijection.

The purpose of the remaining of this section is to provide a characterization of the sets $\set{A}_{Q, \dset{z}}$ and $\set{C}_{Q, \dset{z}}$.
To do so, some mathematical objects are introduced.
Given a real~$\delta\in [0, \infty)$, consider the Rashomon set~\cite{hsu2022rashomon}, $\set{L}_{\dset{z}}(\delta)$, defined as follows
\begin{equation}
\label{EqType2LsetLamb2zero}
	\set{L}_{\dset{z}}(\delta) \triangleq \{\thetav \in \set{M}: \foo{L}_{\dset{z}}(\thetav) \leq \delta \}.
\end{equation}
Consider also the real numbers $\delta^\star_{Q, \dset{z}}$ and $\lambda^\star_{Q, \dset{z}}$ defined as follow:
\begin{IEEEeqnarray}{rCl}
\label{EqDefDeltaStar}
\delta^\star_{Q, \dset{z}} &\triangleq & \inf \{\delta \in [0, \infty): Q(\set{L}_{\dset{z}}(\delta))>0\},
\end{IEEEeqnarray}
and
\begin{IEEEeqnarray}{rCl}
	 \label{EqDefLambdaStar}
		\lambda^{\star}_{Q, \dset{z}} &\triangleq &\inf\set{A}_{Q,\dset{z}}.
\end{IEEEeqnarray}

Let also $\set{L}^{\star}_{Q, \dset{z}}$ be the level set of the empirical risk function $\foo{L}_{\dset{z}}$ in~\eqref{EqLxy} for the value $\delta^\star_{Q, \dset{z}}$. That is,
\begin{equation}
\label{EqDefSetLStarQz}
	\set{L}^{\star}_{Q, \dset{z}} \triangleq \{\thetav \in \set{M}: \foo{L}_{\dset{z}}(\thetav) = \delta^\star_{Q, \dset{z}}\}.
\end{equation}
Using the objects defined above, the following lemma introduces one of the main properties of the function $\bar{K}_{Q, \dset{z}}$ in~\eqref{EqDefNormFunction}.
\begin{lemma}
\label{lemm_InfDevKtype2}
The function~$\bar{K}_{Q, \dset{z}}$ in~\eqref{EqDefNormFunction} is strictly increasing  and continuous.
\end{lemma}
\begin{IEEEproof}
		The proof is presented in Appendix~\ref{AppProofLemmaInfDevKtype2}.
\end{IEEEproof}
%
The following lemma characterizes the sets $\set{A}_{Q, \dset{z}}$ and $\set{C}_{Q, \dset{z}}$ in~\eqref{EqDefMapNormFunction}, which are   the domain and codomain of the function $\bar{K}_{Q, \dset{z}}$ in~\eqref{EqType2Krescaling}.
\begin{lemma}
\label{lemm_Type2_kset}
The set $\set{A}_{Q, \dset{z}}$ in~\eqref{EqDefMapNormFunction} is either empty or an interval of the form 
	\begin{IEEEeqnarray}{rCl}
	\label{Eq_Type2KConstrainOpen}
		\set{A}_{Q, \dset{z}} & = & 
		\begin{cases}
 			[\lambda_{Q,\dset{z}}, \infty) & \text{if } \int \frac{1}{\foo{L}_{\dset{z}}(\thetav) - \delta^{\star}_{Q, \dset{z}}}\diff Q(\thetav)< \infty \\
 			(0, \infty)& \text{otherwise},
 		\end{cases}
	\end{IEEEeqnarray}
	where
	\begin{IEEEeqnarray}{rCl}
		\lambda_{Q,\dset{z}} & = & \bar{K}^{-1}_{Q,\dset{z}}(- \delta^{\star}_{Q, \dset{z}}),
	\end{IEEEeqnarray}
	with $\bar{K}^{-1}_{Q,\dset{z}}$ in~\eqref{Eq_InvKbarEq}, the function $\foo{L}_{\dset{z}}$ is defined in~\eqref{EqLxy}, and $\delta^\star_{Q, \dset{z}}$ is defined in~\eqref{EqDefDeltaStar}. 
Moreover, the set $\set{C}_{Q, \dset{z}}$ in~\eqref{EqDefMapNormFunction} is either empty or an interval of the form
	\begin{IEEEeqnarray}{rCl}
\label{EqSeptember16at17h20in2024}	
 		\set{C}_{Q, \dset{z}} & = & 
		\begin{cases}
 			[-\delta^\star_{Q, \dset{z}}, \infty) & \text{if } \int \frac{1}{\foo{L}_{\dset{z}}(\thetav) - \delta^{\star}_{Q, \dset{z}}}\diff Q(\thetav)< \infty.\\
 			(-\delta^\star_{Q, \dset{z}}, \infty) & \text{otherwise}.
 		\end{cases}
	\end{IEEEeqnarray}
\end{lemma}
\begin{IEEEproof}
		The proof is presented in Appendix~\ref{app_proof_lemm_Type2_kset}.
\end{IEEEproof}

Lemma~\ref{lemm_Type2_kset} shows that the sets $\set{A}_{Q, \dset{z}}$ and $\set{C}_{Q, \dset{z}}$ in~\eqref{EqDefMapNormFunction} are convex sets (intervals).
Appendix~\ref{AppExamplesChangeZ} introduces some examples to illustrate particular cases in which the set $\set{A}_{Q, \dset{z}}$ is open or semi-open. 
The convexity of $\set{A}_{Q, \dset{z}}$ and $\set{C}_{Q, \dset{z}}$ is crucial for analyzing how the choice of $\lambda$ influences whether the Type-II ERM-RER problem in~\eqref{EqOpType2ERMRERNormal} has a solution.
For instance, if $\lambda \in \set{A}_{Q, \dset{z}}$, then the measure $\Pgibbs{\bar{P}}{Q}$ is the unique solution to the problem in~\eqref{EqOpType2ERMRERNormal} (Theorem~\ref{Theo_ERMType2RadNikMutualAbs}). 
Moreover, as $\lambda$ increases, the resulting \mbox{Type-II} ERM-RER problem still possesses a solution, which is formalized by the following corollary.
\begin{corollary}\label{CorSeptember16at16h52in2024}
If the \mbox{Type-II} ERM-RER problem in~\eqref{EqOpType2ERMRERNormal} possesses a solution, then, the following problem
\begin{IEEEeqnarray}{rcl}
\label{EqOpType2ERMRERSubNormal}
    \min_{P \in \bigtriangledown_{Q}(\set{M}) } &\ & \foo{R}_{\dset{z}} ( P )  + \alpha \KL{Q}{P},
\end{IEEEeqnarray}
with $\alpha \geqslant \lambda$, also possesses a solution. 
\end{corollary}
Additionally, Lemma~\ref{lemm_Type2_kset} allows identifying how small $\lambda$ in~\eqref{EqOpType2ERMRERNormal} can be, such that the \mbox{Type-II} ERM-RER problem in~\eqref{EqOpType2ERMRERNormal} still possesses a solution. 
The regularization factor $\lambda$ can be made arbitrarily close to zero in some cases, as shown hereunder.
\begin{corollary}
\label{coro_FiniteKisAZeroInf}
If the set $\set{M}$ is finite, then the set $\set{A}_{Q, \dset{z}}$ in~\eqref{EqDefMapNormFunction} is $(0, \infty)$.
\end{corollary}
Corollary~\ref{coro_FiniteKisAZeroInf} follows by noticing that if the set $\set{M}$ is finite, the subset $\set{L}^{\star}_{Q, \dset{z}}$ in~\eqref{EqDefSetLStarQz} satisfies $Q(\set{L}^{\star}_{Q, \dset{z}}) > 0$. Thus,  the integral in~\eqref{EqSeptember16at17h20in2024} is not finite, which follows from the fact that for all $\thetav \in \set{L}^{\star}_{Q, \dset{z}}$, $\foo{L}_{\dset{z}}(\thetav) - \delta^{\star}_{Q, \dset{z}} = 0$.  
Another immediate consequence of Lemma~\ref{lemm_InfDevKtype2} and Lemma~\ref{lemm_Type2_kset} is the following corollary.
\begin{corollary}
	If the real value~$\delta^\star_{Q, \dset{z}} = 0$, with~$\delta^\star_{Q, \dset{z}}$ in~\eqref{EqDefDeltaStar}, then the function~$\bar{K}_{Q, \dset{z}}$ in~\eqref{EqType2Krescaling} is strictly positive.
\end{corollary}
This section is closed by leveraging Lemma~\ref{lemm_Type2_kset} for presenting a key property of the function~$\bar{K}_{Q, \dset{z}}$ in~\eqref{EqType2Krescaling}.

\begin{lemma}
\label{lemm_Type2_KbarLambda}
	 The function~$\bar{K}_{Q, \dset{z}}$ in~\eqref{EqDefNormFunction} satisfies 
	\begin{equation}
	\label{EqLimKbarRight}
		\lim_{\lambda \rightarrow {\lambda^{\star}_{Q, \dset{z}}}^+} \bar{K}_{Q, \dset{z}}(\lambda) = -\delta^\star_{Q, \dset{z}},
	\end{equation}
	where~$\delta^\star_{Q, \dset{z}}$ and $\lambda^\star_{Q, \dset{z}}$ are defined in~\eqref{EqDefDeltaStar} and~\eqref{EqDefLambdaStar}, respectively.
\end{lemma}
\begin{IEEEproof}
	The proof is presented in Appendix~\ref{app_proof_lemm_Type2_KbarLambda}.
\end{IEEEproof}
The limit in~\eqref{EqLimKbarRight} is determined by the set of models in the support of the prior with the lowest empirical risk determined by the choice of the loss function~$\ell$ and the function~$f$ in~\eqref{EqLxy}.

%
%
\section{Properties of the Solution}

\subsection{Bounds on the Radon-Nikodym Derivative}

Note that from Theorem~\ref{Theo_ERMType2RadNikMutualAbs}, models resulting in lower empirical risks correspond to greater values of the Radon-Nikodym derivative $\frac{\diff \Pgibbs{\bar{P}}{Q}}{\diff Q}$ in~\eqref{EqGenpdfType2}. The following corollary formalizes this observation.
\begin{lemma}
\label{lemm_ERM_RER_Type2_ineq}
	For all~$(\thetav_1, \thetav_2) \in \left( \supp Q \right)^2$, such that~$\foo{L}_{\dset{z}}(\thetav_1)\leq \foo{L}_{\dset{z}}(\thetav_2)$, with~$\foo{L}_{\dset{z}}$ in~\eqref{EqLxy}, the \RadonNikodym derivative $\frac{\diff \Pgibbs{\bar{P}}{Q}}{\diff Q}$ in~\eqref{EqGenpdfType2} satisfies
	\begin{equation}\label{Eq_lem_Type2ineqRNd}
		\frac{\diff \Pgibbs{\bar{P}}{Q}}{\diff Q}(\thetav_2) \leq \frac{\diff \Pgibbs{\bar{P}}{Q}}{\diff Q}(\thetav_1),
	\end{equation}\
	with equality if and only if~$\foo{L}_{\dset{z}}(\thetav_1) = \foo{L}_{\dset{z}}(\thetav_2)$.
\end{lemma}
\begin{IEEEproof}
	The proof is presented in Appendix~\ref{app_lemm_ERM_RER_Type2_ineq}.
\end{IEEEproof}
The intuition that follows from Lemma~\ref{lemm_ERM_RER_Type2_ineq} is that, under the assumption that the ERM problem in~\eqref{EqOriginalOP} possesses a solution in the support of the reference measure, the maximum of the function $\frac{\diff \Pgibbs{\bar{P}}{Q}}{\diff Q}$ in~\eqref{EqGenpdfType2} is achieved by the models in $\set{T}(\dset{z})\cap \supp Q$, provided that it is not empty, with $\set{T}(\dset{z})$ in~\eqref{EqHatTheta}. Furthermore, the Radon-Nikodym derivative $\frac{\diff \Pgibbs{\bar{P}}{Q}}{\diff Q}$ is monotonic with respect to the empirical risk $\foo{L}_{\dset{z}}$ in~\eqref{EqLxy}. This property is similar to that of the solution to the \mbox{Type-I} ERM-RER problem in~\eqref{EqGenpdfType1}, as established in \cite[Corollary~1]{perlaza2024ERMRER}.

The Radon-Nikodym derivative~$\frac{\diff \Pgibbs{\bar{P}}{Q}}{\diff Q}$ in~\eqref{EqGenpdfType2} is always finite and strictly positive.
This observation is formalized in the following lemma.

\begin{lemma}
\label{lemm_ERM_RER_Type2_RNdBounds}
The \RadonNikodym derivative~$\frac{ \diff \Pgibbs{\bar{P}}{Q}}{\diff Q}$ in~\eqref{EqGenpdfType2} satisfies for all~$\thetav \in  \supp Q$,
	\begin{equation}
		0 < \frac{ \diff \Pgibbs{\bar{P}}{Q}}{\diff Q}(\thetav) \leq \frac{\lambda}{\delta^\star_{Q, \dset{z}} + \bar{K}_{Q, \dset{z}}(\lambda)} < \infty,
	\end{equation}
	where the function $\bar{K}_{Q, \dset{z}}$ and the real $\delta^\star_{Q, \dset{z}}$ are defined in 
	\eqref{EqDefMapNormFunction} and~\eqref{EqDefDeltaStar}, respectively. 
	The equality holds if and only if~$\thetav \in \set{L}^{\star}_{Q, \dset{z}}\cap \supp Q$, with $\set{L}^{\star}_{Q, \dset{z}}$ in~\eqref{EqDefSetLStarQz}.
\end{lemma}
\begin{IEEEproof}
	The proof is presented in Appendix~\ref{app_lemm_ERM_RER_Type2_RNdBounds}.
\end{IEEEproof}

%
%
\subsection{Asymptotes of the Radon-Nikodym Derivative}

In the asymptotic regime, when the regularization factor $\lambda$ in~\eqref{EqOpType2ERMRERNormal} grows to infinity, \ie,~$\lambda \to \infty$, the measure~$\Pgibbs{\bar{P}}{Q}$ becomes identical to the reference measure~$Q$, up to sets of measure zero,  as described in the following lemma.

\begin{lemma}
\label{lemm_T2AsymptLamb2Inf} 
	The \RadonNikodym derivative~$\frac{\diff \Pgibbs{\bar{P}}{Q}}{\diff Q}$ in~\eqref{EqGenpdfType2} satisfies for all~$\thetav \in \supp Q$,
	\begin{equation}
		\lim_{\lambda \rightarrow \infty} \frac{\diff \Pgibbs{\bar{P}}{Q}}{\diff Q}(\thetav) = 1.
	\end{equation}
\end{lemma}
\begin{IEEEproof}
	The proof is presented in Appendix~\ref{App_lemm_T2AsymptLamb2Inf}.
\end{IEEEproof}
Lemma~\ref{lemm_T2AsymptLamb2Inf} unveils a similarity between \mbox{Type-I} and \mbox{Type-II} regularization as the \mbox{Type-I} measure~$\Pgibbs{P}{Q}$ in~\eqref{EqGenpdfType1}, also exhibits a similar behavior~\cite[Lemma~7]{perlaza2024ERMRER}.

Alternatively, when the regularization factor decreases to zero from the right, \ie,~$\lambda \to 0^{+}$, the \RadonNikodym derivative~$\frac{\diff \Pgibbs{\bar{P}}{Q}}{\diff Q}$ in~\eqref{EqGenpdfType2} exhibits the following behavior.
\begin{lemma}
\label{lemm_T2AsymptLamb2Zero}
	If~$Q(\set{L}^\star_{Q, \dset{z}}) > 0$, with the set~$\set{L}^\star_{Q, \dset{z}}$ in~\eqref{EqDefSetLStarQz}, then the \RadonNikodym derivative~$\frac{\diff \Pgibbs{\bar{P}}{Q}}{\diff Q}$ in~\eqref{EqGenpdfType2} satisfies  for all~$\thetav \in \supp Q$,
	\begin{IEEEeqnarray}{rCl}
	\label{EqLemm14_s1}
		\lim_{\lambda \rightarrow 0^+} \frac{\diff \Pgibbs{\bar{P}}{Q}}{\diff Q}(\thetav) 
		& = & \frac{1}{Q(\set{L}^{\star}_{Q, \dset{z}})}\ind{\thetav \in \set{L}^{\star}_{Q, \dset{z}}}.
	\end{IEEEeqnarray}
	On the ofther hand, if~$Q(\set{L}^\star_{Q, \dset{z}})=0$ and $\lambda^\star_{Q, \dset{z}}$ in~\eqref{EqDefLambdaStar} satisfies $\lambda^\star_{Q, \dset{z}} = 0$, then for all~$\thetav \in \supp Q$, it holds that
	\begin{IEEEeqnarray}{rCl}
	\label{EqLemm14_s2}
		\lim_{\lambda \rightarrow 0^+} \frac{\diff \Pgibbs{\bar{P}}{Q}}{\diff Q}(\thetav) 
		& = & \begin{cases}
			\infty & \text{if } \thetav \in \set{L}^\star_{Q, \dset{z}}\\
			0  & \text{otherwise.} 
		\end{cases}
		\end{IEEEeqnarray}
		Conversely, if~$Q(\set{L}^\star_{Q, \dset{z}})=0$ and $\lambda^\star_{Q, \dset{z}} > 0$, then for all~$\thetav \in \supp Q$, it holds that
	\begin{IEEEeqnarray}{rCl}
	\label{EqLemm14_s3}
		\lim_{\lambda \rightarrow {\lambda^\star_{Q, \dset{z}}}^+} \frac{\diff \Pgibbs{\bar{P}}{Q}}{\diff Q}(\thetav) 
		& = & \frac{\lambda^\star_{Q, \dset{z}}}{\foo{L}_{\dset{z}}(\thetav)-\delta^\star_{Q, \dset{z}}}.
	\end{IEEEeqnarray}
\end{lemma}
\begin{IEEEproof}
	The proof is presented in Appendix~\ref{App_lemm_T2AsymptLamb2Zero}.
\end{IEEEproof}

Lemma~\ref{lemm_T2AsymptLamb2Zero} highlights that in the asymptotic regime when the regularization factor decreases to zero from the right, \ie,~$\lambda \to 0^{+}$, the value~$\frac{\diff \Pgibbs{\bar{P}}{Q}}{\diff Q}(\vect{\theta})$ does not depend on the exact model $\vect{\theta}$ but rather on whether $\thetav \in \supp Q \cap  \set{L}^{\star}_{Q, \dset{z}}$. In the case in which $\thetav \in \supp Q \cap  \set{L}^{\star}_{Q, \dset{z}}$, it holds that~$\lim_{\lambda \to 0^{+}}\frac{\diff \Pgibbs{\bar{P}}{Q}}{\diff Q}(\vect{\theta}) > 0$. Otherwise,~$\lim_{\lambda \to 0^{+}}\frac{\diff \Pgibbs{\bar{P}}{Q}}{\diff Q}(\vect{\theta}) = 0$.
In the special case in which~$\delta^\star_{Q, \dset{z}} = 0$, with~$\delta^\star_{Q, \dset{z}}$ in~\eqref{EqDefDeltaStar}, the set~$\set{L}^{\star}_{Q, \dset{z}}$ satisfies~$\set{L}^{\star}_{Q, \dset{z}} = \set{T}(\dset{z})$, where~$\set{T}(\dset{z})$ is defined in~\eqref{EqHatTheta}. This implies a concentration of probability over  $\set{T}(\dset{z}) \cap \supp Q$, which establishes a connection with the ERM problem without regularization in~\eqref{EqOriginalOP}.

Furthermore, in the asymptotic regime, when the regularization factor decreases to zero from the right, the solutions to the \mbox{Type-I} and \mbox{Type-II} ERM-RER problems exhibit the same asymptotic behavior, as shown in~\cite[Lemma 6]{perlaza2024ERMRER}.
This aligns with the observation that as~$\lambda$ decreases, the optimization problems in~\eqref{EqOpType1ERMRERNormal} and~\eqref{EqOpType2ERMRERNormal} exhibit a weaker relative entropy constraint.
A stronger result that follows from Lemma~\ref{lemm_T2AsymptLamb2Zero} is presented in the following lemma.
\begin{lemma}
\label{lemm_T2AsymptLamb2ZeroPggibs}
If $\lambda^\star_{Q, \dset{z}}$ in~\eqref{EqDefLambdaStar} satisfies $\lambda^\star_{Q, \dset{z}} = 0$, then the measure~$\Pgibbs{\bar{P}}{Q}$ in~\eqref{EqGenpdfType2} and the set~$\set{L}^\star_{Q, \dset{z}}$ in~\eqref{EqDefSetLStarQz} satisfy
	\begin{IEEEeqnarray}{rCl}
	\label{EqLemm15Pis1}
		\lim_{\lambda \rightarrow 0^+} \Pgibbs{\bar{P}}{Q}(\set{L}^\star_{Q, \dset{z}})
		& = & 1.
	\end{IEEEeqnarray}
	Alternatively, if $\lambda^\star_{Q, \dset{z}} > 0$, then the measure~$\Pgibbs{\bar{P}}{Q}$ in~\eqref{EqGenpdfType2} and the set~$\set{L}^\star_{Q, \dset{z}}$ in~\eqref{EqDefSetLStarQz} satisfy
	\begin{IEEEeqnarray}{rCl}
	\label{EqLemm15Pis0}
		\lim_{\lambda \rightarrow {\lambda^\star_{Q, \dset{z}}}^+} \Pgibbs{\bar{P}}{Q}(\set{L}^\star_{Q, \dset{z}})
		& = & 0.
	\end{IEEEeqnarray}
\end{lemma}
\begin{IEEEproof}
	The proof is presented in Appendix~\ref{proof_lemm_T2AsymptLamb2ZeroPggibs}.
\end{IEEEproof}

Lemma~\ref{lemm_T2AsymptLamb2ZeroPggibs} shows that indeed, if $\lambda^{\star}_{Q, \dset{z}} = 0$ and the regularization factor~$\lambda$ approaches zero from the right, the probability measure~$\Pgibbs{\bar{P}}{Q}$ in~\eqref{EqGenpdfType2} concentrates on a set of models that induce minimum empirical risk, \ie, the set $\set{L}^\star_{Q, \dset{z}}$ in \eqref{EqDefSetLStarQz}.
%
%
\section{The Expected Empirical Risk}
\label{sec:ExpectedER}
 
This section focuses on the expected empirical risk induced by the probability measure~$\Pgibbs{\bar{P}}{Q}$ in~\eqref{EqGenpdfType2}. That is, the value $\foo{R}_{\dset{z}}\left( \Pgibbs{\bar{P}}{Q} \right)$, with the functional~$\foo{R}_{\dset{z}}$ defined in~\eqref{EqRxy}.

The following lemma establishes a relation between $\foo{R}_{\dset{z}}\left( \Pgibbs{\bar{P}}{Q} \right)$,  $\lambda$, and the function $\bar{K}_{Q, \dset{z}}$ in~\eqref{EqType2Krescaling}.
%
\begin{lemma}
\label{lemm_T2PropertiesEmpRiskSolution}
The probability measure $\Pgibbs{\bar{P}}{Q}$ in~\eqref{EqGenpdfType2} satisfies
	\begin{equation}
	\label{EqCharExEmpRiskT2}
		\foo{R}_{\dset{z}}(\Pgibbs{\bar{P}}{Q}) = \lambda - \bar{K}_{Q, \dset{z}}(\lambda),
	\end{equation}
where the functional~$\foo{R}_{\dset{z}}$ and the function~$\bar{K}_{Q, \dset{z}}$ are defined in~\eqref{EqRxy} and~\eqref{EqType2Krescaling}, respectively.
\end{lemma}
\begin{IEEEproof}
	The proof is presented in Appendix~\ref{App_lemm_T2PropertiesEmpRiskSolution}.
\end{IEEEproof}

Lemma~\ref{lemm_T2PropertiesEmpRiskSolution} characterizes the expected empirical risk of the \mbox{Type-II} ERM-RER solution and establishes a direct connection to the regularization factor $\lambda$. For example, from Lemma~\ref{lemm_Type2_KbarLambda} if there exists a model $\thetav^{\star} \in \supp Q$, such that its empirical risk is zero, the function $\bar{K}_{Q, \dset{z}}$ is nonegative, which implies that $\lambda$ serves as an explicit upper bound to the expected empirical risk.
This provides a clear interpretation: the choice of $\lambda$ directly controls and bounds the average expected empirical risk. 
This property gives \mbox{Type-II} ERM-RER a distinct risk management strategy over that of \mbox{Type-I} ERM-RER, mainly lowering the expected empirical risk by appropriately choosing the value of $\lambda$.
Additionally, Lemma~\ref{lemm_T2PropertiesEmpRiskSolution} highlights that the the expected empirical risk $\foo{R}_{\dset{z}}(\Pgibbs{\bar{P}}{Q})$, with $Q$ and $\vect{z}$ fixed, inherits all  properties of the function $\bar{K}_{Q, \dset{z}}$ in~\eqref{EqType2Krescaling}. The following lemma formalizes this observation.
\begin{lemma}
\label{lemm_T2PropertiesEmpRisklambda}
The expected empirical risk $\foo{R}_{\dset{z}}(\Pgibbs[\dset{z}][\lambda]{\bar{P}}{Q})$, with the functional~$\foo{R}_{\dset{z}}$ in~\eqref{EqRxy} and the measure~$\Pgibbs{\bar{P}}{Q}$ in~\eqref{EqGenpdfType2}, is continuous and nondecreasing 
with respect to~$\lambda$. Moreover, it is strictly increasing if and only if the empirical risk function~$\mathsf{L}_{\vect{z}}$ in~\eqref{EqLxy} is separable with respect to the probability measure~$Q$.
\end{lemma}
\begin{IEEEproof}
	The proof is presented in Appendix~\ref{app_proof_lemm_T2PropertiesEmpRisklambda}.
\end{IEEEproof}

\subsection{Bounds on the Expected Empirical Risk}

This section builds on the characterization of the expected empirical risk and its monotonicity with respect to the regularization factor  $\lambda$ in~\eqref{EqOpType2ERMRERNormal} to establish a range of bounds on the expected empirical risk.
%
%
The following lemma highlights a connection existing between the expected empirical risks $\foo{R}_{\dset{z}}(Q)$ and $\foo{R}_{\dset{z}}(\Pgibbs{\bar{P}}{Q})$; and the relative entropy $\KL{Q}{\Pgibbs{\bar{P}}{Q}}$.
\begin{lemma}
	\label{lemm_LowBoundGenGap}
	The functional~$\foo{R}_{\dset{z}}$ defined in~\eqref{EqRxy} and the measures~$Q$ and~$\Pgibbs{\bar{P}}{Q}$ in~\eqref{EqGenpdfType2} satisfy
	\begin{equation}
	\label{EqTheEqThatShallHaveANumber}
		\foo{R}_{\dset{z}}(Q) - \foo{R}_{\dset{z}}(\Pgibbs{\bar{P}}{Q})
		 \geq  \lambda(\exp(\KL{Q}{\Pgibbs{\bar{P}}{Q}}) - 1). 
	\end{equation}
\end{lemma}
\begin{IEEEproof}
	The proof is presented in Appendix~\ref{app_proof_lemm_Type2InequalityJensen}.
\end{IEEEproof}
Note that $\KL{Q}{\Pgibbs{\bar{P}}{Q}} \geq 0$ in~\eqref{EqTheEqThatShallHaveANumber}, which leads to the observation that 
\begin{equation}
(\exp(\KL{Q}{\Pgibbs{\bar{P}}{Q}}) - 1) \geq 0.
\end{equation} 
Hence, from Lemma~\ref{lemm_LowBoundGenGap}, it follows that the solution to the \mbox{Type-II} ERM-RER problem induces an expected empirical risk that is smaller than the one induced by reference measure~$Q$. This is formalized by the following corollary.
\begin{corollary}
\label{lemm_Type2InewRzPvsRzQ}
The probability measures~$Q$ and~$\Pgibbs{\bar{P}}{Q}$ in~\eqref{EqGenpdfType2} satisfy
\begin{equation}
	\foo{R}_{\dset{z}}(\Pgibbs{\bar{P}}{Q}) \leq \foo{R}_{\dset{z}}(Q),
\end{equation}
	where the functional~$\foo{R}_{\dset{z}}$ is defined in~\eqref{EqRxy} and equality holds if and only if the empirical risk function $\foo{L}_{\dset{z}}$ in~\eqref{EqLxy} is nonseparable.
\end{corollary}
The following lemma presents a lower bound and an upper bound on the expected empirical risk $\foo{R}_{\dset{z}}(\Pgibbs{\bar{P}}{Q})$ in which the regularization parameter plays a central role. 
%
\begin{lemma}
\label{lemm_Type2BoundDeltaStar}
The probability measure $\Pgibbs{\bar{P}}{Q}$ in~\eqref{EqGenpdfType2} satisfies
	\begin{IEEEeqnarray}{rCl}
	 \delta^\star_{Q, \dset{z}}  & \leq \foo{R}_{\dset{z}}(\Pgibbs{\bar{P}}{Q})  < & \lambda + \delta^\star_{Q, \dset{z}}, 
	\end{IEEEeqnarray}
	where the functional~$\foo{R}_{\dset{z}}$ is defined in~\eqref{EqRxy} and  $\delta^\star_{Q, \dset{z}}$ is defined in~\eqref{EqDefDeltaStar}. Moreover, the inequality on the left-hand side holds with equality if and only if the empirical risk function $\mathsf{L}_{\vect{z}}$ in~\eqref{EqLxy} is nonseparable. 
\end{lemma}
\begin{IEEEproof}
	The proof is presented in Appendix~\ref{app_lemm_Type2BoundDeltaStar}.
\end{IEEEproof}
The bounds presented in Lemma~\ref{lemm_Type2BoundDeltaStar} highlight that the regularization parameter $\lambda$ in~\eqref{EqOpType2ERMRERNormal} governs the increase of the expected empirical risk  $\foo{R}_{\dset{z}}(\Pgibbs{\bar{P}}{Q})$ with respect to its minimum, i.e, $\delta^\star_{Q, \dset{z}}$ in~\eqref{EqDefDeltaStar}.  
%
Moreover, the lower bound is tight for the probability measure~$\Pgibbs{\bar{P}}{Q}$ in~\eqref{EqGenpdfType2} in the asymptotic regime when $\lambda$ decreases to $\lambda^{\star}_{Q, \dset{z}}$ from right, as shown hereunder. 
\begin{lemma}
	\label{lemm_T2limlamzeroRz}
	The probability measure~$\Pgibbs{\bar{P}}{Q}$ in~\eqref{EqGenpdfType2} satisfies
	\begin{IEEEeqnarray}{rCl}
		\lim_{\lambda \rightarrow {\lambda^{\star}_{Q, \dset{z}}}^{+}}\foo{R}_{\dset{z}}(\Pgibbs{\bar{P}}{Q}) = \lambda^{\star}_{Q, \dset{z}} + \delta^\star_{Q, \dset{z}},
	\end{IEEEeqnarray}
	where~$\delta^\star_{Q, \dset{z}}$ is defined in~\eqref{EqDefDeltaStar} and the functional~$\foo{R}_{\dset{z}}$ is defined in~\eqref{EqRxy}.
\end{lemma}
\begin{IEEEproof}
From Lemma~\ref{lemm_T2PropertiesEmpRiskSolution}, it holds that 
%
\begin{IEEEeqnarray}{rCl}
\IEEEeqnarraymulticol{3}{l}{
	\lim_{\lambda \rightarrow {\lambda^{\star}_{Q, \dset{z}}}^+} \foo{R}_{\dset{z}}(\Pgibbs{\bar{P}}{Q})
	} \nonumber \\ \quad
	& = & \lim_{\lambda \rightarrow {\lambda^{\star}_{Q, \dset{z}}}^+} \lambda - \lim_{\lambda \rightarrow {\lambda^{\star}_{Q, \dset{z}}}^+} \bar{K}_{Q, \dset{z}}(\lambda)
	\label{EqProof_T2limRzp1_s1}\\
	& = & \lambda^{\star}_{Q, \dset{z}}- \lim_{\lambda \rightarrow {\lambda^{\star}_{Q, \dset{z}}}^+} \bar{K}_{Q, \dset{z}}(\lambda)\label{EqProof_T2limRzp1_s2}\\
	& = & \lambda^{\star}_{Q, \dset{z}} +  \delta^\star_{Q, \dset{z}}, \label{EqProof_T2limRzp1_s3}
\end{IEEEeqnarray}
where equality~\eqref{EqProof_T2limRzp1_s3} follows from Lemma~\ref{lemm_Type2_KbarLambda}.
This completes the proof.	
\end{IEEEproof}

Finally, note that the functional~$\foo{R}_{\dset{z}}$ in~\eqref{EqRxy} is nonnegative. This observation together with Lemma~\ref{lemm_T2PropertiesEmpRiskSolution}  lead to a new property for the  function~$\bar{K}_{Q, \dset{z}}$ in~\eqref{EqType2Krescaling}, which is stated by the following corollary
\begin{corollary}
\label{coro_T2PropertiesEmpRiskSolution}
The function~$\bar{K}_{Q, \dset{z}}$ in~\eqref{EqType2Krescaling} satisfies, for all $t > \lambda^{\star}_{Q, \dset{z}}$, with $\lambda^{\star}_{Q, \dset{z}}$ in~\eqref{EqDefLambdaStar}, 
	\begin{equation}
	\label{EqCoroExEmpRiskT2}
		\bar{K}_{Q, \dset{z}}(t) \leq t.
	\end{equation}
\end{corollary}

\subsection{$(\delta, \epsilon)$-Optimality}
This section presents a PAC guarantee for models sampled from the probability measure~$\Pgibbs{\bar{P}}{Q}$ in~\eqref{EqGenpdfType2}, with respect to the  \mbox{Type-II} ERM-RER problem  in~\eqref{EqOpType2ERMRERNormal}.
Such guarantee is presented using the notion of $(\delta, \epsilon)$-optimality introduced in \cite[Definition 6]{perlaza2024ERMRER}.
\begin{definition}[{\cite[Definition 6]{perlaza2024ERMRER}}]
	\label{DefDeltaEpsilonOp}
	Given a pair of positive reals~$(\delta, \epsilon)$, with~$\epsilon< 1$, the probability measure~$P \in \bigtriangleup(\set{M})$ is said to be~$(\delta, \epsilon)$-optimal if the set~$\set{L}_{\dset{z}}(\delta)$ in~\eqref{EqType2LsetLamb2zero} satisfies
	\begin{IEEEeqnarray}{rCl}
	\label{EqDefDeltaEpsilonOp}
		P(\set{L}_{\dset{z}}(\delta)) & > & 1-\epsilon.	
	\end{IEEEeqnarray}
\end{definition}
The following theorem presents a~$(\delta, \epsilon)$-optimality guarantee for the solution to the \mbox{Type-II} ERM-RER problem in~\eqref{EqOpType2ERMRERNormal}.

\begin{theorem}
	\label{theo_DeltaEpsilonOp}
	Assume that $\lambda^{\star}_{Q, \dset{z}} = 0$, then for all~$(\delta, \epsilon) \in (\delta^\star_{Q, \dset{z}}, \infty)\times(0,1)$, with~$\delta^\star_{Q, \dset{z}}$ in~\eqref{EqDefDeltaStar}, there always exists a~$\lambda \in (0, \infty)$, such that the probability measure~$\Pgibbs{\bar{P}}{Q}$ in~\eqref{EqGenpdfType2} is~$(\delta, \epsilon)$-optimal.
\end{theorem}
\begin{IEEEproof}
	The proof is presented in Appendix~\ref{app_proof_theo_DeltaEpsilonOp}.
\end{IEEEproof}


%
%
\section{Interplay Between the Relative Entropy Asymmetry and the Empirical Risk}
\label{sec:logERM_RER}
This section presents a connection between the \mbox{Type-I} ERM-RER in~\eqref{EqOpType1ERMRERNormal} and \mbox{Type-II} ERM-RER in~\eqref{EqOpType2ERMRERNormal} established via a transformation of the empirical risk function. The connection is established by proving the existence of two functions $\foo{W}_{Q, \dset{z}, \lambda}:\set{M}\rightarrow \reals$ and $\foo{V}_{Q, \dset{z}, \lambda}:\set{M}\rightarrow \reals$, such that the solution to the optimization problem in~\eqref{EqOpType1ERMRERNormal} is identical to the solution of the following problem:
\begin{IEEEeqnarray}{rcl}
\label{EqOpERMLinkT2_T1}
    \min_{P \in \bigtriangledown_{Q}(\set{M})} & \quad & \int \foo{W}_{Q, \dset{z}, \lambda}(\thetav) \diff P(\thetav) + \KL{Q}{P},
\end{IEEEeqnarray}
with $\lambda$ and $Q$ in~\eqref{EqOpType1ERMRERNormal}; and the solution to the optimization problem in~\eqref{EqOpType2ERMRERNormal} is identical to the solution of the following problem:
\begin{IEEEeqnarray}{rcl}
\label{EqOpERMLinkT1_T2}
    \min_{P \in \bigtriangleup_{Q}(\set{M})} & \quad & \int \foo{V}_{Q, \dset{z}, \lambda}(\thetav) \diff P(\thetav) + \KL{P}{Q},
\end{IEEEeqnarray}
with $\lambda$ and $Q$ in~\eqref{EqOpType2ERMRERNormal}. 
The main result of this section is presented in the following theorem.

\begin{theorem}
\label{Theo_ERMlogType1To2}
If the problems in~\eqref{EqOpType1ERMRERNormal} and in~\eqref{EqOpType2ERMRERNormal} have solutions, then
\begin{subequations}
\begin{IEEEeqnarray}{rCl}
\min_{P \in \bigtriangledown_{Q}(\set{M})} & \int \foo{L}_{\dset{z}}(\thetav) \diff P(\thetav)  + \lambda \KL{Q}{P} & = \qquad \nonumber
\\
\label{EqOpType2ToType1}
\min_{P \in \bigtriangleup_{Q}(\set{M})} & \int \foo{V}_{Q, \dset{z}, \lambda}(\thetav) \diff P(\thetav)   + \KL{P}{Q}, & \
\end{IEEEeqnarray}
where the function $\foo{V}_{Q, \dset{z}, \lambda}:\set{M} \rightarrow \reals$ is defined as
\begin{IEEEeqnarray}{rCl}
\label{EqLogLxy}
\foo{V}_{Q, \dset{z}, \lambda}(\thetav) & = & \log(\bar{K}_{Q, \dset{z}}(\lambda) + \foo{L}_{\dset{z}}(\thetav)),
\end{IEEEeqnarray}
\end{subequations}
and 
\begin{subequations}
\begin{IEEEeqnarray}{rCl}
\min_{P \in \bigtriangleup_{Q}(\set{M})} & \int \foo{L}_{\dset{z}}(\thetav) \diff P(\thetav)  + \lambda \KL{P}{Q} & = \qquad \nonumber
\\
\label{EqOpType1ToType2}
\min_{P \in \bigtriangledown_{Q}(\set{M})} & \int \foo{W}_{Q, \dset{z}, \lambda}(\thetav) \diff P(\thetav)   + \KL{Q}{P}, & \
\end{IEEEeqnarray}
where the function $\foo{W}_{Q, \dset{z}, \lambda}:\set{M} \rightarrow \reals$ is such that
\begin{IEEEeqnarray}{rCl}
\label{EqFromType2ToType1}
\foo{W}_{Q, \dset{z}, \lambda}(\thetav)
& = & \frac{\lambda}{\exp(-\frac{\foo{L}_{\dset{z}}(\thetav)}{\lambda} - K_{Q,\dset{z}}(-\frac{1}{\lambda}))}-\bar{K}_{Q, \dset{z}}(\lambda),\qquad
\end{IEEEeqnarray}
\end{subequations}
where the functions $K_{Q, \dset{z}}$ and $\bar{K}_{Q, \dset{z}}$ are defined in~\eqref{EqDefKfunction} and~\eqref{EqDefNormFunction}, respectively.
\end{theorem}

\begin{IEEEproof}
Denote by $\Pgibbs{\hat{P}}{Q}$ the solution to the optimization problem in~\eqref{EqOpERMLinkT1_T2}. From Lemma~\ref{lemm_OptimalModelType1}, for all $\thetav\in \supp{Q}$, it follows that
%
\begin{IEEEeqnarray}{rcl}
\IEEEeqnarraymulticol{3}{l}{
	\frac{\diff \Pgibbs{\hat{P}}{Q}}{\diff Q}(\thetav) 
	}\nonumber \\
	& = & \frac{\exp(-\foo{V}_{Q, \dset{z}, \lambda}(\thetav))}{\int \exp(-\foo{V}_{Q, \dset{z}, \lambda}(\nuv))\diff Q(\nuv)}  
	\label{EqProoT1ToT2_s1}\\
	& = & \frac{\exp(\log(\frac{1}{\foo{L}_{\dset{z}}(\thetav) + \bar{K}_{Q, \dset{z}}(\lambda)}))}{\int\! \exp(\log(\!\frac{1}{\foo{L}_{\dset{z}}(\nuv) + \bar{K}_{Q, \dset{z}}(\lambda)}\!)\!)\!\diff Q(\nuv)} 
	\label{EqProoT1ToT2_s2}\\
	& = & \frac{(\int \frac{1}{\foo{L}_{\dset{z}}(\nuv) + \bar{K}_{Q, \dset{z}}(\lambda)}\!\diff Q(\nuv))^{-1}}{\foo{L}_{\dset{z}}(\thetav)\! + \! \bar{K}_{Q, \dset{z}}(\lambda)}
	\label{EqProoT1ToT2_s3}\\
	& = & \frac{\bar{K}^{-1}_{Q,\dset{z}}(\beta)}{\foo{L}_{\dset{z}}(\thetav)\! + \! \bar{K}_{Q, \dset{z}}(\lambda)}
	\label{EqProoT1ToT2_s32}\\
	& = & \frac{\lambda}{\foo{L}_{\dset{z}}(\thetav) + \bar{K}_{Q, \dset{z}}(\lambda)}
	\label{EqProoT1ToT2_s4}\\
	& = & \frac{\diff \Pgibbs{\bar{P}}{Q}}{\diff Q}(\thetav)\label{EqProoT1ToT2_s5} ,
	\end{IEEEeqnarray}
%
where~\eqref{EqProoT1ToT2_s2} follows from~\eqref{EqLogLxy};
\eqref{EqProoT1ToT2_s32} follows from~\eqref{Eq_InvKbarEq};~\eqref{EqProoT1ToT2_s4} follows from~\eqref{EqType2Krescaling};
and~\eqref{EqProoT1ToT2_s5} follows from Theorem~\ref{Theo_ERMType2RadNikMutualAbs}.
This completes the proof of~\eqref{EqOpType1ToType2}.

Similarly, denote by $\Pgibbs{\tilde{P}}{Q}$ the solution to the optimization problem in~\eqref{EqOpERMLinkT2_T1}. From Theorem~\ref{Theo_ERMType2RadNikMutualAbs}, for all $\thetav\in \supp{Q}$, it follows that
\begin{IEEEeqnarray}{rcl}
\IEEEeqnarraymulticol{3}{l}{
	\frac{\diff \Pgibbs{\tilde{P}}{Q}}{\diff Q}(\thetav) 
	}\nonumber \\
	& = & \frac{\lambda}{\foo{W}_{Q, \dset{z}, \lambda}(\thetav) + \bar{K}_{Q, \dset{z}}(\lambda)}  
	\label{EqProoT2ToT1_s1}\\
	& = & \frac{\lambda}{\frac{\lambda}{\exp(-\frac{\foo{L}_{\dset{z}}(\thetav)}{\lambda} - K_{Q,\dset{z}}(-\frac{1}{\lambda}))}-\bar{K}_{Q, \dset{z}}(\lambda) + \bar{K}_{Q, \dset{z}}(\lambda)}  
	\label{EqProoT2ToT1_s2}\\
	& = & \exp(-\frac{\foo{L}_{\dset{z}}(\thetav)}{\lambda} - K_{Q,\dset{z}}(-\frac{1}{\lambda}))
	\label{EqProoT2ToT1_s3}\\
	& = & \frac{\diff \Pgibbs{P}{Q}}{\diff Q}(\thetav)\label{EqProoT2ToT1_s5} ,
	\end{IEEEeqnarray}
where~\eqref{EqProoT2ToT1_s1} follows from~\eqref{EqGenpdfType2WithK};~\eqref{EqProoT2ToT1_s2} follows from~\eqref{EqFromType2ToType1};
and~\eqref{EqProoT2ToT1_s5} follows from Lemma~\ref{lemm_OptimalModelType1}.
 This completes the proof of~\eqref{EqOpType2ToType1}.
\end{IEEEproof}

Theorem~\ref{Theo_ERMlogType1To2} establishes an equivalence between the regularization of \mbox{Type-I} and \mbox{Type-II}. More specifically, 
Theorem~\ref{Theo_ERMlogType1To2} highlights that by modifying the empirical risk function $\foo{L}_{\dset{z}}$ in~\eqref{EqLogLxy} using the function $\foo{V}_{Q, \dset{z}, \lambda}$ in~\eqref{EqLogLxy}, the \mbox{Type-II} ERM-RER problem in~\eqref{EqOpType2ERMRERNormal} can be solved by solving the \mbox{Type-I} ERM-RER problem in~\eqref{EqOpERMLinkT1_T2}.  
It is noteworthy that \mbox{Type-I} regularization imposes the support of the solution to be contained within the support of the reference measure, i.e., $\supp \Pgibbs{P}{Q} \subseteq \supp Q$.
Similarly, \mbox{Type-II} regularization imposes the support of the solution to contain the support of the reference measure, i.e., $\supp Q \subseteq \supp \Pgibbs{\bar{P}}{Q}$.
Interestingly, these inclusions can be shown to be equalities from Theorem~\ref{Theo_ERMType2RadNikMutualAbs} and Lemma~\ref{lemm_OptimalModelType1}.
That is,
\begin{IEEEeqnarray}{rcl}
	\supp \Pgibbs{P}{Q} = \supp Q = \supp \Pgibbs{\bar{P}}{Q}.
\end{IEEEeqnarray}

The remainder of the section focuses on the transformation from \mbox{Type-I} to \mbox{Type-II}.
The function $\foo{V}_{Q, \dset{z}, \lambda}$ in~\eqref{EqLogLxy} is referred to as the \emph{log-empirical} risk function. The \emph{expected log-empirical risk} is defined as follows.

\begin{definition}[Expected Log-Empirical Risk]
\label{DefExLogEmpiricalRisk}
Given the dataset \mbox{$\dset{z} \in  ( \set{X} \times \set{Y} )^n$} in~\eqref{EqTheDataSet} and the log-empirical risk function $\foo{V}_{Q, \dset{z}, \lambda}$ in~\eqref{EqLogLxy}, 
let  the functional $\bar{\foo{R}}_{Q, \dset{z}, \lambda}: \bigtriangleup(\set{M})  \rightarrow \reals$ be such that
\begin{equation}
\label{EqLogRxy}
\bar{\foo{R}}_{Q, \dset{z}, \lambda}( P ) = \int \foo{V}_{Q, \dset{z}, \lambda}(\thetav)  \diff P(\thetav).
\end{equation}
The value $\bar{\foo{R}}_{Q, \dset{z}, \lambda}( P )$ is the expected log-empirical risk induced by the measure~$P$. 
\end{definition}
Using the \emph{expected log-empirical risk} defined above, the optimization problem in~\eqref{EqOpERMLinkT2_T1} can be rewritten as follows
\begin{IEEEeqnarray}{rcl}
\label{EqOpERMLinkT1_T2_v2}
    \min_{P \in \bigtriangleup_{Q}(\set{M})} & \quad & \bar{\foo{R}}_{Q, \dset{z}, \lambda}( P ) + \KL{P}{Q},
\end{IEEEeqnarray}
with $\lambda$ and $Q$ being parameters of the \mbox{Type-I} and \mbox{Type-II} ERM-RER problems in~\eqref{EqOpType1ERMRERNormal} and~\eqref{EqOpType2ERMRERNormal}.
The \mbox{Type-I} - \mbox{Type-II} relation in Theorem~\ref{Theo_ERMlogType1To2} can be used to establish an equality involving the relative entropies $\KL{Q}{\Pgibbs{\bar{P}}{Q}}$ and $\KL{\Pgibbs{\bar{P}}{Q}}{Q}$; and the expected log-empirical risks $\bar{\foo{R}}_{Q, \dset{z}, \lambda}(\Pgibbs{\bar{P}}{Q})$ and $\bar{\foo{R}}_{Q, \dset{z}, \lambda}(Q)$, as shown hereunder. 
\begin{lemma}
\label{lemm_LogEmpiricalRisk}
	The functional~$\bar{\foo{R}}_{Q, \dset{z}, \lambda}$ in~\eqref{EqLogRxy} and the probability measures~$\Pgibbs{\bar{P}}{Q}$ and $Q$ in~\eqref{EqGenpdfType2} satisfy
	\begin{IEEEeqnarray}{rCl}
	\label{EqLemm23_s1}
		\log(\lambda) 
		& = & \bar{\foo{R}}_{Q, \dset{z}, \lambda}(\Pgibbs{\bar{P}}{Q}) + \KL{\Pgibbs{\bar{P}}{Q}}{Q}\\
		\label{EqLemm23_s2}
		& = & \bar{\foo{R}}_{Q, \dset{z}, \lambda}(Q) - \KL{Q}{\Pgibbs{\bar{P}}{Q}}.
	\end{IEEEeqnarray}
\end{lemma}
\begin{IEEEproof}
	The proof is presented in Appendix~\ref{app_proof_lemm_LogEmpiricalRisk}.
\end{IEEEproof}

\subsection{Sensitivity of the Log-Empirical Risk}
The sensitivity of the expected empirical risk, as presented in \cite[Definition 7]{perlaza2024ERMRER}, is defined as follows.%
\begin{definition}[Sensitivity of the Expected Empirical Risk]
\label{DefERMSensitivity}
Consider  the functional~$\foo{R}_{\dset{z}}$ in~\eqref{EqRxy} and let~$\foo{S}_{Q, \lambda}:(\set{X}\times \set{Y})^n\times \bigtriangleup_{Q} (\set{M}) \rightarrow \reals$ be a functional  such that  
\begin{IEEEeqnarray}{rCl}
	\label{EqDefNeoSensitivity}
\foo{S}_{Q, \lambda}(\dset{z},P) & = & \foo{R}_{\dset{z}}(P) - \foo{R}_{\dset{z}}(\Pgibbs{P}{Q}),
\end{IEEEeqnarray}
where the probability measure~$\Pgibbs{P}{Q}$ is defined  in~\eqref{EqGenpdfType1}. The sensitivity of the expected empirical risk $\foo{R}_{\dset{z}}$ due to a deviation from~$\Pgibbs{P}{Q}$ to~$P$ is~$\foo{S}_{Q, \lambda}(\dset{z},P)$.
\end{definition}
Similarly, the sensitivity of the expected log-empirical risk $\foo{V}_{Q, \dset{z}, \lambda}$ in~\eqref{EqLogLxy} is defined as follows.
\begin{definition}[Sensitivity of the Expected Log-Empirical Risk]
\label{DeflogSensitivity}
Consider the functional $\bar{\foo{R}}_{Q, \dset{z}, \lambda}$ in~\eqref{EqLogRxy} and let~$\bar{\foo{S}}_{Q, \lambda}:(\set{X}\times \set{Y})^n\times \bigtriangledown_{Q} (\set{M}) \rightarrow \reals$ be a functional such that  
\begin{IEEEeqnarray}{rCl}
	\label{EqDeflogSensitivity}
	\!\!\!\!\!\!\!\! \bar{\foo{S}}_{Q, \lambda}(\dset{z},P) & = & \bar{\foo{R}}_{Q, \dset{z}, \lambda}(P) -\bar{\foo{R}}_{Q, \dset{z}, \lambda}(\Pgibbs{\bar{P}}{Q}) ,
\end{IEEEeqnarray}
where the probability measure~$\Pgibbs{\bar{P}}{Q}$ is in~\eqref{EqGenpdfType2}. The sensitivity of the expected log-empirical risk due to a deviation from~$\Pgibbs{\bar{P}}{Q}$ to~$P$ is~$ \bar{\foo{S}}_{Q, \lambda}(\dset{z},P)$.
\end{definition}

The sensitivity of the expected log-empirical risk due to a deviation from~$\Pgibbs{\bar{P}}{Q}$ to~$P$ is described by the following closed-form expression.

\begin{lemma}
\label{lemm_Gen_Gap_T2}
	The sensitivity $\bar{\foo{S}}_{Q, \lambda}$ in~\eqref{EqDeflogSensitivity} satisfies for all probability measures~$P \in \bigcirc_{Q}(\set{M})$ that
	\begin{equation}
	\label{EqGen_Gap_T2}
		\bar{\foo{S}}_{Q, \lambda}(\dset{z},P) 
		= \KL{P}{\Pgibbs{\bar{P}}{Q}} - \KL{P}{Q} + \KL{\Pgibbs{\bar{P}}{Q}}{Q},
	\end{equation}
	where the probability measures $Q$ and $\Pgibbs{\bar{P}}{Q}$ are defined in~\eqref{EqGenpdfType2}.
\end{lemma}
\begin{IEEEproof}
	The proof is presented in Appendix~\ref{app_proof_lemm_Gen_Gap_T2}.
\end{IEEEproof}

An interesting interpretation of Lemma~\ref{lemm_Gen_Gap_T2} follows from rewriting~\eqref{EqGen_Gap_T2} using the objective function of the \mbox{Type-I} ERM-RER problem in~\eqref{EqOpERMLinkT1_T2} as follows:
\begin{IEEEeqnarray}{rCl}
\KL{P}{\Pgibbs{\bar{P}}{Q}}
 	& = & 	\bar{\foo{R}}_{Q, \dset{z}, \lambda}(P) -  \bar{\foo{R}}_{Q, \dset{z}, \lambda}(\Pgibbs{\bar{P}}{Q}) 
 	\nonumber\\ &   & 
 	+ \KL{P}{Q} - \> \KL{\Pgibbs{\bar{P}}{Q}}{Q}.
\end{IEEEeqnarray}
That is, the relative entropy $\KL{P}{\Pgibbs{\bar{P}}{Q}}$ represents the variation of the objective function of the  \mbox{Type-I} ERM-RER problem in~\eqref{EqOpERMLinkT1_T2} due to a deviation from the solution $\Pgibbs{\bar{P}}{Q}$ to an alternative probability measure $P$.

In Lemma~\ref{lemm_Gen_Gap_T2}, when~$P$ is chosen to be identical to the reference measure~$Q$, it follows that
\begin{equation}
	\bar{\foo{S}}_{Q, \lambda}(\dset{z}, Q) 
	= \KL{Q}{\Pgibbs{\bar{P}}{Q}} + \KL{\Pgibbs{\bar{P}}{Q}}{Q}, 
\end{equation}
where the right-hand side is a Jeffreys divergence~\cite{jeffreys1946invariant}, also known as the symmetrized Kullback-Leibler divergence between the measures~$\Pgibbs{\bar{P}}{Q}$ and~$Q$.
Furthermore, by observing that \mbox{$\KL{\Pgibbs{\bar{P}}{Q}}{Q} \geq 0$}, and~\mbox{$\KL{Q}{\Pgibbs{\bar{P}}{Q}} \geq 0$}  \cite[Theorem $1$]{perlaza2024ERMRER}, Lemma~\ref{lemm_Gen_Gap_T2} leads to the following corollary.

\begin{corollary}
\label{coro_Gen_Gap_T2}
	The probability measures $Q$ and $\Pgibbs{\bar{P}}{Q}$ in~\eqref{EqGenpdfType2}  satisfy 
	\begin{equation}
		\bar{\foo{R}}_{Q, \dset{z}, \lambda}(\Pgibbs{\bar{P}}{Q}) \leq \bar{\foo{R}}_{Q, \dset{z}, \lambda}(Q),
	\end{equation}
	where the functional~$\bar{\foo{R}}_{Q, \dset{z}, \lambda}$ is defined in~\eqref{EqLogRxy}.
\end{corollary}
%

\subsection{\mbox{Type-I} and \mbox{Type-II} Optimal Measures}

The solutions to the optimization problems~\eqref{EqOpType1ERMRERNormal} and~\eqref{EqOpType2ERMRERNormal}, with regularization factors $\lambda$  and $\alpha$, respectively, are $\Pgibbs{P}{Q}$ and $\Pgibbs[\dset{z}][\alpha]{\bar{P}}{Q}$ in~\eqref{EqGenpdfType1} and in~\eqref{EqGenpdfType2}. These measures exhibit the following property. 

\begin{lemma}
\label{lemm_connectType1vsType2}
	The probability measures~$\Pgibbs{P}{Q}$ and $\Pgibbs[\dset{z}][\alpha]{\bar{P}}{Q}$ in~\eqref{EqGenpdfType1} and in~\eqref{EqGenpdfType2}, respectively, satisfy
	\begin{IEEEeqnarray}{rCl}
		\IEEEeqnarraymulticol{3}{l}{
		\KL{\Pgibbs{P}{Q}}{Q}-\KL{\Pgibbs[\dset{z}][\alpha]{\bar{P}}{Q}}{Q}
		}\nonumber\\ \label{EqLemmTp1Tp2ConnKl}
		& = & \log(\alpha) + K_{Q, \dset{z}}(-\frac{1}{\lambda}),
	\end{IEEEeqnarray}
	where the function~$K_{Q, \dset{z}}$ is defined in~\eqref{EqDefKfunction}.
\end{lemma}
\begin{IEEEproof}
	The proof is presented in Appendix~\ref{app_lemm_connectType1vsType2}.
\end{IEEEproof}
Lemma~\ref{lemm_connectType1vsType2} characterizes the relative entropy difference of \mbox{Type-I} and \mbox{Type-II} ERM-RER solution with respect to the prior $Q$. In doing so, it provides an alternative way to evaluate this difference without directly computing the corresponding relative entropies.


Finally, two important properties of the {\mbox{Type-I} and \mbox{Type-II} optimal measures are presented by the following corollary of Lemma~\ref{lemm_connectType1vsType2}. 
\begin{corollary}
The probability measures $\Pgibbs[\dset{z}][\alpha]{P}{Q}$ and $\Pgibbs{\bar{P}}{Q}$ in~\eqref{EqGenpdfType1} and in~\eqref{EqGenpdfType2}, respectively, satisfy
	\begin{IEEEeqnarray}{rCl}
		\IEEEeqnarraymulticol{3}{l}{
		\bar{\foo{S}}_{Q, \alpha}(\dset{z}, \Pgibbs{P}{Q})
		}\nonumber\\
		& = & \KL{\Pgibbs[\dset{z}][\alpha]{P}{Q}}{\Pgibbs{\bar{P}}{Q}}
		- (\log(\alpha) + K_{Q, \dset{z}}(-\frac{1}{\lambda})) \ \ \
		\label{EqconnT1vsT2DiffT2}\\
		\noalign{\noindent and \vspace{2\jot}}
		\IEEEeqnarraymulticol{3}{l}{
		\frac{1}{\lambda}{\foo{S}}_{Q, \lambda}(\dset{z}, \Pgibbs[\dset{z}][\alpha]{\bar{P}}{Q})
		}\nonumber\\
		& = & \KL{\Pgibbs{\bar{P}}{Q}}{\Pgibbs[\dset{z}][\alpha]{P}{Q}}
		+ \log(\alpha) + K_{Q, \dset{z}}(-\frac{1}{\lambda}),
		\label{EqconnT1vsT2DiffT1}
	\end{IEEEeqnarray}
	where the functionals~$\foo{S}_{Q, \lambda}$ and~$\bar{\foo{S}}_{Q, \alpha}$ are respectively defined in~\eqref{EqDefNeoSensitivity} and  in~\eqref{EqDeflogSensitivity};
	and the function~$K_{Q, \dset{z}}$ is defined in~\eqref{EqDefKfunction}.
\end{corollary}

The equality in~\eqref{EqconnT1vsT2DiffT2} quantifies the variation of the expected log-empirical risk due to a deviation from the probability measure $\Pgibbs{P}{Q}$ in~\eqref{EqGenpdfType1} to the probability measure $\Pgibbs[\dset{z}][\alpha]{\bar{P}}{Q}$ in~\eqref{EqGenpdfType2} via the sensitivity $\bar{\foo{S}}_{Q, \alpha}(\dset{z}, \Pgibbs{P}{Q})$.
The equality in~\eqref{EqconnT1vsT2DiffT1} quantifies the variation of the expected empirical risk due to a deviation from the probability measure $\Pgibbs[\dset{z}][\alpha]{\bar{P}}{Q}$ in~\eqref{EqGenpdfType2} to the probability measure  $\Pgibbs{P}{Q}$ in~\eqref{EqGenpdfType1} via the sensitivity $\foo{S}_{Q, \lambda}(\dset{z}, \Pgibbs[\dset{z}][\alpha]{\bar{P}}{Q})$.

\subsection{Numerical Comparison of \mbox{Type-I} and \mbox{Type-II} Regularization}
\label{sec:subSimExample}

In machine learning, the generalization error describes the capacity of a learning algorithm to select models based on training data that performs well with unseen test data.
The sensitivity (Definition~\ref{DefERMSensitivity}) of the optimization problem in~\eqref{EqOpType1ERMRERNormal} is closely related to the generalization error \cite[Theorem $16$]{perlaza2024ERMRER}. See also \cite{zou2024WorstCase, zou2024generalization}. 
A probability measure $P \in \bigtriangleup_{Q}(\set{M})$, e.g., a machine learning algorithm, that yields larger sensitivity indicates that the learning overfits with respect to the training data,  leading to an increase in the generalization error~\cite{zou2024generalization}. In this context, algorithms arising from the \mbox{Type-I} and \mbox{Type-II} ERM-RER are used for the classification of two handwritten numbers from the MNIST dataset~\cite{lecun1998gradient}.
The MNIST example is simplified to accommodate a parameterized model in $\reals^2$ such that the numerical approximations of the generalization error for different regularization factors are meaningful \cite{perlaza2024HAL}. 

The MNIST dataset consists of $60{,}000$ images for training and $10{,}000$ images for testing. Out of the $60{,}000$ training images, $12{,}183$ are labeled as the digits six and seven, while $1{,}986$ out of the $10{,}000$ test images correspond to these digits. Each image is a $28 \times 28$ grayscale picture and is represented by a  matrix in $[0,1]^{28\times28}$. To reduce the computational complexity, the pictures are processed following the procedure described in Appendix~\ref{AppNumericalSimulation}.
Consider the \mbox{Type-I} ERM-RER problem in~\eqref{EqOpType1ERMRERNormal} and the \mbox{Type-II} ERM-RER problem in~\eqref{EqOpType2ERMRERNormal} and assume that:
	$(i)$ the set of models is $\set{M} = [-50,50]^2$; $(ii)$ the set of patterns $\set{X}$ is formed by computing the histogram of gradients (HOG) of the pictures such that $\set{X} \subset \reals^{1296}$ of the handwritten six and seven in the MNIST dataset; $(iii)$ the set of labels is $\set{Y} = \{6,7\}$; $(iv)$ the reference measure $Q$ is chosen to be a uniform probability measure over the set of models; $(v)$ the function $f$ in~\eqref{EqLxy} is defined as
	\begin{IEEEeqnarray}{rCl}
		f(\thetav, x) 
		& = & 
		\begin{cases}
			6 & \text{if } 0 < (x\vect{W})\thetav,\\
			7 & \text{if } 0 > (x\vect{W})\thetav, 
		\end{cases}
	\end{IEEEeqnarray}
	where the matrix $\vect{W}$ is defined in~\eqref{EqDefWPACsim} in Appendix~\ref{AppNumericalSimulation};
	and $(vi)$ the loss function $\ell$ in~\eqref{EqEll} satisfies
	\begin{IEEEeqnarray}{rCl}
		\ell(f(\thetav, x), y) 
		& = & \ind{f(\thetav, x)\neq y}.
	\end{IEEEeqnarray}
	For the simulation, $8{,}100$ data points are uniformly sampled from the $12{,}183$ available training images, forming the dataset $\dset{z}_1$, referred to as the \emph{training dataset}. Similarly, $1{,}300$ data points are uniformly sampled from the $1{,}986$ available test images, forming the dataset $\dset{z}_2$, referred to as the \emph{test dataset}.
	Not all images are used because the simulation is repeated only $100$ times, and at each iteration, the datasets $\dset{z}_1$ and $\dset{z}_2$ are uniformly resampled.
	\begin{figure}[h]
	\centering
	\begin{minipage}{0.46\linewidth}
		\centering
        \includegraphics[width=0.95\linewidth]{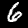}
  		\caption{ $28\times28$ Image of a handwritten 6 from MNIST dataset.}
	\end{minipage}
	\begin{minipage}{0.08\linewidth}	
	\end{minipage}
	\begin{minipage}{0.46\linewidth}
		\centering
    	\includegraphics[width=0.95\linewidth]{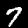}
    	\caption{ $28\times28$ Image of a handwritten 7 from MNIST dataset.}
	\end{minipage}
	\end{figure}
	\begin{figure}[h]
		\centering
        \includegraphics[width=1\linewidth]{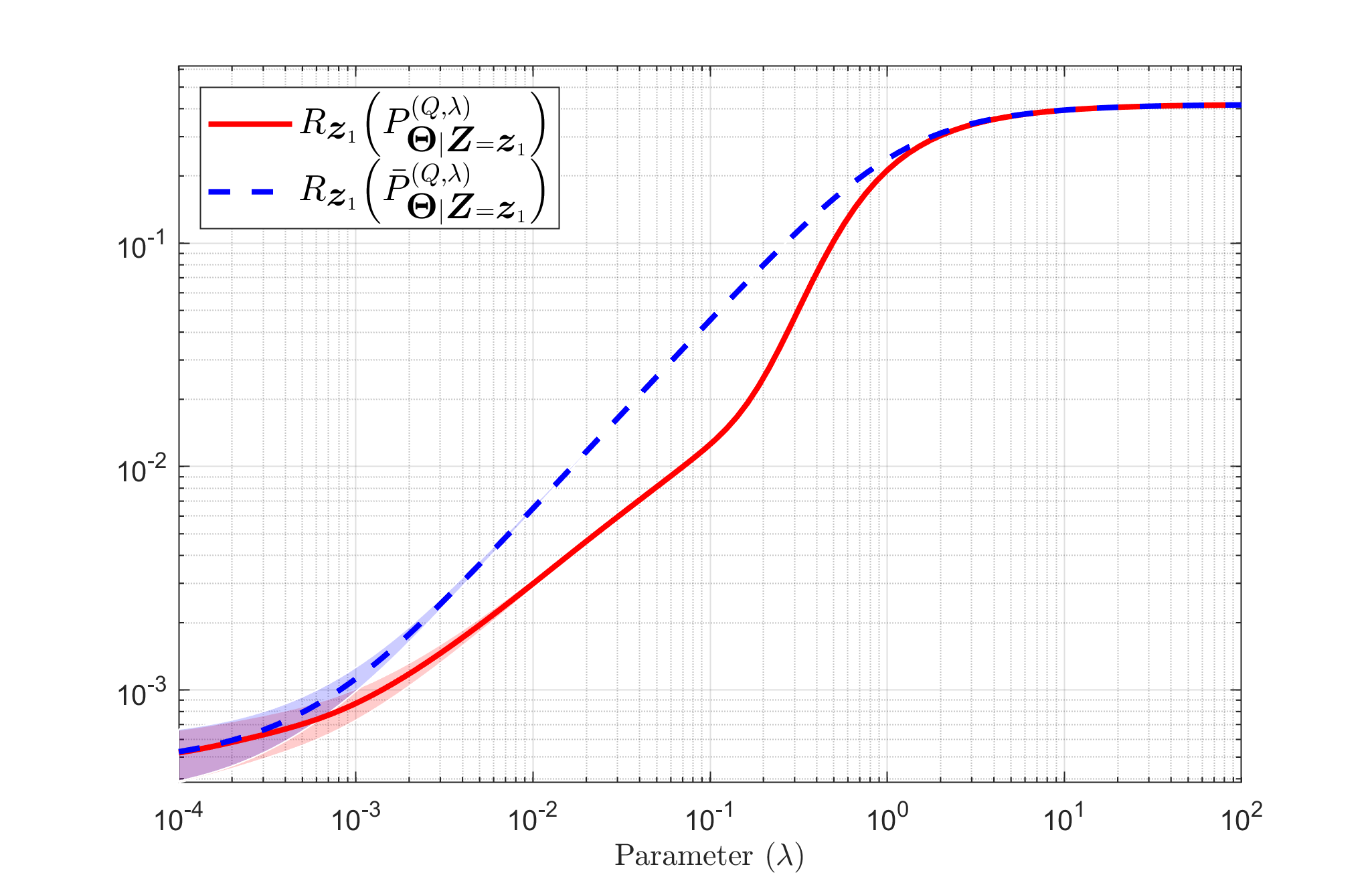}
  		\caption{Average Training Error: average of the expected empirical risks $\foo{R}_{\dset{z}_1}(\Pgibbs[\dset{z}_1]{P}{Q})$ and $\foo{R}_{\dset{z}_1}(\Pgibbs[\dset{z}_1]{\bar{P}}{Q})$, with the measures $\Pgibbs[\dset{z}_1]{P}{Q}$ and $\Pgibbs[\dset{z}_1]{\bar{P}}{Q}$ in~\eqref{EqGenpdfType1} and~\eqref{EqGenpdfType2}, respectively, computed over one hundred different $\dset{z}_1$ (training dataset) random selections.}
  		\label{FigTrainingPlot}
	\end{figure}
	\begin{figure}[h]
		\centering
        \includegraphics[width=1\linewidth]{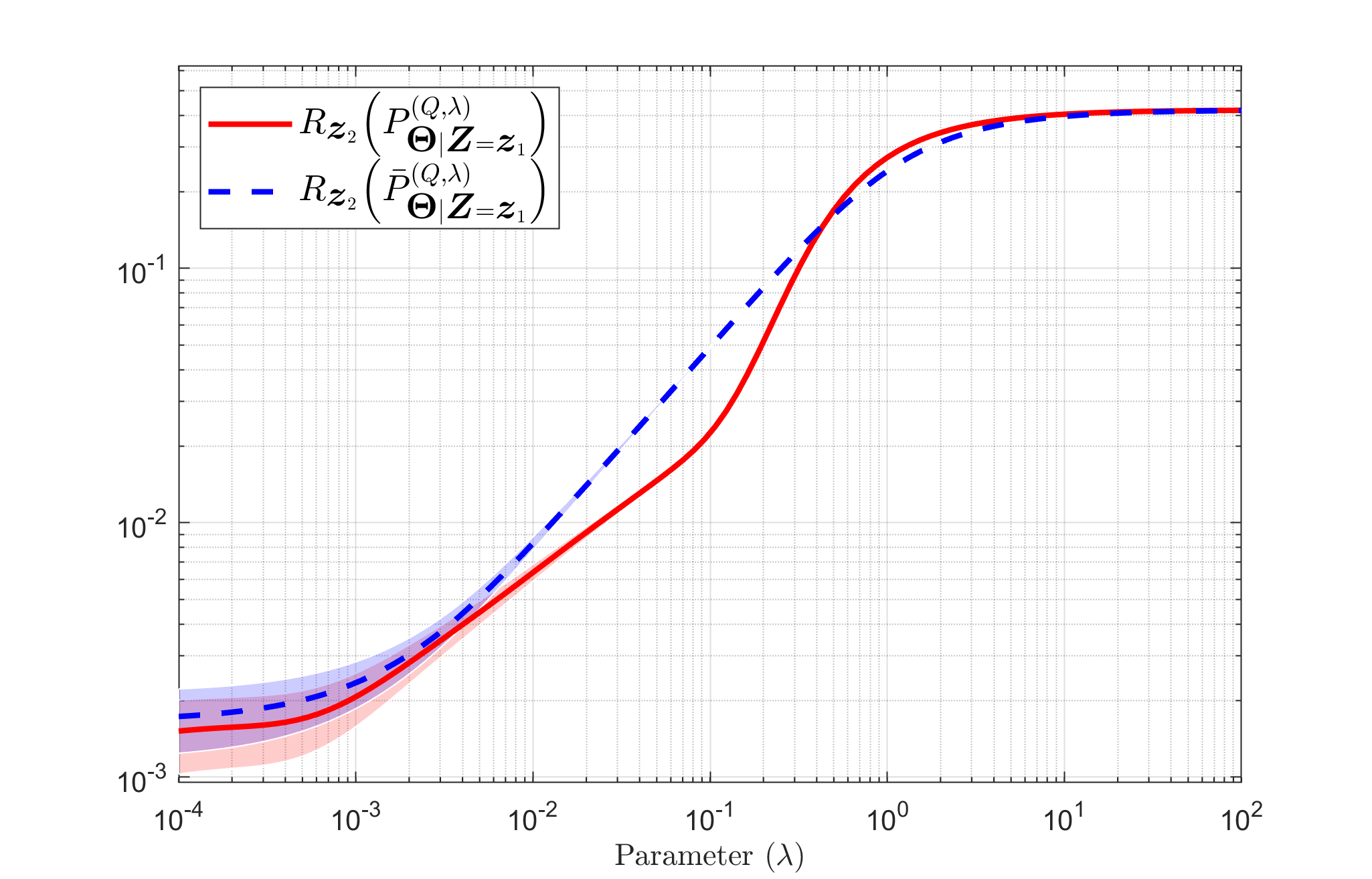}
  		\caption{Average Test Error: average of the expected empirical risks $\foo{R}_{\dset{z}_2}(\Pgibbs[\dset{z}_1]{P}{Q})$ and $\foo{R}_{\dset{z}_2}(\Pgibbs[\dset{z}_1]{\bar{P}}{Q})$, with the measures $\Pgibbs[\dset{z}_1]{P}{Q}$ and $\Pgibbs[\dset{z}_1]{\bar{P}}{Q}$ in~\eqref{EqGenpdfType1} and~\eqref{EqGenpdfType2}, respectively, computed over one hundred different $\dset{z}_1$ (training dataset) and $\dset{z}_2$ (test dataset) random selections.}
  		\label{FigRiskPlot}
	\end{figure}
	\begin{figure}[h]
		\centering
    	\includegraphics[width=1.05\linewidth]{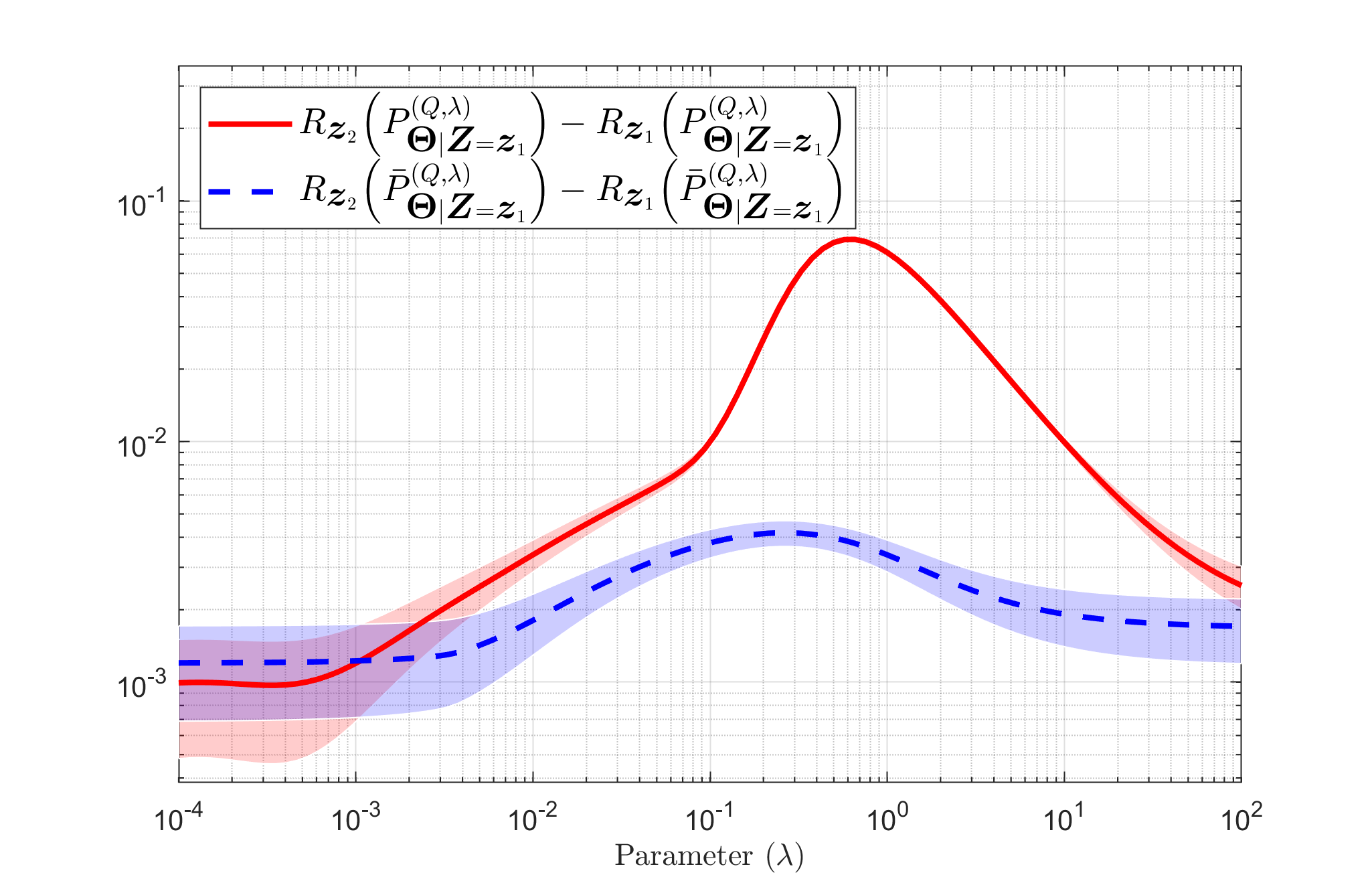}
    	\caption{ Average of the differences (generalization gaps) $\foo{R}_{\dset{z}_2}(\Pgibbs[\dset{z}_1]{P}{Q})- \foo{R}_{\dset{z}_1}(\Pgibbs[\dset{z}_1]{P}{Q})$ and $\foo{R}_{\dset{z}_2}(\Pgibbs[\dset{z}_1]{\bar{P}}{Q})- \foo{R}_{\dset{z}_1}(\Pgibbs[\dset{z}_1]{\bar{P}}{Q})$, with  the measures $\Pgibbs[\dset{z}_1]{P}{Q}$ and $\Pgibbs[\dset{z}_1]{\bar{P}}{Q}$ in~\eqref{EqGenpdfType1} and~\eqref{EqGenpdfType2}, respectively, computed over one hundred different $\dset{z}_1$ (training dataset) and $\dset{z}_2$ (test dataset) random selections.}
    	\label{FigSensPlot}
	\end{figure}
	Figure~\ref{FigSensPlot} displays the average (over 100 repetitions) generalization gap for \mbox{Type-I} and \mbox{Type-II} algorithms. The average generalization gap of \mbox{Type-II} suggests it is less prone to overfitting. In contrast, Figure~\ref{FigTrainingPlot} and Figure~\ref{FigRiskPlot} show that \mbox{Type-I} achieves a lower average training error, which results on a lower average test error. 
	These observations imply that \mbox{Type-II} promotes a more positive conservative learning, reducing the generalization gap by keeping average training and testing error closer, while \mbox{Type-I} achieves lower training error at the cost of a higher avergae generalization gap, indicating greater reliance on the training data. 

	Another key observation is that, for certain ranges of the regularization factor (e.g.,$\lambda \in (0.002, 0.09)$ and $\lambda \in (0.3, 0.8)$ for \mbox{Type-II}), where the average test error are comparable, \mbox{Type-II} exhibits lower average generalization gap. This suggests that \mbox{Type-II} can achieve both small test error and generalization error for certain regularization factor, a highly desirable outcome. However, selecting such regularization ranges remains an open question for future research under the theoretical framework presented in this paper.
%

%
%
\section{Final Remarks}
\label{sec:FinalRemarks}

This work has introduced the \mbox{Type-II} ERM-RER problem and has presented its solution through Theorem~\ref{Theo_ERMType2RadNikMutualAbs}.
The solution highlights that regardless of whether \mbox{Type-I} or \mbox{Type-II} regularization  is used in ERM problems, the models that are considered by the resulting solution are necessarily in the support of the reference measure. In this sense, the restriction over the models introduced by the reference measure cannot be bypassed by the training data when relative entropy is used as the regularizer.
This limitation has been shown to be a consequence of the equivalence that can be established between \mbox{Type-I} and \mbox{Type-II} regularization. 
These analytical results lead to an operationally meaningful characterization of the expected empirical risk induced by the \mbox{Type-II} solution in terms of the regularization parameters.  
The closed-form expressions for the expected empirical risk induced by \mbox{Type-I} and \mbox{Type-II} errors are used to characterize the sensitivity of the expected empirical risk and the sensitivity of the expected log-empirical risk, in terms of the cumulant generating function and Kullback-Leibler divergence.
The analysis of the solution to the optimization problem~\eqref{EqOpType2ERMRERNormal} shows that, under mild assumptions, there always exists a positive real value $\lambda$ such, with probability $1 - \epsilon$, the measure $\Pgibbs{\bar{P}}{Q}$ concentrates on the set of models that minimize the empirical risk.

\appendices

\section{Preliminary}
This appendix introduces a preliminary result, which is central in the proof of Lemma~\ref{lemm_Type2RNDbigcirc}.
\begin{lemma}
\label{lemm_ERM_fDR_DiffZero}
Let $\field{M}$ be the set of measurable functions $h:\set{M} \rightarrow \reals$, with respect to the measurable space $\msblspc{\set{M}}$ and $\bormsblspc{\reals}$. Let $\field{S}$ be the subset of $\field{M}$ including all nonnegative functions that are absolutely integrable with respect to a probability measure $Q$. That is, for all $h \in \field{S}$, it holds that
\begin{IEEEeqnarray}{rCl}
\label{EqApril20at11h54in2024atHome}
\int \abs{h(\thetav)}\diff Q(\thetav) & < & \infty.
\end{IEEEeqnarray}
Let the function $\hat{r}: \reals \rightarrow \reals$ be such that
\begin{IEEEeqnarray}{rCl}
\label{EqHatR_lemmAppx}
	\hat{r}(\alpha) & = & \int -\log(g(\thetav) + \alpha h(\thetav)) \diff Q(\thetav),
\end{IEEEeqnarray}
for some functions $g$ and $h$ in $\field{S}$.
The function $\hat{r}$ in~\eqref{EqHatR_lemmAppx} is differentiable at zero.
\end{lemma}
\begin{IEEEproof}
The objective is to prove that the function $\hat{r}$ in~\eqref{EqHatR_lemmAppx} is differentiable at zero, which reduces to proving that the limit
\begin{IEEEeqnarray}{rCl}
\label{EqLimf}
	& \lim_{\delta \rightarrow 0} \frac{1}{\delta}(\hat{r}(\alpha+\delta) -\hat{r}(\alpha))&
\end{IEEEeqnarray}
exists for all $\alpha \in (-\epsilon, \epsilon)$, with $\epsilon > 0$ arbitrarily small. 
Let the function $f:(0,\infty)\rightarrow \reals$ be a function such that
\begin{IEEEeqnarray}{rCl}
\label{EqfIsLog}
	f(x) & = & -\log(x).
\end{IEEEeqnarray}
Note that the function $\hat{r}$ can be written in terms of $f$ as follows:
\begin{IEEEeqnarray}{rCl}
\label{EqHatR_lemmAppx_fRevKL}
	\hat{r}(\alpha) & = & \int f(g(\thetav) + \alpha h(\thetav)) \diff Q(\thetav).
\end{IEEEeqnarray}
The proof of the existence of such limit in~\eqref{EqLimf} relies on the fact that the function $f$ in~\eqref{EqfIsLog} is strictly convex and differentiable, which implies that $f$ is also Lipschitz continuous. Hence, it follows that for all $\vect{\theta} \in \supp Q$,
\begin{IEEEeqnarray}{rCl}
	\IEEEeqnarraymulticol{3}{l}{
	\abs{f(g(\thetav) + (\alpha+\delta) h(\thetav))-f(g(\thetav) + \alpha h(\thetav))} 
	} \nonumber \\ \label{EqfisLipschitz}
	& \leq & c\abs{h(\thetav)}\abs{\delta},
\end{IEEEeqnarray}
for some positive and finite constant $c$, which implies that
\begin{IEEEeqnarray}{rCl}
	\IEEEeqnarraymulticol{3}{l}{
	\frac{\abs{f(g(\thetav) + (\alpha+\delta) h(\thetav))-f(g(\thetav) + \alpha h(\thetav))}}{\abs{\delta}} 
	} \nonumber \\ \label{EqDfDxLipschitz} 
	& \leq & c\abs{h(\thetav)},
\end{IEEEeqnarray}
and thus, given that $h \in \field{S}$, it holds that
\begin{IEEEeqnarray}{rCl}
\IEEEeqnarraymulticol{3}{l}{
	\int \frac{\abs{f(g(\thetav) + (\alpha+\delta) h(\thetav))-f(g(\thetav) + \alpha h(\thetav))}}{\abs{\delta}} \diff Q(\thetav) 
	} \nonumber \\ \label{EqLimitExists}
	& \leq & \infty.
\end{IEEEeqnarray}
This facilitates using the dominated convergence theorem as follows. From the fact that the function $f$ is differentiable, the limit in~\eqref{EqLimf} satisfies for $\alpha \in (-\epsilon,\epsilon)$, with $\epsilon > 0$ arbitrarily small,
\begin{subequations}
\label{EqLimHatrDCT}
	\begin{IEEEeqnarray}{rCl}
	\IEEEeqnarraymulticol{3}{l}{
	\lim_{\delta \rightarrow 0} \frac{1}{\delta}(\hat{r}(\alpha+\delta) -\hat{r}(\alpha)) 
	}\nonumber\\
	& = & \lim_{\delta \rightarrow 0} \frac{1}{\delta}(\int f(g(\thetav) + (\alpha+\delta) h(\thetav)) \diff Q(\thetav)
	\right. \nonumber \\ &   & \left.
	-\int f(g(\thetav) + \alpha h(\thetav)) \diff Q(\thetav))
	\label{EqLimHatrDCT_s1}\\
	& = & \lim_{\delta \rightarrow 0} \int \frac{1}{\delta}(f(g(\thetav) + (\alpha+\delta) h(\thetav)) 
	\right. \nonumber \\ &   & \left.
	- f(g(\thetav) + \alpha h(\thetav))) \diff Q(\thetav)
	\label{EqLimHatrDCT_s2}\\
	& \leq & \lim_{\delta \rightarrow 0} \int \frac{1}{\abs{\delta}}|f(g(\thetav) + (\alpha+\delta) h(\thetav)) 
	 \nonumber \\ &   & 
	- f(g(\thetav) + \alpha h(\thetav))| \diff Q(\thetav)
	\label{EqLimHatrDCT_s3}\\
	& = & \int \lim_{\delta \rightarrow 0} \frac{1}{\abs{\delta}}|f(g(\thetav) + (\alpha+\delta) h(\thetav)) 
	 \nonumber \\ &   & 
	- f(g(\thetav) + \alpha h(\thetav))| \diff Q(\thetav)
	\label{EqLimHatrDCT_s4}\\
	& = & c \int h(\vect{\theta}) \diff Q(\thetav)
	\label{EqLimHatrDCT_s5}
	\\
	& < &  \infty,
	\label{EqLimHatrDCT_s6}
	\end{IEEEeqnarray}
\end{subequations}
where~\eqref{EqLimHatrDCT_s3}  follows from \cite[Theorem 1.5.9(c)]{ash2000probability}; 
~\eqref{EqLimHatrDCT_s4} follows from the dominated convergence theorem \cite[Theorem 1.6.9]{ash2000probability};~\eqref{EqLimHatrDCT_s5} follows from \eqref{EqDfDxLipschitz}; and~\eqref{EqLimHatrDCT_s6} follows from~\eqref{EqApril20at11h54in2024atHome}.
 Finally, from~\eqref{EqLimHatrDCT_s6}, it follows that the function $\hat{r}$ in~\eqref{EqHatR_lemmAppx} is differentiable at zero, which completes the proof.
\end{IEEEproof}

\section{Proof of Lemma~\ref{lemm_Type2RNDbigcirc}}
%

\label{app_lemm_Type2RNDbigcirc}
The optimization problem in~\eqref{EqOpType2ERMRERancillary} can be re-written in terms of the Radon-Nikodym derivative of the optimization measure~$P$ with respect to the measure $Q$, which yields:
		\begin{subequations}
		\label{EqOpProblemType2fproof}
		\begin{IEEEeqnarray}{ccl}
		\min_{P \in \bigcirc_{Q}(\set{M})}
		&   & \ \int \foo{L}_{\vect{z}}(\thetav)\frac{\diff P}{\diff Q}(\thetav)\, \diff Q(\thetav) 
		\nonumber\\ &   & 
		- \> \lambda \int \log(\frac{\diff P}{\diff Q}(\thetav))\diff Q(\thetav), \\
		\label{Eq_ProofT2Constraint}
		\mathrm{s.t.}  & &  \int \frac{\diff P}{\diff Q}(\thetav)  \diff Q(\thetav) = 1.
		\end{IEEEeqnarray}
		\end{subequations}
The remainder of the proof focuses on the problem in which the optimization is over the function  $\frac{\diff P}{\diff Q}: \set{M} \to \reals$, instead of the measure $P$. This is due to the fact that for all $P \in \bigcirc_{Q}\left( \set{M} \right)$, the Radon-Nikodym derivate $\frac{\diff P}{\diff Q}$ is unique up to sets of zero measure with respect to $Q$~ \cite[Theorem 2.2.1]{ash2000probability}.
Let $\mathscr{S}$ be the set defined in Lemma~\ref{lemm_ERM_fDR_DiffZero}.
Using this notation, the optimization problem of interest is:
\begin{subequations}
\label{EqOpProblemType2fproof9876}
\begin{IEEEeqnarray}{ccl}
\min_{g \in \mathscr{S}} & &  \int \foo{L}_{\vect{z}}(\thetav) g(\thetav)\, \diff Q(\thetav)- \lambda \int \log(g(\thetav))\diff Q(\thetav) \\
\label{Eq_ProofT2Constraint9876}
\mathrm{s.t.}  & &  \int g(\thetav)  \diff Q(\thetav) = 1.
\end{IEEEeqnarray}
\end{subequations}
Let the Lagrangian of the optimization problem in~\eqref{EqOpProblemType2fproof9876} be $L: \mathscr{S}\times \reals \rightarrow \reals$ such that
\begin{IEEEeqnarray}{rcl}
\label{EqFunctionalAll}
   L (g, \beta) 
   & = & \int \foo{L}_{\vect{z}}(\thetav)g(\thetav)\, \diff Q(\thetav) - \lambda \int \log(g(\thetav))\, \diff Q(\thetav)  
   \nonumber\\&   & 
   + \> \beta (\int g(\thetav) \, \diff Q(\thetav)-1) \\
   & = & \int\! \Big(\! g(\thetav) (\foo{L}_{\vect{z}}(\thetav)\! +\! \beta)\! -\! \lambda  \log(g(\thetav))\! \Big)\! \diff Q(\thetav)
   - \> \beta, \qquad
\end{IEEEeqnarray}
where $\beta$ is a  real that acts as a Lagrange multiplier due to the constraint~\eqref{Eq_ProofT2Constraint9876}.
Let $\hat{g}: \set{M} \rightarrow \reals$ be a function in $\mathscr{S}$. 
The Gateaux differential of the functional $L$ in~\eqref{EqFunctionalAll} at $\left(g, \beta\right) \in \mathscr{S}\times \reals$ in the direction of $\hat{g}$ is
\begin{IEEEeqnarray}{rcl}
\label{EqNecessaryCondtionType2f}
    \partial L(g, \beta; \hat{g} ) & \triangleq & \left.\frac{\diff}{\diff \gamma} L(g + \gamma \hat{g}, \beta)\right|_{\gamma = 0}.
\end{IEEEeqnarray}

Let the function $r:\reals \rightarrow \reals$ be defined for some fixed functions $g$ and $\hat{g}$ and some fixed $\beta$ such that for all $\gamma \in (-\epsilon, \epsilon)$, with $\epsilon$ arbitrarily small, 
\begin{IEEEeqnarray}{rcl}
\label{EqApril20at12h41in2025atHome}
    r(\gamma) & = & L(g + \gamma \hat{g}, \beta).
\end{IEEEeqnarray}
The proof follows by showing that the function $r$ in \eqref{EqApril20at12h41in2025atHome} is differentiable at zero, in order to prove the existence of the Gateaux differential in \eqref{EqNecessaryCondtionType2f} for those  functions $g$ and $\hat{g}$ and real $\beta$.
For doing so, note that
\begin{subequations}
\label{EqrType2f}
\begin{IEEEeqnarray}{rcl}
    r(\gamma) 
    & = & \int \foo{L}_{\vect{z}}(\thetav)(g (\thetav) + \gamma \hat{g}(\thetav))\, \diff Q(\thetav)
    \nonumber \\&   &
    - \> \lambda \int \log(g(\thetav) + \gamma \hat{g}(\thetav))\, \diff Q(\thetav) 
    \nonumber \\ &    & 
    + \> \beta(\int(g(\thetav)  + \gamma \hat{g}(\thetav))\diff Q(\thetav) - 1) \\
	& = &  \gamma \int\hat{g}(\thetav)(\foo{L}_{\vect{z}}(\thetav)  +\beta)\diff Q(\thetav) 
	\nonumber\\&   &
	- \> \lambda \int \log(g(\thetav) + \gamma \hat{g}(\thetav))\, \diff Q(\thetav)
	\nonumber\\ &    &
	+ \int g (\thetav) ( \foo{L}_{\vect{z}}(\thetav)  + \beta) \diff Q(\thetav) - \beta,\label{EqYoLloroCorrigiendoEstaPrueba}
\end{IEEEeqnarray}
\end{subequations}
where the first term in~\eqref{EqYoLloroCorrigiendoEstaPrueba} is linear with respect to $\gamma$ and  the third term is independent of $\gamma$. The second term can be written using the function $\hat{r}:\reals \rightarrow \reals$, which is defined for fixed functions $g$ and $\hat{g}$ as follows
\begin{IEEEeqnarray}{rCl}
\label{EqHatR_lemmAppBDef}
	\hat{r}(\gamma) & = & -\int \log(g(\thetav) + \gamma \hat{g}(\thetav)) \diff Q(\thetav),
\end{IEEEeqnarray}
and is verified to be differentiable at zero in Lemma~\ref{lemm_ERM_fDR_DiffZero}.
This implies that
\begin{IEEEeqnarray}{rcl}
    r(\gamma)
	& = &  \gamma \int\hat{g}(\thetav)(\foo{L}_{\vect{z}}(\thetav)  +\beta)\diff Q(\thetav) +\lambda\hat{r}(\gamma) 
	\nonumber\\ &    &
	+ \int g (\thetav) ( \foo{L}_{\vect{z}}(\thetav)  + \beta) \diff Q(\thetav) - \beta,
\end{IEEEeqnarray}
which verifies that the function $r$ in \eqref{EqApril20at12h41in2025atHome} is differentiable at zero and more importantly, verifies that the Gateaux differential $ \partial L(g, \beta; \hat{g} )$ in \eqref{EqNecessaryCondtionType2f} exists.

The proof proceeds by calculating the Gateaux differential  $\partial L(g, \beta; \hat{g} )$ in \eqref{EqNecessaryCondtionType2f}, which requires calculating the derivative of the real function $r$ in~\eqref{EqApril20at12h41in2025atHome}. That is,
\begin{IEEEeqnarray}{rcl}
  \frac{\diff}{\diff \gamma}r(\gamma)\,& = & \int\hat{g}(\thetav)(\foo{L}_{\vect{z}}(\thetav)  +\beta)\diff Q(\thetav)
  \nonumber\\&   &
  - \lambda \int \frac{\hat{g}(\thetav)}{(g(\thetav) + \gamma \hat{g}(\thetav))} \, \diff Q(\thetav)\\
  \label{EqDerrType2f} & = & \!\int\!\hat{g}(\thetav)(\!\foo{L}_{\vect{z}}(\thetav)  +\beta \vphantom{\dot{f}} -  \frac{\lambda}{(g(\thetav) + \gamma \hat{g}(\thetav))} \!)\! \diff Q(\thetav).\IEEEeqnarraynumspace
\end{IEEEeqnarray}
From~\eqref{EqNecessaryCondtionType2f} and~\eqref{EqDerrType2f}, it follows that
\begin{IEEEeqnarray}{rcl}
\label{EqGateauxDiffType2f}
	\partial L(g, \beta; \hat{g} )
	& = & \int\hat{g}(\thetav)(\foo{L}_{\vect{z}}(\thetav)  +\beta \vphantom{\dot{f}} -  \frac{\lambda}{g(\thetav)} )\! \diff Q(\thetav). 
\end{IEEEeqnarray}
The relevance of the Gateaux differential in~\eqref{EqGateauxDiffType2f} stems from \cite[Theorem $1$, page $178$]{luenberger1997bookOptimization}, which unveils the fact that a necessary condition for the functional $L$ in~\eqref{EqFunctionalAll} to have a stationary point at $( \frac{\diff  \Pgibbs{P}{Q}}{\diff Q}, \beta ) \in \set{M} \times [0, \infty)$ is that for all functions $\hat{g} \in \mathscr{S}$,
\begin{equation}
\label{EqConditionhiType2f}
\partial L(\frac{\diff \Pgibbs{P}{Q}}{\diff Q}, \beta; \hat{g}) = 0 .
\end{equation}
From~\eqref{EqGateauxDiffType2f} and~\eqref{EqConditionhiType2f}, it follows that $\frac{\diff \Pgibbs{P}{Q}}{\diff Q} $ must satisfy for all functions $\hat{g}$ in $\mathscr{S}$ that
\begin{equation}
\label{EqIntegralHisZero}
 \int\hat{g}(\thetav)(\foo{L}_{\vect{z}}(\thetav)  + \beta  -  \lambda 
 ( \frac{\diff \Pgibbs{P}{Q}}{\diff Q}(\thetav) )^{-1}  ) \diff Q(\thetav) = 0.
\end{equation}
This implies that for all $\thetav \in \supp Q$,
\begin{equation}
\label{Eq_ProofT2SemiLemma_proof}
\foo{L}_{\vect{z}}(\thetav)  +\beta  -  \lambda  (\frac{\diff \Pgibbs{P}{Q}}{\diff Q}(\thetav)  )^{-1}   = 0,
\end{equation}
and thus,
\begin{equation}
\label{EqKeyNorthStarx}
  \frac{\diff \Pgibbs{P}{Q}}{\diff Q}(\thetav) = \frac{\lambda}{\beta + \foo{L}_{\dset{z}}(\thetav)},
\end{equation}
where $\beta$ is chosen to satisfy~\eqref{Eq_ProofT2Constraint9876} and guarantee that for all $\vect{\theta} \in \supp Q$, it holds that $ \frac{\diff \Pgibbs{P}{Q}}{\diff Q}(\thetav) \in (0, \infty)$. That is, 
\begin{equation}
\beta \in \left\lbrace t\in \reals: \forall \vect{\theta} \in \supp Q , 0 < \frac{\lambda}{t + \foo{L}_{\vect{z}}(\thetav)}  \right\rbrace, \text{ and} 
\end{equation}
\begin{equation}
1 = \int \frac{\lambda}{\foo{L}_{\dset{z}}(\thetav)+\beta}\diff Q(\thetav),
\end{equation}
which is an assumption of the theorem.

The proof continues by verifying that the measure $\Pgibbs{\bar{P}}{Q}$ that satisfies~\eqref{EqKeyNorthStarx} is the unique solution to the optimization problem in~\eqref{EqOpProblemType2fproof}. Such verification is done by showing that the objective function in~\eqref{EqOpProblemType2fproof} is strictly convex with the optimization variable. Let $P_1$ and $P_2$ be two different probability measures in $\msblspc{\set{M}}$ and let $\alpha$ be in $(0,1)$. Hence,
\begin{IEEEeqnarray}{rCl}
	\IEEEeqnarraymulticol{3}{l}{
	\foo{R}_{\dset{z}} (\alpha P_1 +(1-\alpha)P_2) + \lambda\KL{\alpha P_1 +(1-\alpha)P_2}{Q}
	}\nonumber \quad\\
	& = & \foo{R}_{\dset{z}}(\alpha P_1) + \foo{R}_{\dset{z}}((1-\alpha)P_2) 
	\nonumber\\ &   & 
	+ \lambda\KL{\alpha P_1 +(1-\alpha)P_2}{Q}\\
	& > & \alpha\foo{R}_{\dset{z}}( P_1) + (1-\alpha)\foo{R}_{\dset{z}}(P_2) \nonumber\\
	&   & 
	+\> \lambda(\alpha\KL{ P_1}{Q}+ (1-\alpha)\KL{P_2}{Q}),
\end{IEEEeqnarray}
where the functional $\foo{R}_{\dset{z}}$ is defined in~\eqref{EqRxy}. 
The equality above follows from the properties of the Lebesgue integral, while the inequality follows from \cite[Theorem 2]{perlaza2024ERMRER}.
This proves that the solution is unique due to the strict concavity of the objective function, which completes the proof. \proofIEEEend
%

\section{Proof of Lemma~\ref{lemm_RadNikDevMutualIneq}}
\label{AppProof_lemm_RadNikDevMutualIneq}
%
Given a probability measure $V \in \bigtriangledown_{Q}(\set{M})$, with $\bigtriangledown_{Q}(\set{M})$ in~\eqref{DefSetTriangDown}, it follows that 
\begin{IEEEeqnarray}{rCl}
\supp Q & \subseteq  & \supp V.
\end{IEEEeqnarray}
Using this observation, let $V_0$ and $V_1$ be two probability measures on the measurable space $\msblspc{\set{M}}$ such that for all $\set{A} \in \field{F}$, it holds that
\begin{subequations}
\label{Eq_ProofT2V0andV1}
\begin{equation} \label{Eq_ProofT2V0}
	V_0(\set{A}) = \frac{V(\set{A}\setminus \supp Q)}{V(\set{M}\setminus\supp Q)},
\end{equation}
and
\begin{equation} \label{Eq_ProofT2V1}
	V_1(\set{A}) = \frac{V(\set{A} \cap \supp Q)}{V(\set{M} \cap \supp Q)}.
\end{equation}
\end{subequations}
Let the real value $\alpha$ be
\begin{equation}
	\alpha \triangleq V(\set{M} \cap \supp Q) \in (0,1],
\end{equation}
which implies that
\begin{equation}
	1-\alpha \triangleq V(\set{M} \setminus \supp Q) \in [0,1).
\end{equation}
Hence, for all $\set{A} \in \field{F}$, the measure $V$ satisfies that
\begin{equation}
\label{EqProofT2IneqVdefV0V1}
	V(\set{A}) = (1-\alpha) V_0(\set{A}) + \alpha V_1(\set{A}).
\end{equation}
Moreover, 
%
from the definition of $\bigtriangledown_{Q}(\set{M})$ in~\eqref{DefSetTriangDown}, it follows that  the probability measure $Q$ is absolutely continuous with respect to $V$.
Hence, for all measurable sets $\set{A}$, it follows that
\begin{subequations}
\begin{IEEEeqnarray}{rCl}
\label{EqProofT2AbsContQV1_s1}
Q(\set{A}) 
	& = & \int_{\set{A}} \diff Q(\thetav)\\
	& = & \int_{\set{A}} \frac{\diff Q}{\diff V}(\thetav) \diff V(\thetav)\\
	& = & \int_{\set{A}} \frac{\diff Q}{\diff V}(\thetav) \diff ((1-\alpha) V_0 + \alpha V_1)(\thetav)\\
	& = &  (1-\alpha) \int_{\set{A}} \frac{\diff Q}{\diff V}(\thetav) \diff V_0(\thetav)
	\nonumber\\
	\label{EqApril20at13h22in2025atHome}
	&   & + \alpha \int_{\set{A}}\frac{\diff Q}{\diff V}(\thetav) \diff V_1(\thetav).
\end{IEEEeqnarray}
\end{subequations}
Note that if $\set{A} \subseteq \supp Q$, it follows that
\begin{IEEEeqnarray}{rCl}
0 < Q(\set{A}) 
	\label{EqApril21at17h210in2025atHome}	
	& = & \int_{\set{A}} \alpha\frac{\diff Q}{\diff V}(\thetav) \diff V_1(\thetav),\label{EqProofT2AbsContQV1_s5}
\end{IEEEeqnarray}
which follows from the fact that $V_0(\set{A}) = 0$ under the assumption that $\set{A} \subseteq \supp Q$.
%
%
Moreover, noting that the measures $Q$ and $V_1$ are mutually absolutely continuous, it follows  from~the Radon-Nikodym Theorem \cite[Theorem $2.2.1$]{ash2000probability}, that the Radon-Nikodym derivative of $Q$ with respect to $V_1$ exists. Moreover, from  \eqref{EqApril21at17h210in2025atHome}, it follows that for all $\vect{\theta} \in \supp Q$, it holds that 
\begin{IEEEeqnarray}{rCl}
\label{EqProofT2dQsVTodQdV1}
	\frac{\diff Q}{\diff V_1}(\thetav) & = & \alpha \frac{\diff Q}{\diff V}(\thetav). 
\end{IEEEeqnarray}
Alternatively, for all measurable sets $\set{A}$, with $\set{A} \subseteq \set{M} \setminus \supp Q$, it follows that
\begin{IEEEeqnarray}{rCl}
\label{EqApril21at17h07in2025atHome}
0 & = &Q(\set{A}) =\int_{\set{A}} (1-\alpha)\frac{\diff Q}{\diff V}(\thetav) \diff V_0(\thetav),
\end{IEEEeqnarray}
which follows from the fact that $V_1(\set{A}) = 0$ under the assumption that $\set{A} \subseteq \set{M} \setminus \supp Q$.
Hence, it holds that for all $\vect{\theta} \in \set{M} \setminus \supp Q$,  $\frac{\diff Q}{\diff V}(\thetav) = 0$.
In a nutshell, for all $\vect{\theta} \in \supp V$,
\begin{IEEEeqnarray}{rCl}
\label{EqApril21at19h03in2025atHome}
\frac{\diff Q}{\diff V}(\thetav) = \left\lbrace
\begin{array}{cc}
\frac{1}{\alpha} \frac{\diff Q}{\diff V_1}(\thetav) & \text{ if }  \vect{\theta} \in \supp Q\\
0 & \text{otherwise}.
\end{array}
\right.
\end{IEEEeqnarray}

From~\eqref{EqApril21at19h03in2025atHome}, the following holds:
\begin{subequations}
\label{EqProofT2KLVisV1}
\begin{IEEEeqnarray}{rCl}
\label{EqProofT2KLVisV1_s1}
 \KL{Q}{V} 
 & = & \KLshort[\thetav]{Q}{V}\\
 & = & \int \log(\frac{1}{\alpha}\frac{\diff Q}{\diff V_1}(\thetav))\diff Q (\thetav)\label{EqProofT2KLVisV1_s2}\\
 & = & \int \log(\frac{\diff Q}{\diff V_1}(\thetav))\diff Q (\thetav) 
 \nonumber\\ &   & 
 - \> \int \log(\alpha)\diff Q(\thetav)\\
 & = & 	\KL{Q}{V_1} - \log(\alpha).\label{EqProofT2KLVisV1_s4}
\end{IEEEeqnarray}
\end{subequations}
From~\eqref{EqProofT2KLVisV1}, it follows that
\begin{subequations}
\label{EqProofT2RzKLVisV1}
\begin{IEEEeqnarray}{rCl}
\IEEEeqnarraymulticol{3}{l}{
 \!\!\!\foo{R}_{\dset{z}}(V) + \lambda \KL{Q}{V} 
 } \nonumber \\\label{EqProofT2RzKLVisV1_s1}
 & = & \foo{R}_{\dset{z}}((1-\alpha)V_0 + \alpha V_1) + \lambda\KL{Q}{V_1} 
 \nonumber \\ &   &
 - \lambda\log(\alpha)\\
 & = & (1-\alpha)\foo{R}_{\dset{z}}(V_0) + \alpha \foo{R}_{\dset{z}}(V_1)  
 + \lambda\KL{Q}{V_1} 
 \nonumber \\ &   &
 - \> \lambda\log(\alpha)\label{EqProofT2RzKLVisV1_s2}\\
 & \geq & \alpha \foo{R}_{\dset{z}}(V_1) + \lambda\KL{Q}{V_1}, \label{EqProofT2RzKLVisV1_s3}
\end{IEEEeqnarray}
\end{subequations}
with equality if and only if $\alpha = 1$, which implies that for all 
\begin{IEEEeqnarray}{rCl}
P^\star \in \arg \min_{V\in \bigtriangledown_{Q}(\set{M})}	\foo{R}_{\dset{z}}(V) + \lambda \KL{Q}{V},
\end{IEEEeqnarray}
it holds that $P^\star \in \bigcirc_{Q}(\set{M})$, which follows from observing that

%
\begin{equation}
	\supp Q = \supp P^\star,
\end{equation}
which implies \eqref{Eq_ProofT2IneOp2} and completes the proof\proofIEEEend

\section{Proof of Lemma~\ref{lemm_InfDevKtype2}}
\label{AppProofLemmaInfDevKtype2}
The proof is divided into two parts. 
The first part proves the monotonicity of the function $\bar{K}^{-1}_{Q, \dset{z}}$ in~\eqref{Eq_InvKbarEq}; while 
the second part proves the continuity of the function $\bar{K}^{-1}_{Q, \dset{z}}$.
The proof is finalized by using the continuous inverse theorem \cite[Theorem 5.6.5]{bartle2000introduction} to show both the monotonicity and continuity of the function $\bar{K}_{Q, \dset{z}}$ in~\eqref{EqDefNormFunction}.


The first part is as follows. Let $\lambda$ and $\beta$ be two reals that satisfy~\eqref{EqType2Krescaling}.
Hence, $0 < \lambda < \infty$ and from~\eqref{Eq_ProofLambdaIsTheInvOfKbar_s2}, it holds that
\begin{IEEEeqnarray}{rCl}
\label{Eq_ProofKbarNoLambdaIsFinite}	
	0 < \int \frac{1}{\foo{L}_{\dset{z}}(\thetav) + \beta}\diff Q(\thetav)
	& < & \infty,
\end{IEEEeqnarray}
which, together with~\eqref{Eq_InvKbarEq}, imply
\begin{IEEEeqnarray}{rCl}
 \infty > \bar{K}^{-1}_{Q, \dset{z}}(\beta) & > & 0.
\end{IEEEeqnarray}
That is, the function $\bar{K}^{-1}_{Q, \dset{z}}$ in~\eqref{Eq_InvKbarEq} is positive and finite. 
Using this observation, let the reals $\gamma_1$ and  $\gamma_2$ be elements of  the set $\set{C}_{Q, \dset{z}}$, with $\set{C}_{Q, \dset{z}}$ in~\eqref{EqDefMapNormFunction} and $\gamma_1<\gamma_2$. Hence, for all $\thetav \in \supp Q$, it holds that
\begin{IEEEeqnarray}{rCl}
\label{Eq_ProofKbarNoLambdaIsGam1vs2Singular}	
	 \frac{1}{\foo{L}_{\dset{z}}(\thetav) + \gamma_1}
	& > &  \frac{1}{\foo{L}_{\dset{z}}(\thetav) + \gamma_2},
\end{IEEEeqnarray}
which implies that
\begin{IEEEeqnarray}{rCl}
\label{Eq_ProofKbarNoLambdaIsGam1vs2}	
	\int \frac{1}{\foo{L}_{\dset{z}}(\thetav) + \gamma_1}\diff Q(\thetav)
	& > & \int \frac{1}{\foo{L}_{\dset{z}}(\thetav) + \gamma_2}\diff Q(\thetav),
\end{IEEEeqnarray}
and thus, from~\eqref{Eq_InvKbarEq}, it holds that 
\begin{IEEEeqnarray}{rCl}
\label{Eq_ProofKbarStricIncreaseGam1vs2}
\bar{K}^{-1}_{Q, \dset{z}}(\gamma_1)  & < & 	\bar{K}^{-1}_{Q, \dset{z}}(\gamma_2).
\end{IEEEeqnarray}
This proves that the function $\bar{K}^{-1}_{Q, \dset{z}}$ in~\eqref{Eq_InvKbarEq} is strictly increasing, and completes the first part of the proof. 


In the second part, the objective is to prove the continuity of the function $\bar{K}^{-1}_{Q, \vect{z}}$. To do so, two auxiliary functions are introduced and proven to be continuous. Then, the fact that $\bar{K}^{-1}_{Q, \vect{z}}$ in~\eqref{Eq_InvKbarEq} is the composition of the two auxiliary functions is leveraged to prove its continuity.
Let the function $h:(0,\infty) \rightarrow (0,\infty)$ be 
\begin{IEEEeqnarray}{rcl}
\label{EqProofhkWAWALem5}
h(x) = \frac{1}{x}.
\end{IEEEeqnarray}
Let also the function $k: \set{C}_{Q, \dset{z}} \to (0, \infty)$, be such that
\begin{IEEEeqnarray}{rcl}
\label{EqkWAWALem5}
k(\gamma) & = &  \int h(\gamma + \foo{L}_{\dset{z}}(\thetav)) \diff Q (\vect{\theta}).
\end{IEEEeqnarray}
The first step is to prove that the function $k$ in~\eqref{EqkWAWALem5} is continuous in $\set{C}_{Q,\dset{z}}$. This is proved by showing that $k$ always exhibits a limit in $\set{C}_{Q,\dset{z}}$.
Note that if $\gamma \in \set{C}_{Q,\dset{z}}$, with $\set{C}_{Q,\dset{z}}$ in~\eqref{EqDefNormFunction}, then from~\eqref{EqType2KrescConstrain1}, it follows that for all $\thetav \in \supp Q$, the inequality $\foo{L}_{\dset{z}}(\thetav) + \gamma > 0$ holds, which implies that  $\gamma > -\delta^{\star}_{Q, \dset{z}}$, with $\delta^{\star}_{Q, \dset{z}}$ in~\eqref{EqDefDeltaStar}.
Hence, the proof of continuity of the function $k$ in~\eqref{EqkWAWALem5} is restricted to $\left( -\delta^{\star}_{Q, \dset{z}}, \infty \right)$. 

For two models $\thetav_1$ and $\thetav_2$ in $\supp Q$, such that $\foo{L}_{\dset{z}}(\thetav_1) < \foo{L}_{\dset{z}}(\thetav_2)$, the function $h$ satisfies
\begin{IEEEeqnarray}{rcl}
\label{EqThoseEyesISawTodaysubLem5}
h(\gamma + \foo{L}_{\dset{z}}(\thetav_1)) & >  & h(\gamma + \foo{L}_{\dset{z}}(\thetav_2) ). 
\end{IEEEeqnarray}
Then, for all $\gamma \in \left( -\delta^{\star}_{Q, \dset{z}}, \infty \right)$ and for all $\vect{\theta} \in \supp Q$, it holds that 
\begin{IEEEeqnarray}{rcl}
\label{EqThoseEyesISawTodayLem5}
h(\gamma + \delta^{\star}_{Q, \dset{z}} ) & \geq  & h(\gamma + \foo{L}_{\dset{z}}(\thetav)), 
\end{IEEEeqnarray}
where equality holds if and only if $ \foo{L}_{\dset{z}}(\thetav)  = \delta^{\star}_{Q, \dset{z}}$.
The function $h$ is continuous, and thus, for all $\vect{\theta} \in \supp Q$ and for all $a \in (-\delta^{\star}_{Q, \dset{z}}, \infty)$, it holds that
\begin{IEEEeqnarray}{rcl}
\label{EqSuchABeautiflFaceLem5}
\lim_{\gamma \to a} h(\gamma + \foo{L}_{\dset{z}}(\thetav)) & = &  h(a + \foo{L}_{\dset{z}}(\thetav)).
\end{IEEEeqnarray}
Hence, from the dominated convergence theorem~\cite[Theorem~$1.6.9$]{ash2000probability}, the following limit exists and satisfies
\begin{subequations}
\label{EqSuchACrazyDay2Lem5}
\begin{IEEEeqnarray}{rcl}
\label{EqSuchACrazyDay2Lem5_s1}
\lim_{\gamma \to a} k(b) & = & \lim_{\gamma \to a}  \int h(\gamma + \foo{L}_{\dset{z}}(\thetav)) \diff Q (\vect{\theta})\\
& = &  \int  (\lim_{\gamma \to a} h(\gamma + \foo{L}_{\dset{z}}(\thetav)) )\diff Q (\vect{\theta})\\
\label{EqSeptember18at8h48in2024}
& = & \int  h(a + \foo{L}_{\dset{z}}(\thetav)) \diff Q (\vect{\theta})\\
\label{EqSeptember18at8h50in2024}
& = & k(a),
\end{IEEEeqnarray}
\end{subequations} 
where~\eqref{EqSeptember18at8h48in2024} follows from~\eqref{EqSuchABeautiflFaceLem5}.
The equality in~\eqref{EqSeptember18at8h50in2024} proves that the function $k$ in~\eqref{EqkWAWALem5} is continuous in the interval $(-\delta^{\star}_{Q,\dset{z}}, \infty)$.
Note that from~\eqref{Eq_InvKbarEq} and~\eqref{EqkWAWALem5}, it holds that
\begin{IEEEeqnarray}{rCl}
\label{EqProofLimInvKbarOpenPreLem5}
	\bar{K}^{-1}_{Q, \dset{z}}(\gamma) & = & \frac{1}{k(\gamma)}.
\end{IEEEeqnarray}
Using~\eqref{EqProofLimInvKbarOpenPreLem5}, for all $a \in (-\delta^{\star}_{Q, \dset{z}}, \infty)$, it holds that 
\begin{IEEEeqnarray}{rCl}
\label{EqProofLimInvKbarOpenLem5_s1}
	\lim_{\gamma \to a} \bar{K}^{-1}_{Q, \dset{z}}(\gamma)
	& = & \lim_{\gamma \to a} \frac{1}{k(\gamma)}\\
	\label{EqProofLimInvKbarOpenLem5_s2}
	& = & \frac{1}{\displaystyle{\lim_{\gamma \to a}k(\gamma)}}\\
	\label{EqProofLimInvKbarOpenLem5_s3}
	& = & \frac{1}{\int h(a + \foo{L}_{\dset{z}}(\thetav) )\diff Q (\vect{\theta})}\\
	\label{EqProofLimInvKbarOpenLem5_s4}
	& = & \frac{1}{\int \frac{1}{a + \foo{L}_{\dset{z}}(\thetav)}\diff Q (\vect{\theta})}\\
	\label{EqProofLimInvKbarOpenLem5_s5}
	& = & \bar{K}^{-1}_{Q, \dset{z}}(a),
\end{IEEEeqnarray}
where~\eqref{EqProofLimInvKbarOpenLem5_s2} follows from the continuity of the function $h$ in~\eqref{EqProofhkWAWALem5} over the interval $(0,\infty)$;~\eqref{EqProofLimInvKbarOpenLem5_s3} follows from~\eqref{EqSeptember18at8h50in2024}; and~\eqref{EqProofLimInvKbarOpenLem5_s4} follows from~\eqref{Eq_InvKbarEq}.
Thus, the existence of the limit  in~\eqref{EqProofLimInvKbarOpenLem5_s1} implies that the function $\bar{K}^{-1}_{Q, \dset{z}}$ is continuous in $\set{C}_{Q, \dset{z}}$. This completes the second part of the proof.

The proof ends by using the continuous inverse theorem \cite[Theorem 5.6.5]{bartle2000introduction}. That is, given that the function $\bar{K}^{-1}_{Q, \dset{z}}$ is both continuous and strictly increasing, then, so is the function $\bar{K}_{Q, \dset{z}}$  in~\eqref{EqDefNormFunction}.  This concludes the proof. \proofIEEEend
\section{Proof of Lemma~\ref{lemm_Type2_kset}}\label{app_proof_lemm_Type2_kset}

%
The proof is divided into two parts. In the first part,  it is shown that the set $\set{C}_{Q, \dset{z}}$  is an interval of $\reals$. In the second part, the set $\set{A}_{Q, \dset{z}}$ is shown to be also an interval.
The first part uses a partition of $\reals$ formed by the following sets: $\left( -\infty,  -\delta^\star_{Q, \dset{z}}\right)$;  $\left(-\delta^\star_{Q, \dset{z}}, \infty \right)$; and $\lbrace \delta^\star_{Q, \dset{z}} \rbrace$, with $\delta^\star_{Q, \dset{z}}$ in~\eqref{EqDefDeltaStar}.
Each of these intervals is studied separately.   
 
Let $\beta$ be such that $\bar{K}_{Q, \dset{z}}(\lambda)  =  \beta$, with $\lambda$ in~\eqref{EqOpType2ERMRERNormal} and assume that $\beta \in \left( -\infty,  -\delta^\star_{Q, \dset{z}}\right)$. 
Under this assumption, the inclusion in~\eqref{EqType2KrescConstrain1} does not hold. 
This follows from the fact that, if $\beta < -\delta^\star_{Q, \dset{z}}$, for all  $\thetav \in \{\nuv \in \supp Q : \delta^\star_{Q, \dset{z}} \leq \foo{L}_{\dset{z}}(\nuv) < -\beta\}$, it holds that $\foo{L}_{\dset{z}}(\thetav) + \beta <0$, which contradicts~\eqref{EqType2KrescConstrain1}.
%
This implies that 
\begin{IEEEeqnarray}{rCl}
\label{EqCase1NotInCQz}
\left( -\infty,  -\delta^\star_{Q, \dset{z}}\right) \cap \set{C}_{Q, \dset{z}} = \emptyset.
\end{IEEEeqnarray}

Assume now that $\beta\in \left(-\delta^\star_{Q, \dset{z}}, \infty \right)$. Then, from~\eqref{EqDefDeltaStar}, it can be verified that the constraint in~\eqref{EqType2KrescConstrain1} is satisfied. More specifically, for all $\thetav \in \supp Q$, it holds that $\foo{L}_{\dset{z}}(\thetav) + \beta > 0$. 
The proof continuous by showing that~\eqref{EqType2KrescConstrain2} is also verified. For this purpose, note that
\begin{subequations}
\label{Eq_ProofType2DenLambdaEqV} 
\begin{IEEEeqnarray}{rCl}
\label{Eq_ProofType2DenLambdaEqV_s1} 
	\int \frac{1}{\foo{L}_{\dset{z}}(\thetav) + \beta}\diff Q(\thetav)
	& \leq & \int \frac{1}{\delta^\star_{Q, \dset{z}}+\beta} \diff Q(\thetav)\\
	& = &  \frac{1}{\delta^\star_{Q, \dset{z}}+\beta}\\
	& < & \infty.\label{Eq_ProofType2DenLambdaEqV_s2}
\end{IEEEeqnarray}
\end{subequations}
The finiteness of the integral in the left-hand side of~\eqref{Eq_ProofType2DenLambdaEqV_s1} implies that
%
\begin{subequations}
\label{EqProofLamFinite}
\begin{IEEEeqnarray}{rCl}
\label{EqProofLamFinite_s1}
	0 & < & \lambda\\
	 \label{EqProofLamFinite_s2}
	& = & \bar{K}^{-1}_{Q, \dset{z}}(\beta) \\
	\label{EqProofLamFinite_s3}
	& = & \frac{1}{\int \frac{1}{\foo{L}_{\dset{z}}(\thetav) + \beta}\diff Q(\thetav)}\\
	\label{EqProofLamFinite_s4}
	& < & \infty,
\end{IEEEeqnarray}
\end{subequations}
where~\eqref{EqProofLamFinite_s1} follows from the assumption that $\lambda \in \set{A}_{Q, \dset{z}} \subseteq (0,\infty)$;~\eqref{EqProofLamFinite_s2} follows from the fact that  $\bar{K}_{Q, \dset{z}}(\lambda)  =  \beta$;~\eqref{EqProofLamFinite_s3} follows from~\eqref{Eq_InvKbarEq}; and~\eqref{EqProofLamFinite_s4} follows from the inequality in~\eqref{Eq_ProofType2DenLambdaEqV_s2}.
In a nutshell, 
\begin{IEEEeqnarray}{rCl}
	\int \frac{1}{\foo{L}_{\dset{z}}(\thetav) + \beta}\diff Q(\thetav) & < & \infty, \mbox{ and }\\
	\lambda & < & \infty,
\end{IEEEeqnarray}
which, implies that the product
\begin{subequations}
\begin{IEEEeqnarray}{rCl}
\label{EqSep17_1317_s1}
\lambda\int \frac{1}{\foo{L}_{\dset{z}}(\thetav) + \beta}\diff Q(\thetav)	& = & \int \frac{\lambda}{\foo{L}_{\dset{z}}(\thetav) + \beta}\diff Q(\thetav)  \\
\label{EqSep17_1317_s2}
	& = &  \int \frac{\frac{1}{\int \frac{1}{\foo{L}_{\dset{z}}(\nuv) + \beta}\diff Q(\nuv)}}{\foo{L}_{\dset{z}}(\thetav) + \beta}\diff Q(\thetav) \qquad\\
	& = & 1,
\end{IEEEeqnarray}
\end{subequations}
where~\eqref{EqSep17_1317_s2} follows from~\eqref{EqProofLamFinite_s3}. This verifies~\eqref{EqType2KrescConstrain2}, which implies that
\begin{IEEEeqnarray}{rCl}
\label{EqCase2InCQz}
\left(-\delta^\star_{Q, \dset{z}}, \infty \right) & \subseteq & \set{C}_{Q, \dset{z}}.
\end{IEEEeqnarray}

Finally, under the assumption that $\beta = -\delta^{\star}_{Q, \dset{z}}$, two cases are considered: $(a)$ $Q(\set{L}^{\star}_{Q, \dset{z}})>0$; and $(b)$ $Q(\set{L}^{\star}_{Q, \dset{z}})=0$, with $\set{L}^{\star}_{Q, \dset{z}}$ defined in~\eqref{EqDefSetLStarQz}.
In case $(a)$, if $\beta = -\delta^\star_{Q, \dset{z}}$ and $Q(\set{L}^{\star}_{Q, \dset{z}})>0$, then for all $\thetav \in \set{L}^{\star}_{Q, \dset{z}}$, it follows that
\begin{IEEEeqnarray}{rCl}
\frac{1}{\foo{L}_{\dset{z}}(\thetav)-\delta^\star_{Q, \dset{z}}}Q(\set{L}^{\star}_{Q, \dset{z}}) & = & \infty,
\end{IEEEeqnarray}
which implies that,
\begin{IEEEeqnarray}{rCl}
\label{EqProofLambdIIIAInfty}
\int \frac{1}{\foo{L}_{\dset{z}}(\thetav) + \beta}\diff Q(\thetav) & = & \infty.
\end{IEEEeqnarray}
The equality in~\eqref{EqProofLambdIIIAInfty} implies that the constraint~\eqref{EqType2KrescConstrain2} is not satisfied.
Therefore, for case $(a)$ it follows that,
\begin{IEEEeqnarray}{rCl}
\label{EqCase3aNotInCQz}
-\delta^\star_{Q, \dset{z}} & \notin & \set{C}_{Q, \dset{z}}.
\end{IEEEeqnarray}
In the alternative case $(b)$, if $\beta = -\delta^\star_{Q, \dset{z}}$ and $Q(\set{L}^{\star}_{Q, \dset{z}})=0$, then, the integral in~\eqref{Eq_InvKbarEq} is either 
%
\begin{IEEEeqnarray}{rCl}
\label{EqProofT2lowlimDeltaIsLessInfty}
	\int \frac{1}{\foo{L}_{\dset{z}}(\thetav) - \delta^\star_{Q, \dset{z}}}\diff Q(\thetav)
	& < &  \infty,
\end{IEEEeqnarray}
which implies that $-\delta^{\star}_{Q, \dset{z}} \in \set{C}_{Q, \dset{z}}$, with $\set{C}_{Q, \dset{z}}$ defined in~\eqref{EqDefMapNormFunction}, or the integral~is
\begin{IEEEeqnarray}{rCl}
\label{EqProofT2lowlimDeltaIsEqualInfty}
	\int \frac{1}{\foo{L}_{\dset{z}}(\thetav) - \delta^\star_{Q, \dset{z}}}\diff Q(\thetav)
	& = & \infty,
	\end{IEEEeqnarray}
	  which implies that $-\delta^{\star}_{Q, \dset{z}} \notin \set{C}_{Q, \dset{z}}$.
Hence, from~\eqref{EqCase1NotInCQz},~\eqref{EqCase2InCQz},~\eqref{EqCase3aNotInCQz},~\eqref{EqProofT2lowlimDeltaIsLessInfty}, and~\eqref{EqProofT2lowlimDeltaIsEqualInfty} the set $\set{C}_{Q, \dset{z}}$ in~\eqref{EqDefMapNormFunction} is either the open set $(-\delta^{\star}_{Q, \dset{z}}, \infty)$ or the closed set $[-\delta^{\star}_{Q, \dset{z}}, \infty)$. Note that the equality $\set{C}_{Q, \dset{z}} = [-\delta^{\star}_{Q, \dset{z}}, \infty)$ is observed, if and only if, 
\begin{IEEEeqnarray}{rCl}
	\int \frac{1}{\foo{L}_{\dset{z}}(\thetav) - \delta^\star_{Q, \dset{z}}}\diff Q(\thetav)
	& < &  \infty,
\end{IEEEeqnarray}
which completes the first part of the proof.

The second part of the proof is as follows. Two cases are considered: $i\bigl)$ $\set{C}_{Q, \dset{z}} = [-\delta^{\star}_{Q, \dset{z}} , \infty)$; and 
$ii\bigl)$ $\set{C}_{Q, \dset{z}} = (-\delta^{\star}_{Q, \dset{z}} , \infty)$.
In case $i\bigl)$, the value $-\delta^{\star}_{Q, \dset{z}}$ is in the domain of the function $\bar{K}^{-1}_{Q, \dset{z}}$ in~\eqref{Eq_InvKbarEq}, that is, the set $\set{C}_{Q, \dset{z}}$. Given that the function $\bar{K}^{-1}_{Q, \dset{z}}$ is strictly increasing, then, $-\delta^{\star}_{Q, \dset{z}}$ should be mapped to the smallest value in the range of $\bar{K}^{-1}_{Q, \dset{z}}$, denoted by $\lambda_{Q, \dset{z}}$. Hence, 
%
\begin{IEEEeqnarray}{rCl}
	\bar{K}_{Q, \dset{z}}^{-1}(-\delta^{\star}_{Q, \dset{z}}) 
	& = & \lambda_{Q, \dset{z}}\\
	\label{EqSep17_1418}
	& > & 0,
\end{IEEEeqnarray}
where~\eqref{EqSep17_1418} follows from the fact that zero is not in the domain of the function $\bar{K}_{Q, \dset{z}}$, that is, the set $\set{A}_{Q, \dset{z}}$.
 Using these elements, it is concluded that the set $\set{A}_{Q, \dset{z}}$ is the interval  $[\lambda_{Q, \dset{z}} , \infty)$, which ends the analysis of case~$i.\big)$.

In case $ii\bigl)$, from Lemma~\ref{lemm_InfDevKtype2}, the continuity and strict monotonicity of the function $\bar{K}_{Q, \dset{z}}$ in~\eqref{EqDefNormFunction} imply that $\set{A}_{Q, \dset{z}} = (\lambda^{\star}_{Q, \dset{z}}, \infty)$, with $\lambda^{\star}_{Q, \dset{z}}$ in~\eqref{EqDefLambdaStar}. The remaining of the proof focuses on showing that $\lambda^{\star}_{Q, \dset{z}} = 0$ in this case, and thus, $\set{A}_{Q, \dset{z}} = (0, \infty)$.
From Lemma~\ref{lemm_InfDevKtype2} and the continuous inverse theorem \cite[Theorem 5.6.5]{bartle2000introduction}, it follows that function $\bar{K}^{-1}_{Q, \dset{z}}$ is strictly increasing and continuous. Hence, using~\eqref{Eq_InvKbarEq}, it holds that
%
\begin{IEEEeqnarray}{rCl}
\label{EqProofLimInvKbarOpen_s1}
	\lim_{\gamma \to {-\delta^{\star}_{Q,\dset{z}}}^{+}} \bar{K}^{-1}_{Q, \dset{z}}(\gamma)
	& = & \lim_{\gamma \to {-\delta^{\star}_{Q,\dset{z}}}^{+}} \frac{1}{\int \frac{1}{\foo{L}_{\dset{z}}(\thetav) + \gamma}\diff Q(\thetav)}\\
	\label{EqProofLimInvKbarOpen_s2}
	& = & \frac{1}{\displaystyle{\lim_{\gamma \to {-\delta^{\star}_{Q,\dset{z}}}^{+}}\int \frac{1}{\foo{L}_{\dset{z}}(\thetav) + \gamma}\diff Q(\thetav)}}\qquad\\
	\label{EqProofLimInvKbarOpen_s3}
	& = & \frac{1}{\int \frac{1}{\foo{L}_{\dset{z}}(\thetav) -\delta^{\star}_{Q,\dset{z}}}\diff Q(\thetav)}\\
	\label{EqProofLimInvKbarOpen_s4}
	& = & 0,
\end{IEEEeqnarray}
%
where~\eqref{EqProofLimInvKbarOpen_s2} follows from \cite[Theorem 4.4]{rudin1976bookPrinciples}, that permits the change of the limit to the reciprocal; and~\eqref{EqProofLimInvKbarOpen_s3} follows from~\eqref{EqSuchACrazyDay2Lem5}; and~\eqref{EqProofLimInvKbarOpen_s4} follows from~\eqref{EqProofT2lowlimDeltaIsEqualInfty}.
From Lemma~\ref{lemm_InfDevKtype2} and~\eqref{EqProofLimInvKbarOpen_s4}, it follows that in this second case,  in which  $\set{C}_{Q, \dset{z}} = (-\delta^{\star}_{Q, \dset{z}} , \infty)$, it holds that $\set{A}_{Q, \dset{z}} = (0 , \infty)$.
 This completes the proof. \proofIEEEend

\section{Proof of Lemma~\ref{lemm_Type2_KbarLambda}}
\label{app_proof_lemm_Type2_KbarLambda}
%
From Lemma~\ref{lemm_InfDevKtype2}, the function $\bar{K}_{Q, \dset{z}}$ in~\eqref{EqDefNormFunction} is strictly increasing and continuous. Additionally, from Lemma~\ref{lemm_Type2_kset}, the domain and range of the function $\bar{K}_{Q, \dset{z}}$, defined by the sets $\set{A}_{Q, \dset{z}}$ and $\set{C}_{Q, \dset{z}}$, respectively, are convex intervals. Consequently, combining Lemma~\ref{lemm_InfDevKtype2} and Lemma~\ref{lemm_Type2_kset}, it follows that 
\begin{IEEEeqnarray}{rCl}
	\lim_{\lambda \rightarrow {\lambda^{\star}_{Q, \dset{z}}}^+} \bar{K}_{Q, \dset{z}}(\lambda) & = & - \delta^{\star}_{Q, \dset{z}},
\end{IEEEeqnarray}
with $\delta^{\star}_{Q, \dset{z}}$ defined in~\eqref{EqDefDeltaStar} and $\lambda^{\star}_{Q, \dset{z}}$ defined in~\eqref{EqDefLambdaStar}.\proofIEEEend

\section{Proof of Lemma~\ref{lemm_ERM_RER_Type2_ineq}}
\label{app_lemm_ERM_RER_Type2_ineq}
%

%
For all~$(\thetav_1, \thetav_2) \in \left( \supp Q \right)^2$, such that
\begin{equation}
\label{EqApril21at20h43in2025atHome}
\foo{L}_{\dset{z}}(\thetav_1)\leq \foo{L}_{\dset{z}}(\thetav_2),
\end{equation}
it follows that 
\begin{subequations}
\begin{IEEEeqnarray}{rCl}
	\frac{\lambda}{\bar{K}_{Q, \dset{z}}(\lambda) + \foo{L}_{\dset{z}}(\thetav_1)}
	&\geq & \frac{\lambda}{\bar{K}_{Q, \dset{z}}(\lambda) + \foo{L}_{\dset{z}}(\thetav_2)},
\end{IEEEeqnarray}
\end{subequations}
where the function $\bar{K}_{Q, \dset{z}}$ is defined in~\eqref{EqDefNormFunction}; and 
 equality holds if and only if \eqref{EqApril21at20h43in2025atHome} holds with equality. 
The proof is completed by noticing that from \eqref{EqGenpdfType2WithK}, the inequality above can be rewritten as
\begin{equation}
	\frac{\diff \Pgibbs{\bar{P}}{Q}}{\diff Q}(\thetav_1) 
	\geq \frac{\diff \Pgibbs{\bar{P}}{Q}}{\diff Q}(\thetav_2),
\end{equation}
which completes the proof.\proofIEEEend
\section{Proof of Lemma~\ref{lemm_ERM_RER_Type2_RNdBounds}} 
\label{app_lemm_ERM_RER_Type2_RNdBounds}

%
From Lemma~\ref{lemm_ERM_RER_Type2_ineq}, it follows that for all $\thetav \in \supp Q$, and for all $\phiv \in \set{L}^{\star}_{Q, \dset{z}} \cap \supp Q$, it holds that
\begin{subequations}
\begin{align}\label{Eq_lem_Type2RNdBounds1}
	\frac{\diff \Pgibbs{\bar{P}}{Q}}{\diff Q}(\thetav) \leq & \frac{\diff \Pgibbs{\bar{P}}{Q}}{\diff Q}(\phiv)\\
	& =  \frac{\lambda}{\foo{L}_{\dset{z}}(\phiv) + \bar{K}_{Q, \dset{z}}(\lambda)}
	\label{Eq_lem_Type2RNdBounds2}\\
	& \leqslant  \frac{\lambda}{\delta^\star_{Q, \dset{z}} + \bar{K}_{Q, \dset{z}}(\lambda)}
	\label{Eq_lem_Type2RNdBounds3}\\
	& <  \infty, \label{Eq_lem_Type2RNdBounds4}
\end{align}
\end{subequations}
where~\eqref{Eq_lem_Type2RNdBounds2} follows from~\eqref{EqGenpdfType2WithK}; 
the equality in~\eqref{Eq_lem_Type2RNdBounds3} follows from the fact that $\foo{L}_{\dset{z}}(\phiv) \geq \delta^\star_{Q, \dset{z}}$;
and~\eqref{Eq_lem_Type2RNdBounds4} follows from the fact that $\bar{K}_{Q, \dset{z}}(\lambda) < \infty$. 
Note that  equalities in~\eqref{Eq_lem_Type2RNdBounds1} and \eqref{Eq_lem_Type2RNdBounds3} hold if and only if $\thetav \in \set{L}^{\star}_{Q, \dset{z}} \cap \supp Q$ (Lemma~\ref{lemm_Type2_KbarLambda}). 
This completes the proof of finiteness.

For the proof of positivity, observe that from Lemma~\ref{lemm_Type2_kset}, it holds that 
\begin{equation}
	-\delta^\star_{Q, \dset{z}} < \bar{K}_{Q, \dset{z}}(\lambda) < \infty,
\end{equation}
which implies for all $\vect{\theta} \in \supp Q$,
\begin{IEEEeqnarray}{rCl}
	0 & < & \delta^\star_{Q, \dset{z}} + \bar{K}_{Q, \dset{z}}(\lambda)\\
\label{Eq_ProofT2IneqRNderivbound}
	& \leqslant & \foo{L}_{\dset{z}}(\thetav) + \bar{K}_{Q, \dset{z}}(\lambda).
\end{IEEEeqnarray}
Hence, from the fact that $\lambda >0$, it holds from~\eqref{EqGenpdfType2WithK} and~\eqref{Eq_ProofT2IneqRNderivbound} that
\begin{subequations}
\begin{align}
	\frac{\diff \Pgibbs{\bar{P}}{Q}}{\diff Q}(\thetav)  > &\> 0, 
\end{align}
\end{subequations}
%
%
which completes the proof. \proofIEEEend
%
\section{Proof of Lemma~\ref{lemm_T2AsymptLamb2Inf}}
\label{App_lemm_T2AsymptLamb2Inf}
%

From Theorem~$1$, the Radon-Nikodym derivative of the measure $\Pgibbs{\bar{P}}{Q}$ with respect to $Q$, satisfies for all $\thetav \in \supp Q$,
	\begin{subequations}
		\label{EqProofT2ExpEquivalence}
		\begin{IEEEeqnarray}{rCl}
		\IEEEeqnarraymulticol{3}{l}{
			 \frac{\diff \Pgibbs{\bar{P}}{Q}}{\diff Q}(\thetav)
		}\nonumber \\ \quad
		& = & \frac{\lambda}{\beta + \foo{L}_{\dset{z}}(\thetav)}\\
\label{EqApril23at6h50in2024GoingToSophiaByBus}		
		& = & \frac{1}{\beta + \foo{L}_{\dset{z}}(\thetav)}\frac{1}{\int \frac{1}{\beta + \foo{L}_{\dset{z}}(\nuv)}\diff Q(\nuv)}\\
		& = & \frac{1}{\int \frac{\beta}{\beta + \foo{L}_{\dset{z}}(\nuv)} \diff Q(\nuv)+\int \frac{\foo{L}_{\dset{z}}(\thetav)}{\beta + \foo{L}_{\dset{z}}(\nuv)} \diff Q(\nuv)},
		\end{IEEEeqnarray}
	\end{subequations}
where \eqref{EqApril23at6h50in2024GoingToSophiaByBus} follows from \eqref{Eq_InvKbarEq}, which implies that
\begin{IEEEeqnarray}{rCl}
\lambda & = & \bar{K}^{-1}_{Q, \dset{z}}(\beta) 
	 =  \frac{1}{\int \frac{1}{\foo{L}_{\dset{z}}(\thetav) + \beta}\diff Q(\thetav)},
\end{IEEEeqnarray}
with the function $\bar{K}^{-1}_{Q, \dset{z}}$ being the inverse of the function $\bar{K}_{Q, \dset{z}}$ in \eqref{EqDefNormFunction}.
Using the function $\bar{K}_{Q, \dset{z}}$, the equation in~\eqref{EqProofT2ExpEquivalence} can be written in term of $\lambda$ such that 
	\begin{IEEEeqnarray}{rCl}
	\IEEEeqnarraymulticol{3}{l}{
		\!\!\!\!\!\!\! \frac{\diff \Pgibbs{\bar{P}}{Q}}{\diff Q}(\thetav)
		}\nonumber \\
		\label{EqProofT2ExpEquivalenceKbar}
		\!\!\!\!\!\!\!& = & \frac{1}{\int \! \frac{\bar{K}_{Q, \dset{z}}(\lambda)}{\bar{K}_{Q, \dset{z}}(\lambda) + \foo{L}_{\dset{z}}(\nuv)} \diff Q(\nuv)
		\!+\! \int \!\frac{\foo{L}_{\dset{z}}(\thetav)}{\bar{K}_{Q, \dset{z}}(\lambda) + \foo{L}_{\dset{z}}(\nuv)} \diff Q(\nuv)}.
		\end{IEEEeqnarray}
	Furthermore, using Lemma~\ref{lemm_InfDevKtype2} and Lemma~\ref{lemm_Type2_kset}, the following holds from \eqref{EqProofT2ExpEquivalenceKbar},
	\begin{subequations}
	\label{EqPfLimLamInfPre}
	\begin{IEEEeqnarray}{rCl}
		\IEEEeqnarraymulticol{3}{l}{%
		\lim_{\lambda \rightarrow \infty} \frac{\diff \Pgibbs{\bar{P}}{Q}}{\diff Q}(\thetav)
		}\nonumber\\ 
		& = & \!\! \lim_{\lambda\rightarrow \infty} \! \frac{1}{\!\!\int \!\!\frac{\bar{K}_{Q, \dset{z}}(\lambda)}{\bar{K}_{Q, \dset{z}}(\lambda)\! + \foo{L}_{\dset{z}}(\nuv)}\! \diff Q(\nuv)
		\!+\! \int \! \frac{\foo{L}_{\dset{z}}(\thetav)}{\bar{K}_{Q, \dset{z}}(\lambda)\! + \foo{L}_{\dset{z}}(\nuv)} \diff Q(\nuv)}\qquad \quad \\
		& = & \!\!\frac{1}{\!\lim_{\beta\rightarrow \infty}\! \int \! \frac{\beta}{\beta \! + \foo{L}_{\dset{z}}(\nuv)}\! \diff Q(\nuv)\!+\! \lim_{\beta\rightarrow \infty}\! \int \! \frac{\foo{L}_{\dset{z}}(\thetav)}{\beta \! + \foo{L}_{\dset{z}}(\nuv)}\! \diff Q(\nuv)},\quad \label{EqPfLimLamInfPre_s2}
	\end{IEEEeqnarray}
	\end{subequations}
	where the function $\foo{L}_{\dset{z}}$ is defined in $(3)$; and~\eqref{EqPfLimLamInfPre_s2} follows from Theorem~\ref{Theo_ERMType2RadNikMutualAbs}, which implies that the terms in the denominator are positive and the fact that the function $g(x) = \frac{1}{x}$ is continuous.  Recall that from the definition of the function $\foo{L}_{\dset{z}}$ in~\eqref{EqLxy}, for all $\thetav \in \supp Q$, the empirical risk satisfies that $\foo{L}_{\dset{z}}(\thetav) < \infty$. Using this fact, the proof continues by evaluating the limits in the denominator, which yields
	\begin{subequations}
	\label{EqPfLimLamInfDen1}
	\begin{IEEEeqnarray}{rCl}
		\IEEEeqnarraymulticol{3}{l}{%
		\lim_{\beta\rightarrow \infty} \int \frac{\beta}{\beta + \foo{L}_{\dset{z}}(\nuv)} \diff Q(\nuv) 
		}\nonumber \\
		& = & \int \lim_{\beta\rightarrow \infty} \frac{\beta}{\beta + \foo{L}_{\dset{z}}(\nuv)} \diff Q(\nuv)\label{EqPfLimLamInfDen1_s1}\\
		& = & \int \diff Q(\nuv)\label{EqPfLimLamInfDen1_s2}\\
		& = & 1,\label{EqPfLimLamInfDen1_s3}
	\end{IEEEeqnarray}
	\end{subequations}
	where~\eqref{EqPfLimLamInfDen1_s1} follows from the dominated convergence theorem \cite[Theorem 1.6.9]{ash2000probability};
	and,
	\begin{subequations}
	\label{EqPfLimLamInfDen2}
	\begin{IEEEeqnarray}{rCl}
	\IEEEeqnarraymulticol{3}{l}{%
		\lim_{\beta\rightarrow \infty} \int \frac{\foo{L}_{\dset{z}}(\thetav)}{\beta + \foo{L}_{\dset{z}}(\nuv)} \diff Q(\nuv) 
		}\nonumber \\
		& = & \int \lim_{\beta\rightarrow \infty} \frac{\foo{L}_{\dset{z}}(\thetav)}{\beta + \foo{L}_{\dset{z}}(\nuv)} \diff Q(\nuv)\label{EqPfLimLamInfDen2_s1}\\
		& = &\int 0 \diff Q(\nuv)\label{EqPfLimLamInfDen2_s2}\\
		& = & 0,\label{EqPfLimLamInfDen2_s3}
	\end{IEEEeqnarray}
	\end{subequations}
	where~\eqref{EqPfLimLamInfDen2_s1} also follows from the dominated convergence theorem \cite[Theorem 1.6.9]{ash2000probability}.
	Substituting~\eqref{EqPfLimLamInfDen1} and~\eqref{EqPfLimLamInfDen2} into~\eqref{EqPfLimLamInfPre} yields
	\begin{IEEEeqnarray}{rCl}
		\lim_{\lambda \rightarrow \infty} \frac{\diff \Pgibbs{\bar{P}}{Q}}{\diff Q}(\thetav)
		& = & 1,
	\end{IEEEeqnarray}
	 which completes the proof. \proofIEEEend
\section{Proof of Lemma~\ref{lemm_T2AsymptLamb2Zero}}
\label{AppProof_lemm_RadNikDevType2Exp}
%
\label{App_lemm_T2AsymptLamb2Zero}
From Theorem~\ref{Theo_ERMType2RadNikMutualAbs}, the probability measure $\Pgibbs{\bar{P}}{Q}$ satisfies for all $\thetav \in \supp Q$,
\begin{subequations}
\label{EqProofT2RNdervKbar}
\begin{IEEEeqnarray}{rCl}
\label{EqProofT2RNdervKbar_s1}
	\frac{\diff \Pgibbs{\bar{P}}{Q}}{\diff Q}(\thetav)
 	& = & \frac{\lambda}{\foo{L}_{\dset{z}}(\thetav) + \beta}\\
 	& = & \frac{\lambda}{\foo{L}_{\dset{z}}(\thetav) + \bar{K}_{Q, \dset{z}}(\lambda)}\label{EqProofT2RNdervKbar_s2}\\
 	& = & \frac{\bar{K}^{-1}_{Q, \dset{z}}(\bar{K}_{Q, \dset{z}}(\lambda))}{\foo{L}_{\dset{z}}(\thetav) + \bar{K}_{Q, \dset{z}}(\lambda)}\label{EqProofT2RNdervKbar_s3}\\
 	& = & \frac{\frac{1}{\int \frac{1}{\foo{L}_{\dset{z}}(\nuv) + \bar{K}_{Q, \dset{z}}(\lambda)}\diff Q(\nuv)}}{\foo{L}_{\dset{z}}(\thetav) + \bar{K}_{Q, \dset{z}}(\lambda)}\label{EqProofT2RNdervKbar_s4}\\
 	& = & (\int \frac{\foo{L}_{\dset{z}}(\thetav) + \bar{K}_{Q, \dset{z}}(\lambda)}{\foo{L}_{\dset{z}}(\nuv) + \bar{K}_{Q, \dset{z}}(\lambda)}\diff Q(\nuv) )^{-1},\qquad
\end{IEEEeqnarray}
\end{subequations}
where~\eqref{EqProofT2RNdervKbar_s2} follows from~\eqref{EqDefNormFunction}; and~\eqref{EqProofT2RNdervKbar_s4} follows from~\eqref{Eq_InvKbarEq} and observing that $\lambda = \bar{K}^{-1}_{Q, \dset{z}}(\beta)$.
Given $\thetav \in \supp Q$, consider the partition of the $\supp Q$ formed by the sets $\set{A}_0(\thetav)$, $\set{A}_1(\thetav)$, and $\set{A}_2(\thetav)$, which satisfy the following:
\begin{subequations}
\label{EqProofSetsPartitionfbar37}
\begin{IEEEeqnarray}{rCl}
\label{EqProofSetsPartitionfbar37_s1}
\set{A}_0(\thetav) & = &\{ \nuv \in \supp Q: \foo{L}_{\dset{z}}(\thetav) -\foo{L}_{\dset{z}}(\nuv) = 0\}, \\
\set{A}_1(\thetav) & = &\{ \nuv \in \supp Q: \foo{L}_{\dset{z}}(\thetav) -\foo{L}_{\dset{z}}(\nuv) < 0\}, \ \text{and} \qquad
\label{EqProofSetsPartitionfbar37_s2}\\
\set{A}_2(\thetav) & = &\{ \nuv \in \supp Q: \foo{L}_{\dset{z}}(\thetav) -\foo{L}_{\dset{z}}(\nuv) > 0\}.
\label{EqProofSetsPartitionfbar37_s3}
\end{IEEEeqnarray}
\end{subequations}
Using the sets $\set{A}_0(\thetav)$, $\set{A}_1(\thetav)$, and $\set{A}_2(\thetav)$ in~\eqref{EqProofSetsPartitionfbar37}, the following holds for all $\thetav \in \supp Q$.
\begin{subequations}
\label{EqProofT2RNdervKbarA012}
\begin{IEEEeqnarray}{rCl}
	\IEEEeqnarraymulticol{3}{l}{%
	\frac{\diff \Pgibbs{\bar{P}}{Q}}{\diff Q}(\thetav)
	}\nonumber\\
 	& = &  (\int_{\set{A}_0(\thetav)} \frac{\foo{L}_{\dset{z}}(\thetav) + \bar{K}_{Q, \dset{z}}(\lambda)}{\foo{L}_{\dset{z}}(\nuv) + \bar{K}_{Q, \dset{z}}(\lambda)}\diff Q(\nuv) 
 	\right. \nonumber \\ &   &
 	+ \int_{\set{A}_1(\thetav)} \frac{\foo{L}_{\dset{z}}(\thetav) + \bar{K}_{Q, \dset{z}}(\lambda)}{\foo{L}_{\dset{z}}(\nuv) + \bar{K}_{Q, \dset{z}}(\lambda)}\diff Q(\nuv) 
 	\nonumber \\ &   & \left.
 	+ \int_{\set{A}_2(\thetav)} \frac{\foo{L}_{\dset{z}}(\thetav) + \bar{K}_{Q, \dset{z}}(\lambda)}{\foo{L}_{\dset{z}}(\nuv) + \bar{K}_{Q, \dset{z}}(\lambda)}\diff Q(\nuv) )^{-1}\\
 	& = &  (Q(\set{A}_0(\thetav)) + \int_{\set{A}_1(\thetav)} \frac{\foo{L}_{\dset{z}}(\thetav) + \bar{K}_{Q, \dset{z}}(\lambda)}{\foo{L}_{\dset{z}}(\nuv) + \bar{K}_{Q, \dset{z}}(\lambda)}\diff Q(\nuv) 
 	\right.\nonumber\\ &   & \left.
 	+ \int_{\set{A}_2(\thetav)} \frac{\foo{L}_{\dset{z}}(\thetav) + \bar{K}_{Q, \dset{z}}(\lambda)}{\foo{L}_{\dset{z}}(\nuv) + \bar{K}_{Q, \dset{z}}(\lambda)}\diff Q(\nuv) )^{-1}.
\end{IEEEeqnarray}
\end{subequations}

Consider the following partition of the $\supp Q$:
\begin{subequations}
\label{EqProofSetsPartitionfbar36}
\begin{IEEEeqnarray}{Cl}
\label{EqProofSetsPartitionfbar36_s1}
\{ \nuv \in \supp Q: \foo{L}_{\dset{z}}(\nuv) = \delta^\star_{Q, \dset{z}}\},& \\
\{ \nuv \in \supp Q: \foo{L}_{\dset{z}}(\nuv) > \delta^\star_{Q, \dset{z}}\},& \ \text{and}
\label{EqProofSetsPartitionfbar36_s2}\\
\{ \nuv \in \supp Q: \foo{L}_{\dset{z}}(\nuv) < \delta^\star_{Q, \dset{z}}\},&
\label{EqProofSetsPartitionfbar36_s3}
\end{IEEEeqnarray}
\end{subequations}
with $\delta^\star_{Q, \dset{z}}$ in~\eqref{EqDefDeltaStar}.
The proof is divided into two cases. The first case follows under the assumption that 
\begin{IEEEeqnarray}{rCl}
\label{EqProofCase1Condition}
	\int \frac{1}{\foo{L}_{\dset{z}}(\thetav)-\delta^{\star}_{Q, \dset{z}}} \diff Q(\thetav) & = & \infty;
\end{IEEEeqnarray}
and the second case follows under the assumption that
\begin{IEEEeqnarray}{rCl}
\label{EqProofCase2Condition}
	\int \frac{1}{\foo{L}_{\dset{z}}(\thetav)-\delta^{\star}_{Q, \dset{z}}} \diff Q(\thetav) & < & \infty.
\end{IEEEeqnarray}
From Lemma~\ref{lemm_Type2_kset}, it follows that in Case $1$, the set  $\set{A}_{Q, \dset{z}}$ in~\eqref{EqDefNormFunction} is $(0,\infty)$. Similarly, in Case $2$, the set $\set{A}_{Q, \dset{z}}$ is $ [\lambda^{\star}_{Q,\dset{z}},\infty)$. Hence,  Case $1$ considers the limit  ${\lambda \to 0^{+}}$, which comprehends the equalities~\eqref{EqLemm14_s1} and~\eqref{EqLemm14_s2}. Case $2$ considers the limit ${\lambda \to {\lambda^{\star}_{Q,\dset{z}}}^{+}}$, which comprehends the equality~\eqref{EqLemm14_s3}.

\subsection{Case $1$}
This case is divided into three parts.
The first part evaluates $\lim_{\lambda \rightarrow 0^{+}} \frac{\diff \Pgibbs{\bar{P}}{Q}}{\diff Q} (\thetav)$, with $\thetav \in \{ \nuv \in \supp Q: \foo{L}_{\dset{z}}(\nuv) = \delta^\star_{Q, \dset{z}}\}$. The second part considers the case in which $\thetav \in \{ \nuv \in \supp Q: \foo{L}_{\dset{z}}(\nuv) > \delta^\star_{Q, \dset{z}}\}$. The third part considers the remaining case in~\eqref{EqProofSetsPartitionfbar36}.

\subsubsection{Part $1$}
The first part is as follows. Consider that $\thetav \in \{ \nuv \in \supp Q: \foo{L}_{\dset{z}}(\nuv) = \delta^\star_{Q, \dset{z}}\}$ and note that
\begin{IEEEeqnarray}{rCl}
	 \{ \nuv \in \supp Q: \foo{L}_{\dset{z}}(\nuv)  =  \delta^\star_{Q, \dset{z}}\} & = & \set{L}^{\star}_{Q, \dset{z}},
\end{IEEEeqnarray}
with $\set{L}^{\star}_{Q, \dset{z}}$ defined in~\eqref{EqDefSetLStarQz}. Hence, the sets $\set{A}_0(\thetav)$, $\set{A}_1(\thetav)$, and $\set{A}_2(\thetav)$ in~\eqref{EqProofSetsPartitionfbar37} satisfy the following: 
\begin{subequations}
\label{EqProofSetsPartitionfbar35}
\begin{IEEEeqnarray}{rCl}
\label{EqProofSetsPartitionfbar35_s1}
\set{A}_0(\thetav)& = & \set{L}^{\star}_{Q, \dset{z}},\\
\set{A}_1(\thetav)& = &\{ \muv \in \supp Q: \foo{L}_{\dset{z}}(\muv) > \delta^\star_{Q, \dset{z}}\}, \ \text{and}
\label{EqProofSetsPartitionfbar35_s2}\\
\set{A}_2(\thetav)& = &\{ \muv \in \supp Q: \foo{L}_{\dset{z}}(\muv) < \delta^\star_{Q, \dset{z}}\}.
\label{EqProofSetsPartitionfbar35_s3}
\end{IEEEeqnarray}
\end{subequations}
From the definition of $\delta^\star_{Q, \dset{z}}$ in~\eqref{EqDefDeltaStar}, it follows that $Q(\set{A}_2(\thetav))=0$. Substituting the equalities in~\eqref{EqProofSetsPartitionfbar35} in~\eqref{EqProofT2RNdervKbarA012} yields for all $\thetav \in \{ \nuv \in \supp Q: \foo{L}_{\dset{z}}(\nuv) = \delta^\star_{Q, \dset{z}}\}$, 
\begin{IEEEeqnarray}{rCl}
	\IEEEeqnarraymulticol{3}{l}{%
	\frac{\diff \Pgibbs{\bar{P}}{Q}}{\diff Q}(\thetav)
	}\nonumber \\ \label{EqProofT2RNdervKbarA012delta}
 	& = &  (Q(\set{L}^{\star}_{Q, \dset{z}}) + \int_{\set{A}_1(\thetav)} \frac{\foo{L}_{\dset{z}}(\thetav) + \bar{K}_{Q, \dset{z}}(\lambda)}{\foo{L}_{\dset{z}}(\nuv) + \bar{K}_{Q, \dset{z}}(\lambda)}\diff Q(\nuv) )^{-1}\!\!\!\!\!, \qquad
\end{IEEEeqnarray}
which implies that for all $\thetav \in \{ \nuv \in \supp Q: \foo{L}_{\dset{z}}(\nuv) = \delta^\star_{Q, \dset{z}}\}$,
\begin{IEEEeqnarray}{rCl}
	\IEEEeqnarraymulticol{3}{l}{%
	\lim_{\lambda \rightarrow 0^{+}}\frac{\diff \Pgibbs{\bar{P}}{Q}}{\diff Q}(\thetav)
	}\nonumber\\ \label{EqProofT2RNdervKbarLimA012delta}
 	& = & \! (\!Q(\set{L}^{\star}_{Q, \dset{z}}) \! + \! \lim_{\lambda \rightarrow 0^{+}} \int_{\set{A}_1(\thetav)}\! \frac{\foo{L}_{\dset{z}}(\thetav) + \bar{K}_{Q, \dset{z}}(\lambda)}{\foo{L}_{\dset{z}}(\nuv) + \bar{K}_{Q, \dset{z}}(\lambda)}\diff Q(\nuv)\! )^{-1}\quad\\
 	& = & \begin{cases}
			\infty & \text{if } Q(\set{L}^\star_{Q, \dset{z}}) = 0 \\
			\frac{1}{Q(\set{L}^\star_{Q, \dset{z}})}  & \text{otherwise} 
		\end{cases},\label{EqProofT2RNdervKbarLimA012delta_s2}
\end{IEEEeqnarray}
where~\eqref{EqProofT2RNdervKbarLimA012delta_s2} follows from verifying that the dominated convergence theorem \cite[Theorem 2.6.9]{ash2000probability} holds. That is,\\
$(a)$ For all $\nuv \in \set{A}_1(\thetav)$, it holds that $\frac{\foo{L}_{\dset{z}}(\thetav) + \bar{K}_{Q, \dset{z}}(\lambda)}{\foo{L}_{\dset{z}}(\nuv) + \bar{K}_{Q, \dset{z}}(\lambda)} \leq \frac{\lambda}{\delta^\star_{Q, \dset{z}} + \bar{K}_{Q, \dset{z}}(\lambda)}$.\\
$(b)$ For all $\nuv \in \set{A}_1(\thetav)$, it holds that
\begin{subequations}
\begin{IEEEeqnarray}{rCl}
\IEEEeqnarraymulticol{3}{l}{%
	\lim_{\lambda \rightarrow 0^{+}} \frac{\foo{L}_{\dset{z}}(\thetav) + \bar{K}_{Q, \dset{z}}(\lambda)}{\foo{L}_{\dset{z}}(\nuv) + \bar{K}_{Q, \dset{z}}(\lambda)} 
	}\nonumber \\
	& = & \lim_{\lambda \rightarrow 0^{+}} \frac{\delta^\star_{Q, \dset{z}} + \bar{K}_{Q, \dset{z}}(\lambda)}{\foo{L}_{\dset{z}}(\nuv) + \bar{K}_{Q, \dset{z}}(\lambda)}\label{EqLimZeroLambA0_s1}\\
	& = & (\delta^\star_{Q, \dset{z}} + \lim_{\lambda \rightarrow 0^{+}}\bar{K}_{Q, \dset{z}}(\lambda))\lim_{\lambda \rightarrow 0^{+}} \frac{1}{\foo{L}_{\dset{z}}(\nuv) + \bar{K}_{Q, \dset{z}}(\lambda)} \label{EqLimZeroLambA0_s2}\qquad\\
	& = & 0,\label{EqLimZeroLambA0_s3}
\end{IEEEeqnarray}
\end{subequations}
where~\eqref{EqLimZeroLambA0_s2} follows from observing that for all $\nuv \in \set{A}_{1}(\thetav)$, it holds that $\lim_{\lambda \rightarrow 0^{+}}\foo{L}_{\dset{z}}(\nuv) + \bar{K}_{Q, \dset{z}}(\lambda) \neq 0$  and \cite[Theorem 4.4]{rudin1976bookPrinciples}; and~\eqref{EqLimZeroLambA0_s3} follows from Lemma~\ref{lemm_Type2_KbarLambda}.
This completes the first part of Case $1$.

\subsubsection{Part $2$} For all $\delta > \delta^\star_{Q, \dset{z}}$ and for all $\thetav \in \{ \nuv \in \supp Q: \foo{L}_{\dset{z}}(\nuv) = \delta\}$, the sets $\set{A}_0(\thetav)$, $\set{A}_1(\thetav)$, and $\set{A}_2(\thetav)$ in~\eqref{EqProofSetsPartitionfbar37} satisfy the following:
\begin{subequations}
\label{EqProofSetsPartitionfbar34}
\begin{IEEEeqnarray}{rCl}
\label{EqProofSetsPartitionfbar34_s1}
\set{A}_0(\thetav)& = & \{ \muv \in \supp Q: \foo{L}_{\dset{z}}(\muv) = \delta\},\\
\set{A}_1(\thetav)& = &\{ \muv \in \supp Q: \foo{L}_{\dset{z}}(\muv) > \delta\}, \ \text{and}
\label{EqProofSetsPartitionfbar34_s2}\\
\set{A}_2(\thetav)& = &\{ \muv \in \supp Q: \foo{L}_{\dset{z}}(\muv) < \delta\}.
\label{EqProofSetsPartitionfbar34_s3}
\end{IEEEeqnarray}
\end{subequations}

Consider the sets
\begin{subequations}
\label{EqProofSetsPartitionfbar33}
\begin{IEEEeqnarray}{rCl}
\label{EqProofSetsPartitionfbar33_s1}
\set{A}_{2,1}(\thetav)& = & \{ \muv \in \set{A}_2(\thetav): \foo{L}_{\dset{z}}(\muv) < \delta^\star_{Q, \dset{z}} \},\text{ and}\\
\set{A}_{2,2}(\thetav)& = & \{ \muv \in \set{A}_2(\thetav): \delta^\star_{Q, \dset{z}} \leq \foo{L}_{\dset{z}}(\muv) < \delta\},
\label{EqProofSetsPartitionfbar33_s2}
\end{IEEEeqnarray}
\end{subequations}
and note that $\set{A}_{2,1}(\thetav)$ and $\set{A}_{2,2}(\thetav)$ form a partition of $\set{A}_{2}(\thetav)$. 
 Moreover, from the definition of $\delta^\star_{Q, \dset{z}}$ in~\eqref{EqDefDeltaStar}, it holds that 
\begin{IEEEeqnarray}{rCl}
\label{EqProofFun37}
	Q(\set{A}_{2,1}(\thetav)) & = & 0.
\end{IEEEeqnarray}
Hence, substituting the equalities in~\eqref{EqProofSetsPartitionfbar34} and~\eqref{EqProofFun37} in~\eqref{EqProofT2RNdervKbarA012} yields, 
\begin{subequations}
\label{EqProofT2RNdervKbarLimA1delta}
\begin{IEEEeqnarray}{rCl}
	\IEEEeqnarraymulticol{3}{l}{%
	\lim_{\lambda \rightarrow 0^{+}}\frac{\diff \Pgibbs{\bar{P}}{Q}}{\diff Q}(\thetav)
	}\nonumber\\
 	& = &  (Q(\set{A}_0(\thetav)) + \lim_{\lambda \rightarrow 0^{+}} \int_{\set{A}_1(\thetav)} \frac{\foo{L}_{\dset{z}}(\thetav) + \bar{K}_{Q, \dset{z}}(\lambda)}{\foo{L}_{\dset{z}}(\nuv) + \bar{K}_{Q, \dset{z}}(\lambda)}\diff Q(\nuv) 
 	\right. \nonumber \\ &   & \left.
 	+ \lim_{\lambda \rightarrow 0^{+}}\int_{\set{A}_{2}(\thetav)} \frac{\foo{L}_{\dset{z}}(\thetav) + \bar{K}_{Q, \dset{z}}(\lambda)}{\foo{L}_{\dset{z}}(\nuv) + \bar{K}_{Q, \dset{z}}(\lambda)}\diff Q(\nuv) )^{-1}
 	\label{EqProofT2RNdervKbarLimA1delta_s1}\\
 	& = & (Q(\set{A}_0(\thetav)) + \lim_{\lambda \rightarrow 0^{+}} \int_{\set{A}_1(\thetav)} \frac{\foo{L}_{\dset{z}}(\thetav) + \bar{K}_{Q, \dset{z}}(\lambda)}{\foo{L}_{\dset{z}}(\nuv) + \bar{K}_{Q, \dset{z}}(\lambda)}\diff Q(\nuv) 
 	\right. \nonumber \\ &   & \left.
 	+\lim_{\lambda \rightarrow 0^{+}}\int_{\set{A}_{2,2}(\thetav)} \frac{\foo{L}_{\dset{z}}(\thetav) + \bar{K}_{Q, \dset{z}}(\lambda)}{\foo{L}_{\dset{z}}(\nuv) + \bar{K}_{Q, \dset{z}}(\lambda)}\diff Q(\nuv) )^{-1}\label{EqProofT2RNdervKbarLimA1delta_s2}\\
 	& = & (Q(\set{A}_0(\thetav)) +  \int_{\set{A}_1(\thetav)} \lim_{\lambda \rightarrow 0^{+}}\frac{\foo{L}_{\dset{z}}(\thetav) + \bar{K}_{Q, \dset{z}}(\lambda)}{\foo{L}_{\dset{z}}(\nuv) + \bar{K}_{Q, \dset{z}}(\lambda)}\diff Q(\nuv) 
 	\right. \nonumber \\ &   & \left. 
 	+\int_{\set{A}_{2,2}(\thetav)} \lim_{\lambda \rightarrow 0^{+}} \frac{\foo{L}_{\dset{z}}(\thetav) + \bar{K}_{Q, \dset{z}}(\lambda)}{\foo{L}_{\dset{z}}(\nuv) + \bar{K}_{Q, \dset{z}}(\lambda)}\diff Q(\nuv) )^{-1},\label{EqProofT2RNdervKbarLimA1delta_s3}
\end{IEEEeqnarray}
\end{subequations}
where~\eqref{EqProofT2RNdervKbarLimA1delta_s3} follows by verifying that the dominated convergence theorem \cite[Theorem 1.6.9]{ash2000probability} holds. 
That is,\\
$(a)$ For all $\nuv \in \set{A}_{2,2}(\thetav)$, it holds that $\frac{\foo{L}_{\dset{z}}(\thetav) + \bar{K}_{Q, \dset{z}}(\lambda)}{\foo{L}_{\dset{z}}(\nuv) + \bar{K}_{Q, \dset{z}}(\lambda)} \leq \frac{\lambda}{\delta^\star_{Q, \dset{z}} + \bar{K}_{Q, \dset{z}}(\lambda)}< \infty $; and\\
$(b)$ For all $\nuv \in \set{A}_{2,2}(\thetav)$, it holds that
\begin{subequations}
\begin{IEEEeqnarray}{rCl}
\IEEEeqnarraymulticol{3}{l}{%
	\lim_{\lambda \rightarrow 0^{+}} \frac{\foo{L}_{\dset{z}}(\thetav) + \bar{K}_{Q, \dset{z}}(\lambda)}{\foo{L}_{\dset{z}}(\nuv) + \bar{K}_{Q, \dset{z}}(\lambda)} 
	}\nonumber \\
	& = & \lim_{\lambda \rightarrow 0^{+}} \frac{\delta + \bar{K}_{Q, \dset{z}}(\lambda)}{\foo{L}_{\dset{z}}(\nuv) + \bar{K}_{Q, \dset{z}}(\lambda)}\label{EqLimZeroLambA1_s1}\\
	\label{EqLimZeroLambA1_s2}
	& = & (\delta + \lim_{\lambda \rightarrow 0^{+}} \bar{K}_{Q, \dset{z}}(\lambda))\lim_{\lambda \rightarrow 0^{+}}\frac{1}{\foo{L}_{\dset{z}}(\nuv) + \bar{K}_{Q, \dset{z}}(\lambda)}\quad\\
	\label{EqLimZeroLambA1_s3}
	& = & (\delta -\delta^\star_{Q, \dset{z}})\frac{1}{\foo{L}_{\dset{z}}(\nuv) -\delta^\star_{Q, \dset{z}}},
\end{IEEEeqnarray}
\end{subequations}
where~\eqref{EqLimZeroLambA1_s2} follows from observing that for all $\nuv \in \set{A}_{2,2}(\thetav)$, it holds that $\lim_{\lambda \rightarrow 0^{+}}\foo{L}_{\dset{z}}(\nuv) + \bar{K}_{Q, \dset{z}}(\lambda) \neq 0$  and \cite[Theorem 4.2]{rudin1976bookPrinciples}; and~\eqref{EqLimZeroLambA1_s3} follows from Lemma~\ref{lemm_Type2_KbarLambda}.
From~\eqref{EqLimZeroLambA1_s3}, it follows that
\begin{IEEEeqnarray}{rCl}
\IEEEeqnarraymulticol{3}{l}{%
 \int_{\set{A}_{2,2}(\thetav)} \lim_{\lambda \rightarrow 0^{+}}\frac{\foo{L}_{\dset{z}}(\thetav) + \bar{K}_{Q, \dset{z}}(\lambda)}{\foo{L}_{\dset{z}}(\nuv) + \bar{K}_{Q, \dset{z}}(\lambda)}\diff Q(\nuv)
 }\nonumber \\ \label{EqLimA22InInt}
& = &   (\delta -\delta^\star_{Q, \dset{z}})\int_{\set{A}_{2,2}(\thetav)} \frac{1}{\foo{L}_{\dset{z}}(\nuv) - \delta^{\star}_{Q,\dset{z}}}\diff Q(\nuv).
\end{IEEEeqnarray}
Moreover, from the fact that 
\begin{IEEEeqnarray}{rCl}
 \label{EqProofLimFiniteForA0}
 Q(\set{A}_{0}(\thetav)) \leq 1,
\end{IEEEeqnarray}
and the fact that
\begin{IEEEeqnarray}{rCl}
	\IEEEeqnarraymulticol{3}{l}{%
	\int_{\set{A}_{1}(\thetav)} \lim_{\lambda \rightarrow 0^{+}}\frac{\foo{L}_{\dset{z}}(\thetav) + \bar{K}_{Q, \dset{z}}(\lambda)}{\foo{L}_{\dset{z}}(\nuv) + \bar{K}_{Q, \dset{z}}(\lambda)}\diff Q(\nuv) 
	}\nonumber \\
	& = & \int_{\set{A}_{1}(\thetav)} \frac{\delta -\delta^\star_{Q, \dset{z}}}{\foo{L}_{\dset{z}}(\nuv) - \delta^{\star}_{Q,\dset{z}}} \diff Q(\nuv) \\
	& < & \int_{\set{A}_{1}(\thetav)} \frac{\delta -\delta^\star_{Q, \dset{z}}}{\delta - \delta^{\star}_{Q,\dset{z}}}\diff Q(\nuv)\\
	& = & Q(\set{A}_{1}(\thetav))\\
	 \label{EqProofLimFiniteForA1_s3}
	& \leq & 1,
\end{IEEEeqnarray}
the following holds from Lemma~\ref{lemm_Type2_kset} under the assumptions of Case $1$:
\begin{IEEEeqnarray}{rCl}
\infty & = & (\delta -\delta^\star_{Q, \dset{z}})\int \frac{1}{\foo{L}_{\dset{z}}(\nuv) - \delta^{\star}_{Q,\dset{z}}}\diff Q(\nuv)\\
 & = & (\delta -\delta^\star_{Q, \dset{z}})(\int_{\set{A}_{0}(\thetav)} \frac{1}{\foo{L}_{\dset{z}}(\nuv) - \delta^{\star}_{Q,\dset{z}}}\diff Q(\nuv) 
 \right. \nonumber \\ &  & \left.
 +\int_{\set{A}_{1}(\thetav)} \frac{1}{\foo{L}_{\dset{z}}(\nuv) - \delta^{\star}_{Q,\dset{z}}}\diff Q(\nuv) 
 \right. \nonumber \\ &  & \left.
 +\int_{\set{A}_{2,2}(\thetav)} \frac{1}{\foo{L}_{\dset{z}}(\nuv) - \delta^{\star}_{Q,\dset{z}}}\diff Q(\nuv))\\
 \label{EqProofLimInForA22_s3}
  & = & (Q(\set{A}_{0}(\thetav)) +\int_{\set{A}_{1}(\thetav)} \frac{\delta -\delta^\star_{Q, \dset{z}}}{\foo{L}_{\dset{z}}(\nuv) - \delta^{\star}_{Q,\dset{z}}}\diff Q(\nuv) 
   \right. \nonumber \\ &  & \left.
  +\int_{\set{A}_{2,2}(\thetav)} \frac{\delta -\delta^\star_{Q, \dset{z}}}{\foo{L}_{\dset{z}}(\nuv) - \delta^{\star}_{Q,\dset{z}}}\diff Q(\nuv)).
\end{IEEEeqnarray}
From~\eqref{EqProofLimFiniteForA0},~\eqref{EqProofLimFiniteForA1_s3}, and~\eqref{EqProofLimInForA22_s3}, it follows that 
\begin{IEEEeqnarray}{rCl}
\label{EqNosleep}
   (\delta -\delta^\star_{Q, \dset{z}})\int_{\set{A}_{2,2}(\thetav)} \frac{1}{\foo{L}_{\dset{z}}(\nuv) - \delta^{\star}_{Q,\dset{z}}}\diff Q(\nuv) & = & \infty.
\end{IEEEeqnarray}
Finally, from~\eqref{EqProofT2RNdervKbarLimA1delta_s3},~\eqref{EqLimA22InInt}, and~\eqref{EqNosleep}, for all $\thetav \in \{ \nuv \in \supp Q: \foo{L}_{\dset{z}}(\nuv) > \delta^\star_{Q, \dset{z}}\}$,
\begin{IEEEeqnarray}{rCl}
\label{EqLimZeroLambA1m}
	\lim_{\lambda \rightarrow 0^{+}} \frac{\diff \Pgibbs{\bar{P}}{Q}}{\diff Q}(\thetav) 
	& = & 0.
\end{IEEEeqnarray}
This completes the second part of Case $1$.

\subsubsection{Part $3$}
The third part of the proof follows by noticing that the set $\{ \muv \in \supp Q: \foo{L}_{\dset{z}}(\muv) < \delta^\star_{Q, \dset{z}}\}$ is a negligible set with respect to $Q$ and thus, for all $\thetav  \in \{ \muv \in \supp Q: \foo{L}_{\dset{z}}(\muv) < \delta^\star_{Q, \dset{z}}\}$, the value $\frac{\diff \Pgibbs{\bar{P}}{Q}}{\diff Q}(\thetav)$ is immaterial. Hence, it is arbitrarily assumed that for all $\thetav \in \{ \muv \in \supp Q: \foo{L}_{\dset{z}}(\muv) < \delta^\star_{Q, \dset{z}}\}$, it holds that
\begin{IEEEeqnarray}{rCl}
\frac{\diff \Pgibbs{\bar{P}}{Q}}{\diff Q}(\thetav) & = & 0,	
\end{IEEEeqnarray}
which implies that for all $\thetav \in \{ \muv \in \supp Q: \foo{L}_{\dset{z}}(\muv) < \delta^\star_{Q, \dset{z}}\}$, it holds that 
\begin{IEEEeqnarray}{rCl}
\label{EqProofCs1Prt3Lim}
\lim_{\lambda \rightarrow 0^{+}} \frac{\diff \Pgibbs{\bar{P}}{Q}}{\diff Q}(\thetav) & = & 0.	
\end{IEEEeqnarray}
This completes the third part of Case $1$.

Under the assumption that $Q(\set{L}^\star_{Q, \dset{z}}) > 0$, from~\eqref{EqProofT2RNdervKbarLimA012delta_s2},~\eqref{EqLimZeroLambA1m}, and~\eqref{EqProofCs1Prt3Lim}, for all $\thetav \in \supp Q$, it follows that
	\begin{IEEEeqnarray}{rCl}
		\lim_{\lambda \rightarrow 0^+} \frac{\diff \Pgibbs{\bar{P}}{Q}}{\diff Q}(\thetav) 
		& = & \frac{1}{Q(\set{L}^{\star}_{Q, \dset{z}})}\ind{\thetav \in \set{L}^{\star}_{Q, \dset{z}}},
	\end{IEEEeqnarray}
	which completes the proof of~\eqref{EqLemm14_s1}.
	Alternatively, under the assumption that $Q(\set{L}^\star_{Q, \dset{z}}) = 0$, from~\eqref{EqProofT2RNdervKbarLimA012delta_s2},~\eqref{EqLimZeroLambA1m}, and~\eqref{EqProofCs1Prt3Lim}, for all $\thetav \in \supp Q$, it follows that
	\begin{IEEEeqnarray}{rCl}
		\lim_{\lambda \rightarrow 0^+} \frac{\diff \Pgibbs{\bar{P}}{Q}}{\diff Q}(\thetav) 
		& = & \begin{cases}
			\infty & \text{if } \thetav \in \set{L}^\star_{Q, \dset{z}}\\
			0  & \text{otherwise} 
		\end{cases},
		\end{IEEEeqnarray}
which completes the proof of~\eqref{EqLemm14_s2}.

\subsection{Case $2$}
Under the assumptions of Case $2$, namely~\eqref{EqProofCase2Condition}, it holds that
 \begin{IEEEeqnarray}{rCl}
 \label{EqProofQisZeroInL}
 	Q(\set{L}^{\star}_{Q, \dset{z}}) & = & 0.
 \end{IEEEeqnarray}
This can be proved by noticing that if $Q(\set{L}^{\star}_{Q, \dset{z}}) > 0$, then 
\begin{subequations}
\label{EqProofQsetLisZero}
\begin{IEEEeqnarray}{rCl}
\IEEEeqnarraymulticol{3}{l}{%
 	\int \frac{1}{\foo{L}_{\dset{z}}(\thetav) - \delta^{\star}_{Q,\dset{z}}} \diff Q(\thetav)
 	}\nonumber\\
 	& = & \int_{\set{L}^\star_{Q, \dset{z}}}  \frac{1}{\foo{L}_{\dset{z}}(\thetav) - \delta^{\star}_{Q,\dset{z}}}\diff Q(\thetav) 
 	\nonumber \\ &  &
 	+\int_{\comp{\set{L}^\star_{Q, \dset{z}}}} \frac{1}{\foo{L}_{\dset{z}}(\thetav) - \delta^{\star}_{Q,\dset{z}}}\diff Q(\thetav)\\
 	& > & \int_{\set{L}^\star_{Q, \dset{z}}} \frac{1}{\foo{L}_{\dset{z}}(\thetav) - \delta^{\star}_{Q,\dset{z}}}\diff Q(\thetav)\\
 	& = & \frac{1}{\delta^{\star}_{Q,\dset{z}} - \delta^{\star}_{Q,\dset{z}}}Q(\set{L}^\star_{Q, \dset{z}})\\
 	& = & \infty,
\end{IEEEeqnarray}
\end{subequations}
which contradicts~\eqref{EqProofQisZeroInL}.

The proof of Case $2$ is divided into three parts.
The first part evaluates $\lim_{\lambda \rightarrow {\lambda^{\star}_{Q, \dset{z}}}^{+}} \frac{\diff \Pgibbs{\bar{P}}{Q}}{\diff Q} (\thetav)$, with $\thetav \in \{ \nuv \in \supp Q: \foo{L}_{\dset{z}}(\nuv) = \delta^\star_{Q, \dset{z}}\}$. The second part considers the case in which $\thetav \in \{ \nuv \in \supp Q: \foo{L}_{\dset{z}}(\nuv) > \delta^\star_{Q, \dset{z}}\}$. The third part considers the remaining case in~\eqref{EqProofSetsPartitionfbar36}.
\subsubsection{Part $1$}
From~\eqref{EqProofQisZeroInL} it holds that the set the set $\{ \muv \in \supp Q: \foo{L}_{\dset{z}}(\muv) = \delta^\star_{Q, \dset{z}}\}$ is a negligible set with respect to $Q$ and thus, for all $\thetav  \in \{ \muv \in \supp Q: \foo{L}_{\dset{z}}(\muv) = \delta^\star_{Q, \dset{z}}\}$, the value $\frac{\diff \Pgibbs{\bar{P}}{Q}}{\diff Q}(\thetav)$ is immaterial.

\subsubsection{Part $2$}
For all $\thetav \in\{ \nuv \in \supp Q: \foo{L}_{\dset{z}}(\nuv) > \delta^\star_{Q, \dset{z}}\}$, it holds that
\begin{subequations}
\begin{IEEEeqnarray}{rCl}
	\lim_{\lambda \rightarrow {\lambda^{\star}_{Q, \dset{z}}}^{+}} \frac{\diff \Pgibbs{\bar{P}}{Q}}{\diff Q}(\thetav) 
	& = & \lim_{\lambda \rightarrow {\lambda^{\star}_{Q, \dset{z}}}^{+}} \frac{\lambda}{\foo{L}_{\dset{z}}(\thetav) + \bar{K}_{Q, \dset{z}}(\lambda)} 
	\label{EqLimZeroLambC2_s1} \qquad \ \\
	\label{EqLimZeroLambC2_s2}
	& = &  \frac{\lambda^{\star}_{Q, \dset{z}}}{\foo{L}_{\dset{z}}(\thetav) - \delta^{\star}_{Q, \dset{z}} },
\end{IEEEeqnarray}
\end{subequations}
where~\eqref{EqLimZeroLambC2_s2} follows from observing that $\lim_{\lambda \rightarrow {\lambda^{\star}_{Q, \dset{z}}}^{+}}\foo{L}_{\dset{z}}(\thetav) + \bar{K}_{Q, \dset{z}}(\lambda) = \foo{L}_{\dset{z}}(\thetav)  - \delta^{\star}_{Q, \dset{z}} \neq 0$ (Lemma~\ref{lemm_Type2_KbarLambda}) and \cite[Theorem 4.2]{rudin1976bookPrinciples}.
\subsubsection{Part $3$}
The third part of the proof follows by noticing that the set $\{ \muv \in \supp Q: \foo{L}_{\dset{z}}(\muv) < \delta^\star_{Q, \dset{z}}\}$ is a negligible set with respect to $Q$ and thus, for all $\thetav  \in \{ \muv \in \supp Q: \foo{L}_{\dset{z}}(\muv) < \delta^\star_{Q, \dset{z}}\}$, the value $\frac{\diff \Pgibbs{\bar{P}}{Q}}{\diff Q}(\thetav)$ is immaterial. Hence, it is arbitrarily assumed that
\begin{IEEEeqnarray}{rCl}
\label{EqProofCs2Prt3Lim}
\frac{\diff \Pgibbs{\bar{P}}{Q}}{\diff Q}(\thetav) & = & 0.
\end{IEEEeqnarray}
This completes the third part of Case $2$.

From~\eqref{EqLimZeroLambC2_s2}, for all $\thetav \in \supp Q$, it follows that
	\begin{IEEEeqnarray}{rCl}
		\lim_{\lambda \rightarrow {\lambda^{\star}_{Q, \dset{z}}}^+} \frac{\diff \Pgibbs{\bar{P}}{Q}}{\diff Q}(\thetav) 
		& = &  \frac{\lambda^{\star}_{Q, \dset{z}}}{\foo{L}_{\dset{z}}(\thetav) - \delta^{\star}_{Q, \dset{z}}},
	\end{IEEEeqnarray}
which completes the proof of~\eqref{EqLemm14_s3}. 
This completes the proof. \proofIEEEend

\section{Proof of Lemma~\ref{lemm_T2AsymptLamb2ZeroPggibs}}
\label{app_proof_lemm_T2AsymptLamb2ZeroPggibs}
%

\label{proof_lemm_T2AsymptLamb2ZeroPggibs}
The proof is divided into two cases. The first case follows under the assumption that 
\begin{IEEEeqnarray}{rCl}
\label{EqProofLemm15Case1Condition}
	\int \frac{1}{\foo{L}_{\dset{z}}(\thetav)-\delta^{\star}_{Q, \dset{z}}} \diff Q(\thetav) & = & \infty;
\end{IEEEeqnarray}
and the second case follows under the assumption that
\begin{IEEEeqnarray}{rCl}
\label{EqProofLemm15Case2Condition}
	\int \frac{1}{\foo{L}_{\dset{z}}(\thetav)-\delta^{\star}_{Q, \dset{z}}} \diff Q(\thetav) & < & \infty,
\end{IEEEeqnarray}
with $\delta^\star_{Q, \dset{z}}$ in~\eqref{EqDefDeltaStar} and the function $\foo{L}_{\dset{z}}$ in~\eqref{EqLxy}.
From Lemma~\ref{lemm_Type2_kset}, it follows that in Case $1$, the set  $\set{A}_{Q, \dset{z}}$ in~\eqref{EqDefNormFunction} is $(0,\infty)$. Similarly, in Case $2$, the set $\set{A}_{Q, \dset{z}}$ is $ [\lambda^{\star}_{Q,\dset{z}},\infty)$. Hence,  Case $1$ considers the limit  ${\lambda \to 0^{+}}$, which comprehends the equality~\eqref{EqLemm15Pis1}. Case $2$ considers the limit ${\lambda \to {\lambda^{\star}_{Q,\dset{z}}}^{+}}$, which comprehends the equality~\eqref{EqLemm15Pis0}.

\subsection{Case $1$}
The first case is as follows.
Consider the following partition of the set $\set{M}$ formed by the sets	
\begin{subequations}
\label{EqProofSetsPartitionfZeroPggibs}
\begin{IEEEeqnarray}{rCl}
\label{EqProofSetsPartitionfZeroPggibs_s1}
\set{A}_0 & = &\{ \thetav \in \set{M}: \foo{L}_{\dset{z}}(\thetav) = \delta^\star_{Q, \dset{z}}\}, \\
\set{A}_1 & = &\{ \thetav \in \set{M}: \foo{L}_{\dset{z}}(\thetav) > \delta^\star_{Q, \dset{z}}\}, \text{ and}
\label{EqProofSetsPartitionfZeroPggibs_s2}\\
\set{A}_2 & = &\{ \thetav \in \set{M}: \foo{L}_{\dset{z}}(\thetav) < \delta^\star_{Q, \dset{z}}\}.
\label{EqProofSetsPartitionfZeroPggibs_s3}
\end{IEEEeqnarray}
\end{subequations}
Note that $\set{A}_0 = \set{L}^\star_{Q, \dset{z}}$, with $\set{L}^\star_{Q, \dset{z}}$ in~\eqref{EqDefSetLStarQz}.
For all $\lambda \in (0, \infty)$, it holds that
\begin{subequations}
\label{EqProofZeroPggibs}
\begin{IEEEeqnarray}{rCl}
\label{EqProofZeroPggibs_s1}
	1 
	& = & \Pgibbs{\bar{P}}{Q}(\set{A}_0) + \Pgibbs{\bar{P}}{Q}(\set{A}_1) + \Pgibbs{\bar{P}}{Q}(\set{A}_2)\quad\\
	& = & \Pgibbs{\bar{P}}{Q}(\set{A}_0) + \Pgibbs{\bar{P}}{Q}(\set{A}_1)\label{EqProofZeroPggibs_s2}\\
	& = & \Pgibbs{\bar{P}}{Q}(\set{A}_0) + \int_{\set{A}_1} \diff \Pgibbs{\bar{P}}{Q}(\thetav), \label{EqProofZeroPggibs_s3}
\end{IEEEeqnarray}
\end{subequations}
where~\eqref{EqProofZeroPggibs_s2} follows from the fact that $\Pgibbs{\bar{P}}{Q}(\set{A}_2) = 0$, which follows from the mutual absolute continuity of $\Pgibbs{\bar{P}}{Q}$ and $Q$ (Corollary~\ref{coro_mutuallyAbsCont}).
The above implies that
\begin{subequations}
\label{EqProoflimZeroPggibs}
\begin{IEEEeqnarray}{rCl}
	\IEEEeqnarraymulticol{3}{l}{
	\lim_{\lambda \rightarrow 0^+}( \Pgibbs{\bar{P}}{Q}(\set{A}_0) + \int_{\set{A}_1} \diff \Pgibbs{\bar{P}}{Q}(\thetav))
	}\nonumber\\ \qquad \quad 
	& = & \lim_{\lambda \rightarrow 0^+} \Pgibbs{\bar{P}}{Q}(\set{A}_0)
	\nonumber \\&   & 
	\label{EqProoflimZeroPggibs_s1}
	+ \lim_{\lambda \rightarrow 0^+}\int_{\set{A}_1} \frac{\diff \Pgibbs{\bar{P}}{Q}}{\diff Q}(\thetav) \diff Q(\thetav)\\
	& = & \lim_{\lambda \rightarrow 0^+} \Pgibbs{\bar{P}}{Q}(\set{A}_0) 
	\nonumber\\&   & 
	+ \int_{\set{A}_1}  \lim_{\lambda \rightarrow 0^+} \frac{\diff \Pgibbs{\bar{P}}{Q}}{\diff Q}(\thetav) \diff Q(\thetav)\label{EqProoflimZeroPggibs_s2}\\
	& = & \lim_{\lambda \rightarrow 0^+} \Pgibbs{\bar{P}}{Q}(\set{A}_0), \label{EqProoflimZeroPggibs_s3}\\
	& = & 1,
\end{IEEEeqnarray}
\end{subequations}
where,~\eqref{EqProoflimZeroPggibs_s2} follows from Lemma~\ref{lemm_ERM_RER_Type2_RNdBounds} and the dominated convergence theorem \cite[Theorem 1.6.9 page~50]{ash2000probability};
and~\eqref{EqProoflimZeroPggibs_s3} follows from Lemma~\ref{lemm_T2AsymptLamb2Zero}.
Hence, it holds that
\begin{IEEEeqnarray}{rCl}
	\lim_{\lambda \rightarrow 0^+} \Pgibbs{\bar{P}}{Q}(\set{L}^\star_{Q, \dset{z}}) & = & 1,
\end{IEEEeqnarray}
which completes the proof of~\eqref{EqLemm15Pis1}.

\subsection{Case $2$}
Under the assumptions of Case $2$, namely~\eqref{EqProofLemm15Case2Condition}, it holds that
 \begin{IEEEeqnarray}{rCl}
 \label{EqProofLemm15QisZero}
 	Q(\set{L}^{\star}_{Q, \dset{z}}) & = & 0,
 \end{IEEEeqnarray}
which can be shown using the arguments in~\eqref{EqProofQsetLisZero}.
Hence, the probability measure $\Pgibbs{\bar{P}}{Q}$ satisfies
\begin{subequations}
\label{EqProoflimZeroPggibsCs2}
\begin{IEEEeqnarray}{rCl}
	\IEEEeqnarraymulticol{3}{l}{
	\lim_{\lambda \rightarrow {\lambda^{\star}_{Q, \dset{z}}}^+} \Pgibbs{\bar{P}}{Q}(\set{A}_0) 
	}\nonumber\\ \qquad \quad 
	& = & \lim_{\lambda \rightarrow {\lambda^{\star}_{Q, \dset{z}}}^+} \int_{\set{A}_0} \frac{\diff \Pgibbs{\bar{P}}{Q}}{\diff Q}(\thetav) \diff Q(\thetav)\\
	& = & \lim_{\lambda \rightarrow {\lambda^{\star}_{Q, \dset{z}}}^+} \int_{\set{A}_0} \frac{\lambda}{\foo{L}_{\dset{z}}(\thetav)+\bar{K}_{Q, \dset{z}}(\lambda)} \diff Q(\thetav) \qquad\label{EqProoflimZeroPggibsCs2_s2}\\
	& = & \lim_{\lambda \rightarrow {\lambda^{\star}_{Q, \dset{z}}}^+} \int_{\set{A}_0} \frac{\lambda}{\delta^{\star}_{Q, \dset{z}}+\bar{K}_{Q, \dset{z}}(\lambda)} \diff Q(\thetav)\label{EqProoflimZeroPggibsCs2_s3}\\
	& = & \lim_{\lambda \rightarrow {\lambda^{\star}_{Q, \dset{z}}}^+} \frac{\lambda}{\delta^{\star}_{Q, \dset{z}}+\bar{K}_{Q, \dset{z}}(\lambda)}Q(\set{A}_0)\label{EqProoflimZeroPggibsCs2_s4}\\
	& = & \lim_{\lambda \rightarrow {\lambda^{\star}_{Q, \dset{z}}}^+} \frac{\lambda}{\delta^{\star}_{Q, \dset{z}}+\bar{K}_{Q, \dset{z}}(\lambda)}Q(\set{L}^{\star}_{Q, \dset{z}})\label{EqProoflimZeroPggibsCs2_s5}\\
	& = & \lim_{\lambda \rightarrow {\lambda^{\star}_{Q, \dset{z}}}^+} \frac{\lambda}{\delta^{\star}_{Q, \dset{z}}+\bar{K}_{Q, \dset{z}}(\lambda)}0\label{EqProoflimZeroPggibsCs2_s6}\\
	& = & 0,
\end{IEEEeqnarray}
\end{subequations}
which completes the proof of~\eqref{EqLemm15Pis0}.
This completes the proof.\proofIEEEend

\section{Proof of Lemma~\ref{lemm_T2PropertiesEmpRiskSolution}}
\label{App_lemm_T2PropertiesEmpRiskSolution}
%

%
From Lemma~\ref{lemm_Type2RNDbigcirc} and Corollary~\ref{coro_mutuallyAbsCont}, it holds that for all $\thetav \in \supp Q$,
\begin{IEEEeqnarray}{rCl}
	\frac{\diff Q}{\diff \Pgibbs{\bar{P}}{Q}}(\thetav)
	& = &
	(\frac{\diff \Pgibbs{\bar{P}}{Q}}{\diff Q}(\thetav))^{-1}\\
	& = & \frac{ \bar{K}_{Q, \dset{z}}(\lambda) + \foo{L}_{\dset{z}}(\thetav)}{\lambda}, \label{EqDevZeroElegantT2}
\end{IEEEeqnarray}
where the functions $\foo{L}_{\dset{z}}$ and $\bar{K}_{Q, \dset{z}}$ are in~\eqref{EqLxy} and~\eqref{EqType2Krescaling}, respectively. From~\eqref{EqDevZeroElegantT2}, it follows that for all $\thetav \in \supp Q$,
\begin{equation}
\label{EqProofT2RNdCharExpctER}
	0 = \lambda\frac{\diff Q}{\diff \Pgibbs{\bar{P}}{Q}}(\thetav) - \foo{L}_{\dset{z}}(\thetav) - \bar{K}_{Q, \dset{z}}(\lambda).  
\end{equation}
Integrating both sides of~\eqref{EqProofT2RNdCharExpctER} with respect to the probability measure $\Pgibbs{\bar{P}}{Q}$ yields
\begin{subequations}
\begin{IEEEeqnarray}{rCl}
	0 
	& = & \int ( \foo{L}_{\dset{z}}(\thetav)-\lambda(\frac{\diff \Pgibbs{\bar{P}}{Q}}{\diff Q}(\thetav))^{-1}
	\right.\nonumber\\&   &\left.\vphantom{\frac{\diff \Pgibbs{\bar{P}}{Q}}{\diff Q}}
	+ \bar{K}_{Q, \dset{z}}(\lambda) ) \diff \Pgibbs{\bar{P}}{Q}(\thetav)\\
	& = & \int \foo{L}_{\dset{z}}(\thetav) \diff \Pgibbs{\bar{P}}{Q}(\thetav) 
	\nonumber\\&   &
	- \lambda \int (\frac{\diff \Pgibbs{\bar{P}}{Q}}{\diff Q}(\thetav))^{-1} \diff \Pgibbs{\bar{P}}{Q}(\thetav)
	\nonumber\\ &   & 
	+ \int \bar{K}_{Q, \dset{z}}(\lambda) \diff \Pgibbs{\bar{P}}{Q}(\thetav)\\
	& = & \foo{R}_{\dset{z}}(\Pgibbs{\bar{P}}{Q}) - \lambda \int \diff Q(\thetav) + \bar{K}_{Q, \dset{z}}(\lambda)\\
	& = & \foo{R}_{\dset{z}}(\Pgibbs{\bar{P}}{Q}) - \lambda + \bar{K}_{Q, \dset{z}}(\lambda).\label{EqEqualityRe}
\end{IEEEeqnarray}
\end{subequations}
From~\eqref{EqEqualityRe}, it holds that
\begin{equation}
	\foo{R}_{\dset{z}}(\Pgibbs{\bar{P}}{Q}) =  \lambda - \bar{K}_{Q, \dset{z}}(\lambda),
\end{equation}
which completes the proof.\proofIEEEend
\section{Proof of Lemma~\ref{lemm_T2PropertiesEmpRisklambda}}
\label{app_proof_lemm_T2PropertiesEmpRisklambda}
%

%
The proof of continuity is immediate from Lemma~\ref{lemm_InfDevKtype2} and Lemma~\ref{lemm_T2PropertiesEmpRiskSolution}.
The proof of monotonicity is divided into two parts.
The first part presents the first derivative of the functional inverse $\bar{K}^{-1}_{Q, \dset{z}}$ in~\eqref{Eq_InvKbarEq} and shows that its derivative is strictly positive.
The second part shows the expected empirical risk $\foo{R}_{\dset{z}}(\Pgibbs{\bar{P}}{Q})$ is decreasing with $\lambda$.

The first part is as follows.
For all $\thetav \in \set{M}$, the partial derivative of $\frac{1}{\foo{L}_{\dset{z}}(\thetav) + \beta}$, with respect to $\beta \in (-\delta^{\star}_{Q,\dset{z}}, \infty)$, with $\delta^{\star}_{Q,\dset{z}}$ in~\eqref{EqDefDeltaStar}, is
\begin{IEEEeqnarray}{rCl}
\label{Eq_ProofType2PartRl2GeqRl1} 
	 \frac{\partial }{\partial \beta}(\frac{1}{\foo{L}_{\dset{z}}(\thetav) + \beta})
	& = & - \frac{1}{(\beta + \foo{L}_{\dset{z}}(\thetav))^{2}}.
\end{IEEEeqnarray}
From~\cite[Theorem 6.28, page~160]{klenke2020ergodic}, the following holds 
\begin{IEEEeqnarray}{rCl}
\label{Eq_ProofType2LeibnizRuleRl2GeqRl1}
		\!\!\!\!\frac{\diff}{\diff \beta}\!\int \frac{1}{\foo{L}_{\dset{z}}(\thetav) + \beta}\diff Q(\thetav)
		& = &\!\! \int \frac{\partial}{\partial \beta}\frac{1}{\foo{L}_{\dset{z}}(\thetav) + \beta}\diff Q(\thetav)\\
		\label{Eq_ProofType2LeibnizRuleRl2GeqRl1_s2}
		& = &- \int  \frac{1}{(\beta + \foo{L}_{\dset{z}}(\thetav))^{2}} \diff Q(\thetav). \qquad
\end{IEEEeqnarray}	
From Lemma~\ref{lemm_InfDevKtype2}, the derivative of the function $\bar{K}^{-1}_{Q, \dset{z}}$ in~\eqref{Eq_InvKbarEq} satisfies:
\begin{subequations}
\label{EqK_dev1Type2Rl2GeqRl1}
\begin{IEEEeqnarray}{rCl} 
	\IEEEeqnarraymulticol{3}{l}{%
		\frac{\diff}{\diff \beta} \bar{K}^{-1}_{Q, \dset{z}}(\beta)%
	}\nonumber\\
	& = & \frac{\diff }{\diff \beta}(\int \frac{1}{\beta + \foo{L}_{\dset{z}}(\thetav)}\diff Q(\thetav))^{-1}	\label{EqType2Rl2GeqRl1_s1}\\
	& = & - (\frac{1}{\int \frac{1}{\beta + \foo{L}_{\dset{z}(\thetav)}}\diff Q(\thetav)})^{2}
	\frac{\diff }{\diff \beta}\int \frac{1}{\beta + \foo{L}_{\dset{z}(\thetav)}}\diff Q(\thetav)\qquad\label{EqType2Rl2GeqRl1_s3}\\
	& = &- \frac{\int -\frac{1}{(\beta + \foo{L}_{\dset{z}}(\thetav))^{2}}\diff Q(\thetav)}{(\int \frac{1}{\beta + \foo{L}_{\dset{z}(\thetav)}}\diff Q(\thetav))^{2}}\label{EqType2Rl2GeqRl1_s5}\\
	& = & \frac{\int \frac{1}{(\beta + \foo{L}_{\dset{z}}(\thetav))^{2}}\diff Q(\thetav)}{(\int \frac{1}{\beta + \foo{L}_{\dset{z}(\thetav)}}\diff Q(\thetav))^{2}}\label{EqType2Rl2GeqRl1_s6},
\end{IEEEeqnarray}
\end{subequations}
where~\eqref{EqType2Rl2GeqRl1_s5} follows from~\eqref{Eq_ProofType2LeibnizRuleRl2GeqRl1_s2}.

Jensen's inequality \cite[Theorem 2.6.2]{cover2006book} leads to the following inequality:
\begin{IEEEeqnarray}{rCl}
\label{EqType2JensenDenNumRl2GeqRl1}
(\int \frac{1}{\beta + \foo{L}_{\dset{z}}(\thetav)}\diff Q(\thetav))^{2} 
& \leq & \int \frac{1}{(\beta + \foo{L}_{\dset{z}}(\thetav))^{2}}\diff Q(\thetav),	\quad
\end{IEEEeqnarray}
with equality if and only if the function $\foo{L}_{\dset{z}}$ in~\eqref{EqLxy} is nonseparable (Definition~\ref{DefSeparableLxy}).
Then, from~\eqref{EqType2Rl2GeqRl1_s6} and~\eqref{EqType2JensenDenNumRl2GeqRl1}, for all $\beta \in (-\delta^{\star}_{Q, \dset{z}}, \infty)$, it holds that
\begin{IEEEeqnarray}{rCl}
\label{EqType2KInvGeq1Rl2GeqRl1}
	\frac{\diff}{\diff \beta} \bar{K}^{-1}_{Q, \dset{z}}(\beta) & = & \frac{\int \frac{1}{(\beta + \foo{L}_{\dset{z}}(\thetav))^{2}}\diff Q(\thetav)}{(\int \frac{1}{\beta + \foo{L}_{\dset{z}(\thetav)}}\diff Q(\thetav))^{2}}\\
	\label{EqType2KInvGeq1Rl2GeqRl1_s2}
	&  \geq  & 1.
\end{IEEEeqnarray}
This completes the first part of the proof.

The second part is as follows.
Consider the pairs $(\lambda_1, \beta_1) \in \set{A}_{Q, \dset{z}}\times \set{C}_{Q, \dset{z}}$ and $(\lambda_2, \beta_2) \in \set{A}_{Q, \dset{z}}\times \set{C}_{Q, \dset{z}}$, such that $\lambda_2 > \lambda_1$, which implies that $ \bar{K}_{Q, \dset{z}}(\lambda_2)> \bar{K}_{Q, \dset{z}}(\lambda_1)$ (Lemma~\ref{lemm_InfDevKtype2}). Then, from Lemma~\ref{lemm_T2PropertiesEmpRiskSolution}, it follows that
\begin{subequations}
\begin{IEEEeqnarray}{rCl}
	\IEEEeqnarraymulticol{3}{l}{
	\!\!\foo{R}_{\dset{z}}(\Pgibbs[\dset{z}][\lambda_2]{\bar{P}}{Q}) - \foo{R}_{\dset{z}}(\Pgibbs[\dset{z}][\lambda_1]{\bar{P}}{Q})
	}\nonumber\\ \qquad \qquad 
	& = & \lambda_2 - \lambda_1 + \bar{K}_{Q, \dset{z}}(\lambda_1) -  \bar{K}_{Q, \dset{z}}(\lambda_2)\label{EqProofT2EERlambda}\\
	& = & \bar{K}^{-1}_{Q, \dset{z}}(\beta_2) - \bar{K}^{-1}_{Q, \dset{z}}(\beta_1) + \beta_1  - \beta_2, \label{EqProofT2EERlambdaInv}
\end{IEEEeqnarray}
\end{subequations}
where~\eqref{EqProofT2EERlambdaInv} follows from substituting~\eqref{Eq_InvKbarEq} into~\eqref{EqProofT2EERlambda}.
Note that~\eqref{EqType2KInvGeq1Rl2GeqRl1_s2} implies that
\begin{IEEEeqnarray}{rCl}
\label{EqType2BetaKInvIneqRl2GeqRl1}
	\bar{K}^{-1}_{Q, \dset{z}}(\beta_2)-\bar{K}^{-1}_{Q, \dset{z}}(\beta_1) & \geq &  \beta_2 - \beta_1,
\end{IEEEeqnarray}
with equality if and only if the function $\foo{L}_{\dset{z}}$ is nonseparable.
Thus, from~\eqref{EqProofT2EERlambdaInv} and~\eqref{EqType2BetaKInvIneqRl2GeqRl1} it follows that
\begin{IEEEeqnarray}{rCl}
\label{EqProofT2EERl1vsl2}
	\foo{R}_{\dset{z}}(\Pgibbs[\dset{z}][\lambda_2]{\bar{P}}{Q}) - \foo{R}_{\dset{z}}(\Pgibbs[\dset{z}][\lambda_1]{\bar{P}}{Q}) &\geq & 0,
\end{IEEEeqnarray}
with equality if and only if the function $\foo{L}_{\dset{z}}$ is nonseparable.
This completes the second part of the proof.\proofIEEEend

\section{Proof of Lemma~\ref{lemm_LowBoundGenGap}}
\label{app_proof_lemm_Type2InequalityJensen}

%
From Theorem~\ref{Theo_ERMType2RadNikMutualAbs} and Corollary~\ref{coro_mutuallyAbsCont}, it holds that
\begin{IEEEeqnarray}{rCl}
	 \IEEEeqnarraymulticol{3}{l}{
	 \KL{Q}{\Pgibbs[\dset{z}][\lambda]{\bar{P}}{Q}} 
	 }\nonumber\\\quad
	 & = & \int \log(\frac{\diff Q}{\diff \Pgibbs[\dset{z}][\lambda]{\bar{P}}{Q}})\diff Q(\thetav)\\
	 & \leq & \log(\int \frac{\diff Q}{\diff \Pgibbs[\dset{z}][\lambda]{\bar{P}}{Q}}(\thetav)\diff Q (\thetav))\label{EqJensenIneqType2Lemma2_s2}\\
	 & = & \log( \int \frac{ \bar{K}_{Q, \dset{z}}(\lambda) + \foo{L}_{\dset{z}}(\thetav)}{\lambda } \diff Q (\thetav))
	 \label{EqJensenIneqType2Lemma2_s3}\\
	 & = & \log( \frac{1}{\lambda}\int \bar{K}_{Q, \dset{z}}(\lambda) + \foo{L}_{\dset{z}}(\thetav)\diff Q(\thetav))\\
	 & = & \log( \frac{1}{\lambda}( \bar{K}_{Q, \dset{z}}(\lambda) + \foo{R}_{\dset{z}}(Q))), \label{EqJensenIneqType2Lemma2_s4}
\end{IEEEeqnarray}
where~\eqref{EqJensenIneqType2Lemma2_s2} follows from  Jensen's inequality \cite[Theorem 2.6.2]{cover2006book};~\eqref{EqJensenIneqType2Lemma2_s3} follows from~\eqref{EqGenpdfType2}; and~\eqref{EqJensenIneqType2Lemma2_s4} follows from~\eqref{EqRxy}.
From~\eqref{EqJensenIneqType2Lemma2_s4}, it follows that
\begin{IEEEeqnarray}{rCl}
\label{EqProofIneqRzPvsRzQ_Qineq}
	\foo{R}_{\dset{z}}(Q) 
	& \geq & \lambda\exp(\KL{Q}{\Pgibbs[\dset{z}][\lambda]{\bar{P}}{Q}})-\bar{K}_{Q, \dset{z}}(\lambda)\\
	\label{EqProofIneqRzPvsRzQ_Qineq_s2}
	&  = & \lambda\exp(\KL{Q}{\Pgibbs[\dset{z}][\lambda]{\bar{P}}{Q}})\! +\! \foo{R}_{\dset{z}}(\Pgibbs{\bar{P}}{Q})\! -\! \lambda,\qquad
\end{IEEEeqnarray}
where~\eqref{EqProofIneqRzPvsRzQ_Qineq_s2} follows from Lemma~\ref{lemm_T2PropertiesEmpRiskSolution}.
Hence, the difference between the expected empirical risk of the probability measures $\Pgibbs{\bar{P}}{Q}$ and $Q$, from~\eqref{EqProofIneqRzPvsRzQ_Qineq_s2}, satisfies that
\begin{equation}
	\foo{R}_{\dset{z}}(Q) - \foo{R}_{\dset{z}}(\Pgibbs{\bar{P}}{Q}) 
	\geq  \lambda(\exp(\KL{Q}{\Pgibbs[\dset{z}][\lambda]{\bar{P}}{Q}}) - 1),
\end{equation}
which completes the proof.\proofIEEEend
%
\section{Proof of Lemma~\ref{lemm_Type2BoundDeltaStar}}
\label{app_lemm_Type2BoundDeltaStar}
%
From Lemma~\ref{lemm_InfDevKtype2}, Lemma~\ref{lemm_Type2_kset}, and Lemma~\ref{lemm_T2PropertiesEmpRiskSolution}, it holds that
\begin{IEEEeqnarray}{rCl}
 \foo{R}_{\dset{z}}(\Pgibbs{\bar{P}}{Q}) & = & \lambda - \bar{K}_{Q, \dset{z}}(\lambda)\\
 & < & \lambda+\delta^\star_{Q, \dset{z}}.
\end{IEEEeqnarray}
%
Note also that
\begin{IEEEeqnarray}{rCl}
\label{EqProof17LowBoundEER}
 \foo{R}_{\dset{z}}(\Pgibbs{\bar{P}}{Q}) 
 & = & \int \foo{L}_{\dset{z}}(\thetav) \diff \Pgibbs{\bar{P}}{Q}(\thetav) \\
 \label{EqProof17LowBoundEER_s2}
  & \geq  & \int \delta^\star_{Q, \dset{z}} \diff \Pgibbs{\bar{P}}{Q}(\thetav) \\
  \label{EqProof17LowBoundEER_s3}
  & = & \delta^\star_{Q, \dset{z}},
\end{IEEEeqnarray}
with $\foo{L}_{\dset{z}}$ in~\eqref{EqLxy}.
%
The proof continues by determining the conditions for which~\eqref{EqProof17LowBoundEER_s3} holds with equality. Assume the empirical risk $\foo{L}_{\dset{z}}$ in~\eqref{EqLxy} is separable with respect to the probability measure $\Pgibbs{\bar{P}}{Q}$ in~\eqref{EqGenpdfType2} (see Definition~\ref{DefSeparableLxy}). Then, there exists a real value $\epsilon>0$ and two nonnegligible sets $\set{A}$ and $\set{B}$  with respect to the probability measure $\Pgibbs{\bar{P}}{Q}$ in~\eqref{EqGenpdfType2}, such that 
\begin{IEEEeqnarray}{rCl}
\set{A}
	& = & \{\thetav \in \set{M}: \foo{L}_{\dset{z}}(\thetav) < \delta^\star_{Q, \dset{z}} + \epsilon\}, \text{and}\\
	\label{EqProofSetBLem18}
	\set{B} & = & \{\thetav \in \set{M}: \foo{L}_{\dset{z}}(\thetav) \geq \delta^\star_{Q, \dset{z}} + \epsilon\}.
\end{IEEEeqnarray}
Note that the sets $\set{A}$ and $\set{B}$ form a partition of the set $\set{M}$. 
Hence, the expected empirical risk satisfies
\begin{IEEEeqnarray}{rCl}
\foo{R}_{\dset{z}}(\Pgibbs{\bar{P}}{Q}) 
	& = &	\int \foo{L}_{\dset{z}}(\thetav)\diff \Pgibbs{\bar{P}}{Q}(\thetav)\\
	& = &	\int_{\set{A}}\foo{L}_{\dset{z}}(\thetav)\diff \Pgibbs{\bar{P}}{Q}(\thetav) \nonumber\\
	&   & + \> \int_{\set{B}}\foo{L}_{\dset{z}}(\thetav)\diff \Pgibbs{\bar{P}}{Q}(\thetav)\\
	& \geq & \int_{\set{A}} \delta^{\star}_{Q, \dset{z}}\diff \Pgibbs{\bar{P}}{Q}(\thetav) \nonumber\\
	&   & + \> \int_{\set{B}}(\delta^\star_{Q, \dset{z}} + \epsilon)\diff \Pgibbs{\bar{P}}{Q}(\thetav)
	\label{EqProofIFFseparability_s2}\\
	& = & \delta^{\star}_{Q, \dset{z}}\Pgibbs{\bar{P}}{Q}(\set{A}) \nonumber\\
	&   & + \>(\delta^\star_{Q, \dset{z}} + \epsilon) \Pgibbs{\bar{P}}{Q}(\set{B}), \\
	& = & \delta^{\star}_{Q, \dset{z}}\Pgibbs{\bar{P}}{Q}(\set{A}) \nonumber \\
	&   & + (\delta^\star_{Q, \dset{z}} + \epsilon) (1-\Pgibbs{\bar{P}}{Q}(\set{A}))\label{EqProofIFFseparability_s4}\\
	& = & \delta^\star_{Q, \dset{z}} + \epsilon(1-\Pgibbs{\bar{P}}{Q}(\set{A}))\\
	& > & \delta^\star_{Q, \dset{z}}, \label{EqProofIFFseparability_s6}
\end{IEEEeqnarray}
where inequality~\eqref{EqProofIFFseparability_s2} follows from~\eqref{EqDefDeltaStar} and~\eqref{EqProofSetBLem18};~\eqref{EqProofIFFseparability_s4} follows from the fact that the sets $\set{A}$ and $\set{B}$ form a partition of the set $\set{M}$; and~\eqref{EqProofIFFseparability_s6} follows from the fact that the sets $\set{A}$ and $\set{B}$ are nonnegligible with respect to the probability measure $\Pgibbs{\bar{P}}{Q}$, which implies $\Pgibbs{\bar{P}}{Q}(\set{A})<1$. This proves the strict inequality in~\eqref{EqProof17LowBoundEER_s2}.

Consider the case in which the empirical risk $\foo{L}_{\dset{z}}$ in~\eqref{EqLxy} is not separable with respect to $\Pgibbs{\bar{P}}{Q}$ in~\eqref{EqGenpdfType2}. Then, for all $\thetav \in \supp \Pgibbs{\bar{P}}{Q}$, the empirical risk satisfies $\foo{L}_{\dset{z}}(\thetav) =\delta^\star_{Q, \dset{z}}$, which implies $\foo{R}_{\dset{z}}(\Pgibbs{\bar{P}}{Q})=\delta^\star_{Q, \dset{z}}$. 
Hence, if the function $\foo{L}_{\dset{z}}$ is nonseparable, then~\eqref{EqProof17LowBoundEER_s2} holds with equality.
Therefore,~\eqref{EqProof17LowBoundEER_s2} holds with equality, if and only if the function $\foo{L}_{\dset{z}}$ is nonseparable,   which completes the proof.\proofIEEEend

\section{Proof of Theorem~\ref{theo_DeltaEpsilonOp}}
\label{app_proof_theo_DeltaEpsilonOp}
%

	Let $\delta$ be a real in $(\delta^\star_{Q, \dset{z}}, \infty)$, with $\delta^\star_{Q, \dset{z}}$ in~\eqref{EqDefDeltaStar}.
	Let also $\gamma \in (0, \infty)$ satisfy the following equality:
	%
	\begin{IEEEeqnarray}{rCl}
	\label{EqProof_T2DeltaEpsilonOp}
		\foo{R}_{\dset{z}}(\Pgibbs[\dset{z}][\gamma]{\bar{P}}{Q}) 
		& \leq & \delta,
	\end{IEEEeqnarray}
	%
	where the existence of such a $\gamma$ is ensured by the continuity of $\foo{R}_{\dset{z}}(\Pgibbs[\dset{z}][\gamma]{\bar{P}}{Q})$ with respect to $\gamma$  (Lemma~\ref{lemm_T2PropertiesEmpRisklambda}); and Lemma~\ref{lemm_Type2BoundDeltaStar} and Lemma~\ref{lemm_T2limlamzeroRz}.
	From~\eqref{EqType2LsetLamb2zero}, it holds that
	\begin{IEEEeqnarray}{rCl}
	\label{EqProofT2DltEpslnOpSets}
		\set{L}_{\dset{z}}(\delta) & \supseteq & \set{L}^\star_{Q, \dset{z}},
	\end{IEEEeqnarray}
	and thus,
	\begin{IEEEeqnarray}{rCl}
		\Pgibbs[\dset{z}][\gamma]{\bar{P}}{Q} (\set{L}_{\dset{z}}(\delta)) 
		& \geq & \Pgibbs[\dset{z}][\gamma]{\bar{P}}{Q} (\set{L}^\star_{Q, \dset{z}}),
	\end{IEEEeqnarray}
	with $\set{L}^\star_{Q, \dset{z}}$ defined in~\eqref{EqDefSetLStarQz}.
	Let $\lambda$ be a positive real such that $\lambda \leq \gamma$, and 
	\begin{IEEEeqnarray}{rCl}
	\label{EqProofT2DltEpslnOptGamma}
		\Pgibbs[\dset{z}]{\bar{P}}{Q}(\set{L}^\star_{Q, \dset{z}}) > 1- \epsilon.	
	\end{IEEEeqnarray}
	The existence of such a positive real $\lambda$ follows from Lemma~\ref{lemm_T2AsymptLamb2ZeroPggibs}.
	From~\eqref{EqProofT2DltEpslnOpSets}, it follows that
	\label{EqProofT2DeltaEpsilonOptgamma}
	\begin{IEEEeqnarray}{rCl}
	\Pgibbs[\dset{z}][\gamma]{\bar{P}}{Q} (\set{L}_{\dset{z}}(\delta))
	& \geq & \Pgibbs[\dset{z}][\gamma]{\bar{P}}{Q} (\set{L}^\star_{Q, \dset{z}}).\label{EqProofT2DeltaEpsilonOptgamma_s6}
	\end{IEEEeqnarray}
	%
	%
	Hence, from~\eqref{EqProofT2DltEpslnOptGamma} and~\eqref{EqProofT2DeltaEpsilonOptgamma_s6}, it holds that
	\begin{IEEEeqnarray}{rCl}
	\label{EqProofT2DltEpslnOptPgibbs_s1}
		1- \epsilon & < & \Pgibbs[\dset{z}][\lambda]{\bar{P}}{Q} (\set{L}^\star_{Q, \dset{z}})\\
		& \leq & \Pgibbs[\dset{z}][\lambda]{\bar{P}}{Q} (\set{L}_{\dset{z}}(\delta)).\label{EqProofT2DltEpslnOptPgibbs_s2}
	\end{IEEEeqnarray}
	The equality in~\eqref{EqProofT2DltEpslnOptPgibbs_s2} implies that the probability measure $\Pgibbs{\bar{P}}{Q}$ is $(\delta, \epsilon)$-optimal (Definition~\ref{DefDeltaEpsilonOp}), which completes the proof.
	
	\proofIEEEend

\section{Proof of Lemma~\ref{lemm_LogEmpiricalRisk}}
\label{app_proof_lemm_LogEmpiricalRisk}

%
The proof is divided into two parts.
The first part is as follows, from Theorem~\ref{Theo_ERMType2RadNikMutualAbs}, it follows that for all $\thetav \in \set{M}$, 
\begin{align}
	\log(\frac{\diff \Pgibbs{\bar{P}}{Q}}{\diff Q}(\thetav)) 
	= & \log(\frac{\lambda}{\bar{K}_{Q, \dset{z}}(\lambda) + \foo{L}_{\dset{z}}(\thetav)})\\
	= & \log(\lambda) - \log(\bar{K}_{Q, \dset{z}}(\lambda) + \foo{L}_{\dset{z}}(\thetav))\\
	\label{Eq_Proof_logToKL}
	= & \log(\lambda) - \foo{V}_{Q, \dset{z}, \lambda}(\thetav),
\end{align}
where the function $\foo{V}_{Q, \dset{z}, \lambda}$ is defined in~\eqref{EqLogLxy}.
Thus,
\begin{align}
	\KL{\Pgibbs{\bar{P}}{Q}}{Q}
	= & \int \log(\frac{\diff \Pgibbs{\bar{P}}{Q}}{\diff Q}(\thetav)) \diff \Pgibbs{\bar{P}}{Q}(\thetav)\\
	= & \log(\lambda) - \int \foo{V}_{Q, \dset{z}, \lambda}(\thetav) \diff \Pgibbs{\bar{P}}{Q}(\thetav)\\
	= & \log(\lambda) - \bar{\foo{R}}_{Q, \dset{z}, \lambda}(\Pgibbs{\bar{P}}{Q}),
	\label{Eq_Proof_log_and_KLPtoQ_Type2}
\end{align}
where the functional $\bar{\foo{R}}_{Q, \dset{z}, \lambda}$ is defined in~\eqref{EqLogRxy}.
Hence, it follows from~\eqref{Eq_Proof_log_and_KLPtoQ_Type2} that
\begin{equation}
	\log(\lambda) = \bar{\foo{R}}_{Q, \dset{z}, \lambda}(\Pgibbs{\bar{P}}{Q}) + \KL{\Pgibbs{\bar{P}}{Q}}{Q},
\end{equation}
which completes the proof of~\eqref{EqLemm23_s1} and concludes the first part.

The second part is as follows.
From~\eqref{Eq_Proof_logToKL}, it follows that
\begin{align}
	\KL{Q}{\Pgibbs{\bar{P}}{Q}}
	= & -\int \log(\frac{\diff \Pgibbs{\bar{P}}{Q}}{\diff Q}(\thetav)) \diff Q(\thetav)\\
	= & -\log(\lambda) + \int \foo{V}_{Q, \dset{z}, \lambda}(\thetav) \diff Q(\thetav)\\
	\label{EqProofDQP}
	= & -\log(\lambda) + \bar{\foo{R}}_{Q, \dset{z}, \lambda}(Q).
\end{align}
Hence, it follows from~\eqref{EqProofDQP} that
\begin{equation}
	\log(\lambda) = \bar{\foo{R}}_{Q, \dset{z}, \lambda}(Q) - \KL{Q}{\Pgibbs{\bar{P}}{Q}},
\end{equation}
which completes the proof of~\eqref{EqLemm23_s2}.
This completes the proof.\proofIEEEend

\section{Proof of Lemma~\ref{lemm_Gen_Gap_T2}}
\label{app_proof_lemm_Gen_Gap_T2}

%
The proof uses the mutual absolute continuity between $\Pgibbs{\bar{P}}{Q}$ in~\eqref{EqGenpdfType2} and $Q$ (Corollary~\ref{coro_mutuallyAbsCont}).
Hence, a probability measure $P \in \bigcirc_{Q}(\set{M})$ is mutually absolutely continuous with $\Pgibbs{\bar{P}}{Q}$.
The proof follows by noticing that for such $P$ and for all $\thetav \in \set{M}$, it holds that
\begin{IEEEeqnarray}{rCl}
\IEEEeqnarraymulticol{3}{l}{
\log(\frac{\diff P}{\diff \Pgibbs{\bar{P}}{Q}}(\thetav)) 
}\nonumber \\
	& = & \log(\frac{\diff P}{\diff Q}(\thetav)\frac{\diff Q}{\diff \Pgibbs{\bar{P}}{Q}}(\thetav))\\
	& = & \log(\frac{\diff P}{\diff Q}(\thetav)) - \log(\frac{\diff \Pgibbs{\bar{P}}{Q}}{\diff Q}(\thetav))
	\label{Eq_ProofLogGenGap_s2}\\
	& = & \log(\frac{\diff P}{\diff Q}(\thetav)) - \log(\frac{\lambda}{\bar{K}_{Q, \dset{z}}(\lambda) + \foo{L}_{\dset{z}}(\thetav)})
	\label{Eq_ProofLogGenGap_s3}\\
	& = & \log(\frac{\diff P}{\diff Q}(\thetav)) - \log(\lambda) 
	\nonumber \\ &   & 
	+ \> \log(\bar{K}_{Q, \dset{z}}(\lambda) + \foo{L}_{\dset{z}}(\thetav))\\
	\label{Eq_ProofLogGenGapFull}
	& = & \log(\frac{\diff P}{\diff Q}(\thetav)) - \log(\lambda) + \foo{V}_{Q, \dset{z}, \lambda}(\thetav),
\end{IEEEeqnarray}
where the functions $\foo{L}_{\dset{z}}$, $\bar{K}_{Q, \dset{z}}$ and $\foo{V}_{Q, \dset{z}, \lambda}$ are defined in~\eqref{EqLxy},~\eqref{EqDefNormFunction} and in~\eqref{EqLogLxy}, respectively.
The equality~\eqref{Eq_ProofLogGenGap_s3} follows from~\eqref{EqGenpdfType2}.
Hence, the relative entropy $\KL{P}{\Pgibbs{\bar{P}}{Q}}$ satisfies,
%
\begin{IEEEeqnarray}{rCl}
	\IEEEeqnarraymulticol{3}{l}{
	\KL{P}{\Pgibbs{\bar{P}}{Q}}
	}\nonumber \\ \qquad
	& = & \int \log(\frac{\diff P}{\diff \Pgibbs{\bar{P}}{Q}}(\thetav)) \diff P(\thetav)
	\label{Eq_ProofGemGapType2_s1}\\
	& = & \int ( \log(\frac{\diff P}{\diff Q}(\thetav)) - \log(\lambda) 
	\right. \nonumber\\ &   & \left.
	+ \vphantom{\frac{\diff P}{\diff Q}} \foo{V}_{Q, \dset{z}, \lambda}(\thetav) ) \diff P(\thetav)
	\label{Eq_ProofGemGapType2_s2}\\
	& = & \int \log(\frac{\diff P}{\diff Q}(\thetav)) \diff P(\thetav) - \log(\lambda)
	\nonumber\\&   & 
	+ \int \foo{V}_{Q, \dset{z}, \lambda}(\thetav) \diff P(\thetav)
	\label{Eq_ProofGemGapType2_s3}\\
	& = & \KL{P}{Q} - \log(\lambda) + \bar{\foo{R}}_{Q, \dset{z}, \lambda}(P)
	\label{Eq_ProofGemGapType2_s4}\\
	& = & \KL{P}{Q} - \bar{\foo{R}}_{Q, \dset{z}, \lambda}(\Pgibbs{\bar{P}}{Q}) 
	\nonumber\\&   &
	- \KL{\Pgibbs{\bar{P}}{Q}}{Q} 
	+ \> \bar{\foo{R}}_{Q, \dset{z}, \lambda}(P),
	\label{Eq_ProofGemGapType2_s5}
\end{IEEEeqnarray}
where~\eqref{Eq_ProofGemGapType2_s2} follows from~\eqref{Eq_ProofLogGenGapFull};~\eqref{Eq_ProofGemGapType2_s4} follows from~\eqref{EqLogRxy};
and~\eqref{Eq_ProofGemGapType2_s5} follows from Lemma~\ref{lemm_LogEmpiricalRisk}.
Thus, from~\eqref{Eq_ProofGemGapType2_s5}, it follows that
\begin{IEEEeqnarray}{rCl}
	\IEEEeqnarraymulticol{3}{l}{
	\bar{\foo{R}}_{Q, \dset{z}, \lambda}(P) - \bar{\foo{R}}_{Q, \dset{z}, \lambda}(\Pgibbs{\bar{P}}{Q})
	}\nonumber \\
	& = & \KL{P}{\Pgibbs{\bar{P}}{Q}} - \KL{P}{Q}
	+ \KL{\Pgibbs{\bar{P}}{Q}}{Q},\quad
\end{IEEEeqnarray}
which completes the proof. \proofIEEEend

\section{Proof of Lemma~\ref{lemm_connectType1vsType2}}
\label{app_lemm_connectType1vsType2}
%

%
%

From Lemma~\ref{lemm_LogEmpiricalRisk}, for all $\alpha \in (0, \infty)$, it holds that
\begin{IEEEeqnarray}{rCl}
\label{EqProofExpectlogT1}
  \KL{\Pgibbs[\dset{z}][\alpha]{\bar{P}}{Q}}{Q} & = & - \bar{\foo{R}}_{Q, \dset{z}, \alpha}(\Pgibbs[\dset{z}][\alpha]{\bar{P}}{Q}) + \log(\alpha),
\end{IEEEeqnarray}
where the functional $\bar{\foo{R}}_{Q, \dset{z}, \alpha}$ is defined in~\eqref{EqLogRxy}.

Similarly, from \cite[Lemma~20]{perlaza2024ERMRER}, for all $\lambda \in (0, \infty)$, it holds that
\begin{IEEEeqnarray}{rCl}
\KL{\Pgibbs{P}{Q}}{Q}
\label{EqProofExpectT1}
 & \!= &  - (\!\frac{1}{\lambda}\foo{R}_{\dset{z}}(\Pgibbs{P}{Q}) \!+\! K_{Q, \dset{z}}(-\frac{1}{\lambda})\!),\qquad 
\end{IEEEeqnarray}
with the functional $\foo{R}_{\dset{z}}$ defined in~\eqref{EqRxy}.
From \cite[Theorem~3]{perlaza2024ERMRER}, the function $\foo{S}_{Q,\lambda}$ in \cite[Definition~7]{perlaza2024ERMRER} satisfies that
\begin{IEEEeqnarray}{rCl}
	\IEEEeqnarraymulticol{3}{l}{
	\foo{S}_{Q, \lambda}(\dset{z}, \Pgibbs[\dset{z}][\alpha]{\bar{P}}{Q})
	} \nonumber \\
	 	& = & \foo{R}_{\dset{z}}(\Pgibbs[\dset{z}][\alpha]{\bar{P}}{Q}) - \foo{R}_{\dset{z}}(\Pgibbs{P}{Q})
	 	\label{EqProofSens}\\
		& = & \lambda(\KL{\Pgibbs[\dset{z}][\alpha]{\bar{P}}{Q}}{\Pgibbs{P}{Q}} + \KL{\Pgibbs{P}{Q}}{Q}  
		\right.\nonumber\\&   &\left. 
		-\KL{\Pgibbs[\dset{z}][\alpha]{\bar{P}}{Q}}{Q})\\
		\label{EqProofSens_s2}
		& = & \lambda(\KL{\Pgibbs[\dset{z}][\alpha]{\bar{P}}{Q}}{\Pgibbs{P}{Q}} + \KL{\Pgibbs{P}{Q}}{Q} 
		\right.\nonumber\\&   &\left.
		+\> \bar{\foo{R}}_{Q, \dset{z}, \alpha}(\Pgibbs[\dset{z}][\alpha]{\bar{P}}{Q}) - \log(\alpha))\\
		\label{EqProofSens_s3}
		& = & \lambda(\KL{\Pgibbs[\dset{z}][\alpha]{\bar{P}}{Q}}{\Pgibbs{P}{Q}} - \frac{1}{\lambda}\foo{R}_{\dset{z}}(\Pgibbs{P}{Q}) 
		\right.\nonumber\\&   &\left.
		- K_{Q, \dset{z}}(-\frac{1}{\lambda})
		+\! \bar{\foo{R}}_{Q, \dset{z}, \alpha}(\Pgibbs[\dset{z}][\alpha]{\bar{P}}{Q}) - \log(\alpha)\!),
\end{IEEEeqnarray}
where~\eqref{EqProofSens_s2} follows from~\eqref{EqProofExpectlogT1}; and~\eqref{EqProofSens_s3} follows from~\eqref{EqProofExpectT1}.
Rearranging~\eqref{EqProofSens_s3} yields
\begin{IEEEeqnarray}{rCl}
	\IEEEeqnarraymulticol{3}{l}{
	\frac{1}{\lambda} \foo{R}_{\dset{z}}(\Pgibbs[\dset{z}][\alpha]{\bar{P}}{Q}) - \bar{\foo{R}}_{Q, \dset{z}, \alpha}(\Pgibbs[\dset{z}][\alpha]{\bar{P}}{Q}) 
	}\nonumber \\
	& = & \KL{\Pgibbs[\dset{z}][\alpha]{\bar{P}}{Q}}{\Pgibbs{P}{Q}} - \log(\alpha)
	\!- K_{Q, \dset{z}}(-\frac{1}{\lambda}).\quad\label{EqProofconnT1vsT2DiffT2}
\end{IEEEeqnarray}
Similarly, from Lemma~\ref{lemm_Gen_Gap_T2} the function $\bar{\foo{S}}_{Q, \alpha}$ in~\eqref{EqDeflogSensitivity} satisfies that 
\begin{IEEEeqnarray}{rCl}
\IEEEeqnarraymulticol{3}{l}{
\bar{\foo{S}}_{Q, \alpha}(\dset{z}, \Pgibbs{P}{Q})
}\nonumber \\ \qquad 
\label{EqProoflogSens}
	& = & \bar{\foo{R}}_{Q, \dset{z}, \alpha}(\Pgibbs{P}{Q}) - \bar{\foo{R}}_{Q, \dset{z}, \alpha}(\Pgibbs[\dset{z}][\alpha]{\bar{P}}{Q})\\
		& = & \KL{\Pgibbs{P}{Q}}{\Pgibbs[\dset{z}][\alpha]{\bar{P}}{Q}} - \KL{\Pgibbs{P}{Q}}{Q}
		\nonumber\\&   &
		+ \KL{\Pgibbs[\dset{z}][\alpha]{\bar{P}}{Q}}{Q}\\
		\label{EqProoflogSens_s3}
		& = & \KL{\Pgibbs{P}{Q}}{\Pgibbs[\dset{z}][\alpha]{\bar{P}}{Q}} - \KL{\Pgibbs{P}{Q}}{Q}
		\nonumber\\&   &
		- \bar{\foo{R}}_{Q, \dset{z}, \alpha}(\Pgibbs[\dset{z}][\alpha]{\bar{P}}{Q}) + \log(\alpha)\\
		& = & \KL{\Pgibbs{P}{Q}}{\Pgibbs[\dset{z}][\alpha]{\bar{P}}{Q}} 
		\nonumber\\&   &
		+\frac{1}{\lambda}\foo{R}_{\dset{z}}(\Pgibbs{P}{Q}) + K_{Q, \dset{z}}(-\frac{1}{\lambda})
		\nonumber\\&   &
		-\> \bar{\foo{R}}_{Q, \dset{z}, \alpha}(\Pgibbs[\dset{z}][\alpha]{\bar{P}}{Q}) + \log(\alpha),
		\label{EqProoflogSens_s4}
\end{IEEEeqnarray}
where~\eqref{EqProoflogSens_s3} follows from~\eqref{EqProofExpectlogT1}; and~\eqref{EqProoflogSens_s4} follows from~\eqref{EqProofExpectT1}.
Rearranging~\eqref{EqProoflogSens_s4} yields
\begin{IEEEeqnarray}{rCl}
\IEEEeqnarraymulticol{3}{l}{
\frac{1}{\lambda} \foo{R}_{\dset{z}}(\Pgibbs{P}{Q}) - \bar{\foo{R}}_{Q, \dset{z}, \alpha}(\Pgibbs{P}{Q}) 
}\nonumber \\ \qquad \qquad
		& = & -\KL{\Pgibbs{P}{Q}}{\Pgibbs[\dset{z}][\alpha]{\bar{P}}{Q}}
		\nonumber\\&   & 
		- \> (\log(\alpha) + K_{Q, \dset{z}}(-\frac{1}{\lambda})).
		\label{EqProofconnT1vsT2DiffT1}
\end{IEEEeqnarray}
The proof proceeds by subtracting~\eqref{EqProofconnT1vsT2DiffT1} from~\eqref{EqProofconnT1vsT2DiffT2}, resulting in
	\begin{IEEEeqnarray}{rCl}
	\IEEEeqnarraymulticol{3}{l}{ \frac{1}{\lambda}\foo{S}_{Q, \lambda}(\dset{z}, \Pgibbs[\dset{z}][\alpha]{\bar{P}}{Q}) - \bar{\foo{S}}_{Q, \alpha}(\dset{z}, \Pgibbs{P}{Q}) } \nonumber\\ 
	\qquad \qquad & = & \KL{\Pgibbs[\dset{z}][\alpha]{\bar{P}}{Q}}{\Pgibbs{P}{Q}} - \KL{\Pgibbs{P}{Q}}{\Pgibbs[\dset{z}][\alpha]{\bar{P}}{Q}} 
	\nonumber\\&   & 
	+ 2(\log(\alpha) + K_{Q, \dset{z}}(-\frac{1}{\lambda})),
	\label{EqDiffST1ST2}
\end{IEEEeqnarray}
where the functions~$\foo{S}_{Q, \lambda}$ and~$\bar{\foo{S}}_{Q, \alpha}$ are respectively defined in~\cite[Definition~7]{perlaza2024ERMRER} and~\eqref{EqDeflogSensitivity}. From \cite[Theorem~1]{perlazaISIT2023b} and Lemma~\ref{lemm_Gen_Gap_T2}, it follows that
\begin{IEEEeqnarray}{rCl}
	\IEEEeqnarraymulticol{3}{l}{ 
	 \frac{1}{\lambda}\foo{S}_{Q, \lambda}(\dset{z}, \Pgibbs[\dset{z}][\alpha]{\bar{P}}{Q}) - \bar{\foo{S}}_{Q, \alpha}(\dset{z}, \Pgibbs{P}{Q})
	}\nonumber\\ \qquad  
	& = & \KL{\Pgibbs[\dset{z}][\alpha]{\bar{P}}{Q}}{\Pgibbs{P}{Q}} - \KL{\Pgibbs{P}{Q}}{\Pgibbs[\dset{z}][\alpha]{\bar{P}}{Q}}
	\nonumber\\ &   & 
	+\> 2(\KL{\Pgibbs{P}{Q}}{Q}-\KL{\Pgibbs[\dset{z}][\alpha]{\bar{P}}{Q}}{Q}).
	\label{EqProofLogSandSidentity}
\end{IEEEeqnarray}
Substituting~\eqref{EqProofLogSandSidentity} into~\eqref{EqDiffST1ST2} yields
\begin{IEEEeqnarray}{rCl}
	\IEEEeqnarraymulticol{3}{l}{
	\KL{\Pgibbs{P}{Q}}{Q}-\KL{\Pgibbs[\dset{z}][\alpha]{\bar{P}}{Q}
	}{Q}
	}\nonumber\\ \qquad\qquad\qquad
	& = & \log(\alpha) + K_{Q, \dset{z}}(-\frac{1}{\lambda}),
\end{IEEEeqnarray}
which completes the proof.\proofIEEEend

\section{Examples}
\label{AppExamplesChangeZ}

\subsection{Example 1}
\label{AppExamplesEx1T2}

	Consider the \mbox{Type-II} ERM-RER problem in~\eqref{EqOpType2ERMRERNormal} and assume that: 
	$(a)$ $\set{M} = \set{X} = \set{Y} = [0,\infty)$; $(b)$ $\dset{z} = \left( (1,0) \right)$, which represents a single data point; and $(c)$ $Q \ll \mu$, with $\mu$ the Lebesgue measure, such that for all $\thetav \in \supp Q$,
	\begin{IEEEeqnarray}{rCl}
	\label{EqEx1dQdmu}
	 \frac{\diff Q}{\diff \mu}(\thetav) & = &  4\thetav^2\exp(-2\thetav).
	\end{IEEEeqnarray} 
	Let also the function $f:\set{M}\times\set{X}\rightarrow\set{Y}$ be 
	\begin{IEEEeqnarray}{rCl}
		f(\thetav,x)& = & x\thetav,
	\end{IEEEeqnarray}
	and the loss function $\ell$ in~\eqref{EqEll} be
	\begin{IEEEeqnarray}{rCl}
	\label{EqEx1ellDef}
		\ell(f(\thetav,x),y)& = & (x\thetav-y)^2,
	\end{IEEEeqnarray}
	which implies
	%
	\begin{IEEEeqnarray}{rCl}
	\label{EqEx1LxyDef}
		\foo{L}_{\dset{z}}(\thetav) 
		& = & (x\thetav-y)^2,
	\end{IEEEeqnarray}
	%
	with the function $\foo{L}_{\dset{z}}$ defined in~\eqref{EqLxy}. Furthermore, from assumptions $(a)$, $(b)$, and~\eqref{EqEx1LxyDef}, it follows that there exists $\thetav^{\star} \in \supp Q$ such that $\foo{L}_{\dset{z}}(\thetav^{\star})=0$, which implies that 
	\begin{equation}
	\label{EqEx1deltaStarZero}
		\delta^{\star}_{Q, \dset{z}} = 0.
	\end{equation}
	Under the current assumptions, the objective of this example is to show that $\set{C}_{Q, \dset{z}} = [\delta^{\star}_{Q, \dset{z}},\infty)$. For this purpose, from Lemma~\ref{lemm_Type2_kset}, it is sufficient to show that the condition in~\eqref{Eq_Type2KConstrainOpen} holds.
	From Theorem~\ref{Theo_ERMType2RadNikMutualAbs}, it follows that $\Pgibbs{\bar{P}}{Q}$ in~\eqref{EqGenpdfType2} satisfies for all $\thetav \in \supp Q$,  
	\begin{IEEEeqnarray}{rCl}
	\label{EqRDdPdmu}
	 \frac{\diff \Pgibbs{\bar{P}}{Q}}{\diff \mu}(\thetav) & = &  \frac{\lambda}{\foo{L}_{\dset{z}}(\thetav)+\beta}4\thetav^2\exp(-2\thetav),
	\end{IEEEeqnarray}
	with $\beta$ satisfying~\eqref{EqType2KrescConstrainAll}.
	Thus,
	%
	\begin{subequations}
	\label{EqEx1dPdmu}
	\begin{IEEEeqnarray}{rCl}
	\IEEEeqnarraymulticol{3}{l}{
	 \int \frac{1}{\foo{L}_{\dset{z}}(\thetav)-\delta^{\star}_{Q,\dset{z}}} \diff Q(\thetav)
	 }\nonumber \\
	 & = & \int \frac{1}{\foo{L}_{\dset{z}}(\thetav)-\delta^{\star}_{Q,\dset{z}}}4\thetav^2\exp(-2\thetav) \diff \mu (\thetav) \label{EqEx1dPdmu_s1}\\
	 & = &\int^{\infty}_{0}\frac{4\thetav^2\exp(-2\thetav)}{(x\thetav-y)^2-\delta^{\star}_{Q,\dset{z}}} \diff \thetav\\
	 & = & \int^{\infty}_{0}\frac{4\thetav^2\exp(-2\thetav)}{\thetav^2 - \delta^{\star}_{Q,\dset{z}}} \diff \thetav
	 \label{EqEx1dPdmu_s5}\\
	 & = &\int^{\infty}_{0}\frac{4\thetav^2\exp(-2\thetav)}{\thetav^2} \diff \thetav \label{EqEx1dPdmu_s6}\\
	 & = & \int^{\infty}_{0} 4\exp(-2\thetav) \diff \thetav\\
	 & = & 2,
	\end{IEEEeqnarray}
	\end{subequations}
	where~\eqref{EqEx1dPdmu_s1} follows from~\eqref{EqEx1dQdmu};~\eqref{EqEx1dPdmu_s5} follows from the assumption that $(x,y) = (1,0)$; and~\eqref{EqEx1dPdmu_s6} follows from the fact that $\delta^{\star}_{Q, \dset{z}} = 0$.
	Finally, the function $\bar{K}_{Q, \dset{z}}$ in~\eqref{EqDefNormFunction} satisfies $\bar{K}_{Q, \dset{z}}(\frac{1}{2})= 0$, which implies $-\delta^{\star}_{Q,\dset{z}} \in \set{C}_{Q, \dset{z}}$, thus the set $\set{A}_{Q, \dset{z}} = (0, \infty)$.

\subsection{Example 2}
\label{AppExamplesEx2T2}

Consider Example~$1$ in Appendix~\ref{AppExamplesEx1T2} with \mbox{$\dset{z} = ((1,1))$}. Note that~\eqref{EqEx1deltaStarZero} holds for this example. Under the current assumptions, the objective of this example is to show that $\set{A}_{Q, \dset{z}} = (\delta^{\star}_{Q, \dset{z}},\infty)$. For this purpose, from Lemma~\ref{lemm_Type2_kset}, it is sufficient to show that the condition in~\eqref{Eq_Type2KConstrainOpen} does not hold. That is,
\begin{subequations}
\begin{IEEEeqnarray}{rCl}
\IEEEeqnarraymulticol{3}{l}{
\int \frac{1}{\foo{L}_{\dset{z}}(\thetav)-\delta^{\star}_{Q,\dset{z}}}\diff Q(\thetav) 
}\nonumber\\
& = &  \int \frac{1}{\foo{L}_{\dset{z}}(\thetav)-\delta^{\star}_{Q,\dset{z}}}  \frac{\diff Q}{\diff \mu}(\thetav) \diff \mu(\thetav)
\label{EqEx2dPdmu_s2}\\
& = & \int \frac{4\thetav^2\exp(-2\thetav)}{\foo{L}_{\dset{z}}(\thetav)-\delta^{\star}_{Q,\dset{z}}} \diff \mu(\thetav)
\label{EqEx2dPdmu_s3}\\
& = & \int \frac{4\thetav^2\exp(-2\thetav)}{(x\thetav-y)^2-\delta^{\star}_{Q,\dset{z}}} \diff \mu(\thetav)
\label{EqEx2dPdmu_s4}\\
& = & \int \frac{4\thetav^2\exp(-2\thetav)}{(\thetav -1)^2} \diff \mu(\thetav)
\label{EqEx2dPdmu_s5}\\
& = & 4\int \frac{\thetav^2\exp(-2\thetav)}{(\thetav -1)^2} \diff \mu(\thetav)
\label{EqEx2dPdmu_s6}\\
& = & 4\int (\frac{\thetav\exp(-2\thetav)+\frac{\exp(-2\thetav)}{2}}{\thetav -1} 
\right. \nonumber \\ &  & \left.
- \frac{-\frac{\thetav\exp(-2\thetav)}{2}-\frac{\exp(-2\thetav)}{2}}{(\thetav - 1)^2}) \diff \mu(\thetav)
\label{EqEx2dPdmu_s7}\\
& = & 4(\int \frac{\thetav\exp(-2\thetav)+\frac{\exp(-2\thetav)}{2}}{\thetav -1}\diff \mu(\thetav) 
\right. \nonumber \\ &  & \left.
+ \int\frac{\frac{\thetav\exp(-2\thetav)}{2}+\frac{\exp(-2\thetav)}{2}}{(\thetav - 1)^2} \diff \mu(\thetav))
\label{EqEx2dPdmu_s8}\\
& = & 4(\frac{1}{2}\int \frac{2\thetav\exp(-2\thetav)+\exp(-2\thetav)}{\thetav -1}\diff \mu(\thetav) 
\right. \nonumber \\ &  & \left.
+ \frac{1}{2}\int\frac{\thetav\exp(-2\thetav)+\exp(-2\thetav)}{(\thetav - 1)^2} \diff \mu(\thetav)),\quad
\label{EqEx2dPdmu_s9}
\end{IEEEeqnarray}
\end{subequations}
where~\eqref{EqEx2dPdmu_s2} follows from the assumption that $Q \ll \mu$,~\eqref{EqEx2dPdmu_s3} follows from~\eqref{EqEx1dQdmu},~\eqref{EqEx2dPdmu_s5} follows from~\eqref{EqEx1deltaStarZero} and the assumption that $(x,y) = (1,1)$. Using integration by parts on the second integral in~\eqref{EqEx2dPdmu_s9}, let the functions $\phi: \set{M} \rightarrow \reals$ and $\psi:\set{M} \rightarrow \reals$ be 
\begin{subequations}
\label{EqEx2IntByPartsPhiall}
\begin{IEEEeqnarray}{rCl}
\label{EqEx2IntByPartsPhi}
 \phi(\thetav) & = & \thetav \exp(-2\thetav) + \exp(-2\thetav), \text{ and}\\
 \psi(\thetav) & = & -\frac{1}{\thetav - 1}.
\end{IEEEeqnarray}
\end{subequations}
The derivatives of $\phi$ and $\psi$ satisfy
\begin{subequations}
\label{EqEx2IntByPartsPhi2all}
\begin{IEEEeqnarray}{rCl}
\label{EqEx2IntByPartsPhi2}
 \frac{\diff \phi}{\diff \mu}(\thetav) & = & -2\thetav \exp(-2\thetav) - \exp(-2\thetav), \text{ and}\\
 \frac{\diff \psi}{\diff \mu}(\thetav) & = & \frac{1}{(\thetav - 1)^{2}},
\end{IEEEeqnarray}
\end{subequations}
respectively. Note that given a subset $[a, b] \subset \set{M}$ with $a,b \in \reals$ such that $a<b$ it holds that,
\begin{subequations}
\label{EqEx2dIntByParts}
\begin{IEEEeqnarray}{rCl}
\IEEEeqnarraymulticol{3}{l}{
\int_{[a,b]}\frac{\exp(-2\thetav)+\exp(-2\thetav)}{(\thetav - 1)^2} \diff \mu(\thetav) 
}\nonumber\\
& = & \int_{[a,b]} \phi(\thetav) \frac{\diff \psi}{\diff \mu}(\thetav) \mu (\thetav)\label{EqEx2dIntByParts_s1}\\
& = & \Big[\phi(\thetav)\psi(\thetav)\Big]^{b}_{a} - \int_{[a,b]} \frac{\diff \phi}{\diff \mu}(\thetav)\psi(\thetav) \diff \mu(\thetav)
\label{EqEx2dIntByParts_s2}\\
& = & \Big[-\frac{\thetav \exp(-2\thetav) + \exp(-2\thetav)}{\thetav - 1}\Big]^{b}_{a} 
\nonumber\\ &  &
+ \int_{[a,b]} \frac{-2\thetav \exp(-2\thetav) - \exp(-2\thetav)}{\thetav - 1} \diff \mu(\thetav),\ 
\label{EqEx2dIntByParts_s3}
\end{IEEEeqnarray}
\end{subequations}
where~\eqref{EqEx2dIntByParts_s3} follows the equalities~\eqref{EqEx2IntByPartsPhiall} and~\eqref{EqEx2IntByPartsPhi2all}. 
Substituting~\eqref{EqEx2dIntByParts_s3} into~\eqref{EqEx2dPdmu_s9} yields
\begin{subequations}
\label{EqEx2dInt}
\begin{IEEEeqnarray}{rCl}
\IEEEeqnarraymulticol{3}{l}{
\int \frac{1}{\foo{L}_{\dset{z}}(\thetav)-\delta^{\star}_{Q,\dset{z}}}\diff Q(\thetav)
}\nonumber\\
& = & 4\int_{[0,\infty)} \frac{\thetav^2\exp(-2\thetav)}{(\thetav -1)^2} \diff \mu(\thetav)
\label{EqEx2dInt_s1}\\
& = & 4\int_{[0,1]} \frac{\thetav^2\exp(-2\thetav)}{(\thetav -1)^2} \diff \mu(\thetav) 
\nonumber\\ &  &
+\> 4\int_{(1,\infty)} \frac{\thetav^2\exp(-2\thetav)}{(\thetav -1)^2} \diff \mu(\thetav)
\label{EqEx2dInt_s2}\\
& \geq & 4\int_{[0,1]} \frac{\thetav^2\exp(-2\thetav)}{(\thetav -1)^2} \diff \mu(\thetav)
\label{EqEx2dInt_s3}\\
& = & 2(\int_{[0,1]} \frac{2\thetav\exp(-2\thetav)+\exp(-2\thetav)}{\thetav -1}\diff \mu(\thetav) 
\right. \nonumber \\ &   & \left.
+ \> \int_{[0,1]}\frac{\exp(-2\thetav)+\exp(-2\thetav)}{(\thetav - 1)^2} \diff \mu(\thetav))
\label{EqEx2dInt_s4}\\
& = & 2(\int_{[0,1]} \frac{2\thetav\exp(-2\thetav)+\exp(-2\thetav)}{\thetav -1}\diff \mu(\thetav) 
\right.\nonumber\\ &   & \left.
+ \Big[-\frac{\thetav \exp(-2\thetav) + \exp(-2\thetav)}{\thetav - 1}\Big]^{1}_{0}
\right.\nonumber\\ &   & \left.
- \int_{[0,1]} \frac{2\thetav \exp(-2\thetav) + \exp(-2\thetav)}{\thetav - 1} \diff \mu(\thetav))
\label{EqEx2dInt_s5}\\
& = & 2\Big[-\frac{\thetav \exp(-2\thetav) + \exp(-2\thetav)}{\thetav - 1}\Big]^{1}_{0}
\label{EqEx2dInt_s6}\\
& = & \infty,
\label{EqEx2dInt_s10}
\end{IEEEeqnarray}
\end{subequations}
where~\eqref{EqEx2dInt_s1} follows from the assumption that $\set{M} = [0,\infty)$,~\eqref{EqEx2dInt_s3} follows from observing that for all $\thetav \in [0,\infty)$, it holds that $\frac{\thetav^2\exp(-2\thetav)}{(\thetav -1)^2} > 0$, in~\eqref{EqEx2dInt_s4} follows from~\eqref{EqEx2dPdmu_s9}, and~\eqref{EqEx2dInt_s5} follows from substituting~\eqref{EqEx2dIntByParts_s3} into~\eqref{EqEx2dInt_s4}.
From~\eqref{EqEx2dInt_s10}, it follows that the function $\bar{K}_{Q, \dset{z}}$ in~\eqref{EqDefNormFunction} is undefined at zero, which implies $\delta^{\star}_{Q,\dset{z}} \not\in \set{A}_{Q, \dset{z}}$, and this, $\set{A}_{Q, \dset{z}} = (0,\infty)$.

\subsection{Example 3}
\label{AppExamplesEx3T2}


%
Consider the \mbox{Type-II} ERM-RER problem in~\eqref{EqOpType2ERMRERNormal} and assume that: $(a)$ the set $\set{B}$ is a proper subset of $\set{M}$, and $(b)$ the probability measure $Q$ satisfies
	\begin{subequations}
	\label{EqPoneAll}
	\begin{IEEEeqnarray}{rCl}\label{EqPone}
	Q(\set{B})  & = &  \epsilon, \qquad \text{ and}\\
	Q(\set{M}\setminus\set{B}) & = & 1 - \epsilon,
	\end{IEEEeqnarray}
	\end{subequations}
	with $\epsilon>0$.
Let the empirical risk function $\foo{L}_{\dset{z}}$ in~\eqref{EqLxy} be
\begin{IEEEeqnarray}{rCl}
\label{EqLZeroOne}
	\foo{L}_{\dset{z}}(\thetav) 
	& = & \left\lbrace 
		\begin{array}{lcl}   
			0   & \text{ if } &  \vect{\theta} \in \set{B}\\
			c & \text{ if } &  \vect{\theta} \in \set{M}\setminus\set{B},
		\end{array}
		\right. 
\end{IEEEeqnarray}
with $c >0$.
Under the current assumptions, the objective of this example is to show that for all $\dset{z} \in (\set{X}\times\set{Y})^{n}$, it holds that $\set{C}_{Q, \dset{z}} = (-\delta^{\star}_{Q, \dset{z}},\infty)$ and $\set{A}_{Q, \dset{z}} = (0,\infty)$. To show this, it is necessary to characterize the function $\bar{K}_{Q, \dset{z}}$ in~\eqref{EqType2Krescaling}. Hence, from the fact that the Lagrangian multiplier $\beta$ for the optimization problem in~\eqref{EqOpType2ERMRERNormal} satisfies
	\begin{equation}
	\label{EqEx2ProofConstrain}
		\int \frac{\lambda}{\beta +\foo{L}_{\dset{z}}(\nuv)}\, \diff Q(\nuv) = 1,
	\end{equation}
	which follows from Theorem~\ref{Theo_ERMType2RadNikMutualAbs}, the empirical risk function $\foo{L}_{\dset{z}}: \set{M} \rightarrow \posreals$ in~\eqref{EqLZeroOne}, which is a simple function, and the probability measure $Q$ in~\eqref{EqPone}, it holds that
	\begin{subequations}
	\begin{IEEEeqnarray}{rCl}
	\IEEEeqnarraymulticol{3}{l}{
	\int \frac{\lambda}{\beta +\foo{L}_{\dset{z}}(\nuv)}\, \diff Q(\nuv) 
	}\nonumber \\
	& = & \lambda(\frac{1}{\beta+c_0}Q(\set{T}(\dset{z})) + \frac{1}{\beta +c_1}Q(\set{M} \setminus \set{T}(\dset{z})))\label{EqPolyExpanBeta1} \\
	& = & \lambda(\frac{1}{\beta+c_0}Q(\set{T}(\dset{z})) + \frac{1}{\beta +c_1}(1-Q(\set{T}(\dset{z}))))\label{EqPolyExpanBeta12} \\
	& = & \lambda(\frac{(\beta + c_1)Q(\set{T}(\dset{z})) + (\beta + c_0)(1- Q(\set{T}(\dset{z})))}{\beta^2+\beta(c_0+c_1) +c_0c_1})\label{EqPolyExpanBeta13} \qquad\\
	& = & \lambda(\frac{(c_1-c_0)Q(\set{T}(\dset{z})) + \beta + c_0}{\beta^2+\beta(c_0+c_1) +c_0c_1}).\label{EqPolyExpanBeta}
	\end{IEEEeqnarray}
	\end{subequations}
	From~\eqref{EqEx2ProofConstrain} and~\eqref{EqPolyExpanBeta}, it follows that 
	\begin{subequations}
	\begin{align}
	    0 = & \beta^2+\beta(c_0+c_1) +c_0c_1 - \lambda((c_1-c_0)Q(\set{T}(\dset{z}))+\beta + c_0) \\
	    = & \beta^2+\beta(c_0 + c_1 - \lambda) + c_0c_1 - \lambda c_0 -\lambda(c_1-c_0)Q(\set{T}(\dset{z})). \label{EqConstBeta22Poly}
	\end{align}
		Hence, 
	\begin{equation}
		0 = \beta^2+\beta(c_1 - \lambda) -\lambda c_1 Q(\set{T}(\dset{z})). \label{EqConstBeta2Poly}
	\end{equation}
	\end{subequations}
	\begin{subequations}
	Observe that the expression in~\eqref{EqConstBeta2Poly} is a quadratic polynomial that has two roots~$r_1$ and~$r_2$. Hence,~\eqref{EqConstBeta2Poly} in terms of~$r_1$ and~$r_2$ satisfies  
	\begin{align}
		0 =& \beta^2 - (r_1 + r_2)\beta +r_1 r_2\\
		=& (\beta - r_1)(\beta - r_2),
	\end{align}
	\end{subequations}
	where the roots~$r_1$ and~$r_2$ are given by the quadratic formula such that 
	\begin{subequations}\label{EqRootsBeta}	
	\begin{align}\label{EqRootsBeta1}
		r_1 =& -\frac{(c_1 - \lambda)}{2}-\sqrt{(\frac{c_1 - \lambda}{2})^2 + \lambda c_1 Q(\set{T}(\dset{z}))}, \intertext{and} 
		r_2 =& -\frac{(c_1 - \lambda)}{2}+\sqrt{(\frac{c_1 - \lambda}{2})^2 + \lambda c_1 Q(\set{T}(\dset{z}))}. \label{EqRootsBeta2}
	\end{align}
	\end{subequations}
	The proof continues by verifying that the roots in~\eqref{EqRootsBeta1} and~\eqref{EqRootsBeta2} are real and there is only one positive root for all $\lambda \in (0,+\infty)$ and for all $Q(\set{T}(\dset{z})) \in [0,1)$.

	Note that for all $c_1 \in (0, \infty)$ and for all $\lambda \in [0,+\infty)$, it holds that
		\begin{align}
			-\frac{c_1 - \lambda}{2} \leq & \abs{\frac{c_1 - \lambda}{2}}\\
			 =   & \sqrt{(\frac{c_1 - \lambda}{2})^2} \\
			\leq & \sqrt{(\frac{c_1 - \lambda}{2})^2 + \lambda c_1 Q(\set{T}(\dset{z}))}.\label{EqPosNegBeta}
		\end{align}

	Observe that for all $Q(\set{T}(\dset{z})) \in [0,1)$, $c_1 \in (0, \infty)$ and $\lambda \in [0, \infty)$ the expressions $(\frac{c_1 - \lambda}{2})^2$ and $\lambda c_1 Q(\set{T}(\dset{z}))$ are always positive. Thus, the square roots in~\eqref{EqRootsBeta1} and~\eqref{EqRootsBeta2} are real, which implies that $r_1$ and $r_2$ are real.
	From~\eqref{EqRootsBeta} and \eqref{EqPosNegBeta}, for all $\lambda \in [0,+\infty)$ and for all $Q(\set{T}(\dset{z})) \in [0,1)$, it holds that
	\begin{subequations}
	\begin{align}
		r_1 <& 0;
		\intertext{and following the same arguments} 
		r_2 >&0.
	\end{align}
	\end{subequations}
	Hence, the solution for the Lagrange Multiplier $\beta$ that satisfies~\eqref{EqEx2ProofConstrain} given the empirical risk function $\foo{L}_{\dset{z}}$ in~\eqref{EqLZeroOne} and the probability measure $Q$ in~\eqref{EqPone} is
	\begin{equation}\label{EqProofBetaExam1}
		\beta = -\frac{(c_1 - \lambda)}{2}+\sqrt{(\frac{c_1 - \lambda}{2})^2 + \lambda c_1 Q(\set{T}(\dset{z}))},
	\end{equation}
	which implies that the function $\bar{K}_{Q, \dset{z}}$ in~\eqref{EqType2Krescaling} under the current assumptions in~\eqref{EqPoneAll} and~\eqref{EqLZeroOne} satisfies
	\begin{equation}
	\label{EqRootsBeta2Example1}
		\bar{K}_{Q, \dset{z}}(\lambda) = -\frac{(c - \lambda)}{2}+\sqrt{(\frac{c - \lambda}{2})^2 + \lambda c Q(\set{B})}. 
	\end{equation}
From Theorem~\ref{Theo_ERMType2RadNikMutualAbs}, it follows that $\Pgibbs{\bar{P}}{Q}$ in~\eqref{EqGenpdfType2} satisfies for all $\thetav \in \supp Q$,  
	\begin{IEEEeqnarray}{rCl}
	\IEEEeqnarraymulticol{3}{l}{
	 \frac{\diff \Pgibbs{\bar{P}}{Q}}{\diff Q}(\thetav)
	 }\nonumber\\
	 & = &  \frac{\lambda}{\foo{L}_{\dset{z}}(\thetav) -\frac{(c - \lambda)}{2}+\sqrt{(\frac{c - \lambda}{2})^2 + \lambda c Q(\set{B})}}.
	 \label{EqRDdPdmuEx3}
	\end{IEEEeqnarray}
Under the current assumptions, from Lemma~\ref{lemm_Type2_kset}, it is sufficient to show that for all $c \in (0,\infty)$ and $\lambda \in (0, \infty)$ in~\eqref{EqLZeroOne}, the function $\bar{K}_{Q, \dset{z}}$ in~\eqref{EqType2Krescaling} is strictly greater than $-\delta^{\star}_{Q, \dset{z}}$.
From equality~\eqref{EqRootsBeta2Example1}, it holds that
\begin{subequations}
\begin{IEEEeqnarray}{rCl}
	\bar{K}_{Q, \dset{z}}(\lambda)
	& = & -\frac{(c - \lambda)}{2}+\sqrt{(\frac{c - \lambda}{2})^2 + \lambda c Q(\set{B})}\\
	& > & -\frac{(c - \lambda)}{2}+\sqrt{(\frac{c - \lambda}{2})^2}\\
	& = & -\frac{(c - \lambda)}{2}+\abs{\frac{c - \lambda}{2}}\\
	& \geq & 0\\
	& = & -\delta^{\star}_{Q, \dset{z}},
\end{IEEEeqnarray}
\end{subequations}
which proves that for all $c \in (0, \infty)$ and for all $\lambda \in (0,\infty)$, it holds that $\bar{K}_{Q, \dset{z}}(\lambda) > -\delta^{\star}_{Q, \dset{z}}$ which implies that $-\delta^{\star}_{Q, \dset{z}} \not\in \set{C}_{Q, \dset{z}}$ with the set $\set{C}_{Q, \dset{z}}$ defined in~\eqref{EqDefNormFunction} and thus $\set{A}_{Q, \dset{z}} = (0,\infty)$.

%
\section{Numerical Simulation}
\label{AppNumericalSimulation}

The MNIST dataset consists of $60{,}000$ images for training and $10{,}000$ images for testing. Out of the $60{,}000$ training images, $12{,}183$ are labeled as the digits six or seven, while $1{,}986$ out of the $10{,}000$ test images correspond to these digits. Each image is a $28 \times 28$ grayscale picture and is represented by the matrix $I \in [0,1]^{28\times28}$.

\subsection{Feature Extraction of the Histogram of Oriented Gradients}
\label{SubsectionHOG}

The grayscale images are processed by calculating their corresponding \emph{histogram of oriented gradients} (HOG) \cite{dalal2005histograms}. The HOG for each image is computed through the following steps:

$1.\big)$
For each pixel location $(i, j) \in \{1,2,...,28\}^{2}$ in the image, the gradients in the $w$- and $h$-directions (\emph{width, height}) are computed using finite differences given by the functions $\foo{G}_w:\{1,2,...,28\}^{2}\rightarrow \reals$ and $\foo{G}_h:\{1,2,...,28\}^{2}\rightarrow \reals$, which are defined as
\begin{IEEEeqnarray}{rCl}
\foo{G}_w(i, j) & = & \begin{cases}
	I(i+1, j) - I(i-1, j)& \!\!\!\text{if } i \in \{2,\ldots,27\} \\
	I(i+1, j) - I(i, j)  & \!\!\!\text{if } i =1 \\
	I(i, j) - I(i-1, j)  & \!\!\!\text{if } i =28
\end{cases}\!\!,\qquad \\
\noalign{\noindent and \vspace{2\jot}}
\foo{G}_h(i, j) & = & \begin{cases}
	I(i, j+1) - I(i, j-1)& \!\!\!\text{if } j \in \{2,\ldots,27\} \\
	I(i, j+1) - I(i, j)  & \!\!\!\text{if } j =1 \\
	I(i, j) - I(i, j-1)  & \!\!\!\text{if } j =28
\end{cases}\!\!\!, \qquad
\end{IEEEeqnarray}
where $I(i, j) \in [0,1]$ represents the pixel intensity at location $(i, j)$.

$2.\big)$
Given a pixel location $(i, j) \in \{1,2,...,28\}^{2}$, the magnitude and orientation of a pixel at location $(i, j)$ is given by the functions $\foo{M}:\{1,2,...,28\}^{2}\rightarrow \reals$ and  $\phi:\{1,2,...,28\}^{2}\rightarrow \reals$, such that
\begin{IEEEeqnarray}{rCl}
\label{EqDefMagnitudeM}
\foo{M}(i, j) & = & \sqrt{\foo{G}_w(i, j)^2 + \foo{G}_h(i, j)^2}, \text{ and} \\
\label{EqDefOrientationP}
\phi(i, j) & = & \arctan\left(\frac{\foo{G}_h(i, j)}{\foo{G}_w(i, j)}\right).
\end{IEEEeqnarray}

$3.\big)$ The matrix $I$ is divided into sub-matrices of size $4 \times 4$, such that the number of sub-matrices is $7$. 
These sub-matrices are referred to as \emph{cells} and are denoted, for all $(w,h) \in \{1,\cdots, 7\}^2$, by
\begin{IEEEeqnarray}{rCl}
\label{EqDefCellCm}
C_{w,h} & = & \begin{bmatrix}
 	 I(a_w, b_h)  & \cdots  & I(a_w+3, b_h)\\
 	 \vdots    & \ddots     & \vdots   \\
 	 I(a_w, b_h+3)  & \cdots  & I(a_w+3, b_h+3)
 \end{bmatrix},\quad
\end{IEEEeqnarray}
where the real values $a_w$ and $b_h$ are
\begin{IEEEeqnarray}{rCl}
\label{EqDef_aw}
a_w & = & 4(w-1)+1\\
b_h & = & 4(h-1)+1.
\label{EqDef_bh}
\end{IEEEeqnarray}
This implies that the matrix $I$ can be represented as
\begin{IEEEeqnarray}{rCl}
\label{EqDefImgAsCell}
I & = & \begin{bmatrix}
 	C_{1,1}    & \cdots  & C_{7,1}\\
 	 \vdots    & \ddots  & \vdots   \\
 	 C_{1,7} & \cdots  & C_{7,7}
 \end{bmatrix}.
\end{IEEEeqnarray}
From~\eqref{EqDefCellCm}, the set of all pairs $(i,j)$ of pixel coordinates in $I$ that lie within the cell $C_{w,h}$ is given by:
\begin{IEEEeqnarray}{rCl}
\set{A}_{w,h} & = & \{a_w,a_w+3\}\times\{b_h,b_h+3\},
\end{IEEEeqnarray}
with $a_w$ in~\eqref{EqDef_aw} and $b_h$ in~\eqref{EqDef_bh}.

$4.\big)$ For each cell $C_{w,h}$ in~\eqref{EqDefCellCm} the orientations $\phi(i, j)$ in~\eqref{EqDefOrientationP} are divided into $9$ bins. That is, the $n^{th}$ bin, with $1\leq n \leq 9$, satisfies that
\begin{IEEEeqnarray}{rCl}
\IEEEeqnarraymulticol{3}{l}{
\set{B}^{(n)}_{w,h} 
}\nonumber \\
	& = & \{\phi(i, j) \in \reals:
 	180(\frac{n-1}{9})\leq \phi(i, j) < 180(\frac{n}{9}):
	\right. \nonumber\\ &  & \left.
 	(i,j)\in \set{A}_{w,h} \vphantom{180(\frac{n-1}{9})} \}. 
 	\label{EqDefNthbin}
\end{IEEEeqnarray}
The contribution of each pixel to its corresponding bin is based on its gradient magnitude. That is, the value of the $n$-th bin from the $(w,h)$-th cell $C_{w,h}$ in~\eqref{EqDefCellCm} is given by the function $H_{w,h}(n): \{1,2, \ldots, 9\} \rightarrow \reals$, such that
\begin{IEEEeqnarray}{rCl}
\label{EqDefHistBinN}
H_{w,h}(n) & = & \sum_{(i, j) \in \set{A}_{w,h}} \foo{M}(i, j)\ind{\phi(i, j) \in \set{B}^{(n)}_{w,h}},
\end{IEEEeqnarray}
with $\foo{M}$ in~\eqref{EqDefMagnitudeM}; and $\set{B}^{(n)}_{w,h}$ in~\eqref{EqDefNthbin}. Thus, the histogram of gradient orientations of the cell $C_{w,h}$ is represented by the vector $\vect{H}_{{w,h}} \in \reals^{9}$, such that
\begin{IEEEeqnarray}{rCl}
\label{EqDefHistVect}
\vect{H}_{w,h} & = & [H_{w,h}(1),H_{w,h}(2),\cdots H_{w,h}(9)],
\end{IEEEeqnarray}
with the function $H_{w,h}$ in~\eqref{EqDefHistBinN}.

$5.\big)$
To account for illumination and contrast variations, the histogram $\vect{H}_{w,h}$ in~\eqref{EqDefHistVect} is normalized. To normalize the histograms for all cells $C_{w,h}$ in~\eqref{EqDefCellCm}, the cells are grouped into sub-matrices formed by $2 \times 2$ cells with a \emph{cell overlap} set to $1$ pixel, such that  the number of sub-matrices is:
\begin{IEEEeqnarray}{rCl}
\label{EqNumBlocks}
(7-1)\times (7-1)	& = & 36.
\end{IEEEeqnarray}
These sub-matrices of the matrix $I$ in~\eqref{EqDefImgAsCell}, with $(m,s) \in \{1, \cdots \sqrt{36}\}^2$ are referred to as \emph{blocks}, and denoted by
\begin{IEEEeqnarray}{rCl}
\label{EqDefBlockAsCell}
B_{m,s} & = & \begin{bmatrix}
 	C_{m,s}  & C_{m+1,s}\\
 	C_{m,s+1}& C_{m+1,s+1}
 \end{bmatrix},
\end{IEEEeqnarray}
with the matrix $C_{m,s}$ in~\eqref{EqDefCellCm}.
From~\eqref{EqDefImgAsCell} and~\eqref{EqDefBlockAsCell}, a block $B_{m,s}$ is a sub-matrix of size $8\times8$, \ie, $B_{m,s} \in \reals^{8\times8}$. The \emph{size} of a block is given by the ratio of the total number of pixels in a block to the number of pixels in a cell:
\begin{subequations}
\label{EqNumSizeBlock}
\begin{IEEEeqnarray}{rCl}
\frac{8\times 8}{4\times 4}
  & = & 4,
\end{IEEEeqnarray}
\end{subequations}
The normalized histogram of a cell $C_{w,h}$ in a block $B_{m,s}$ is denoted by the vector $\hat{\vect{H}}^{(m,s)}_{w,h} \in \reals^9$. This normalization of is typically done using the $\ell_2$-norm, such that
\begin{IEEEeqnarray}{rCl}
\hat{\vect{H}}^{(m,s)}_{w,h} & = & \frac{\vect{H}_{w,h}}{\sqrt{\displaystyle{\sum_{(i,j) \in \{m,m+1\}\times\{s,s+1\}}}\vect{H}_{i,j}^2 + \epsilon^2}},
\end{IEEEeqnarray}
where $\vect{H}_{w,h}$ in~\eqref{EqDefHistVect} is the unnormalized histogram, and the $\epsilon = 10^{-4}$ to avoid division by zero.

$6.\big)$
For an image with 36 blocks (see~\eqref{EqNumBlocks}), 9 orientation bins, and a size of block 4 (see~\eqref{EqNumSizeBlock}), the dimension of the HOG feature vector $\hat{\mathbf{x}}$ is:
\begin{IEEEeqnarray}{rCl}
\label{EqIntHOGVectLength}
36 \times 4 \times 9 	& = & 1296.
\end{IEEEeqnarray}
The HOG feature vector $\hat{\mathbf{x}} \in \reals^{1296}$ is formed by concatenating all the normalized histograms $\hat{\vect{H}}^{(m,s)}_{w,h}$ such that
\begin{IEEEeqnarray}{rCl}
\label{EqDefHOGPattern}
\hat{\mathbf{x}} 
	& = & [ \hat{\vect{H}}^{(1,1)}_{1,1},\hat{\vect{H}}^{(1,1)}_{1,2},\hat{\vect{H}}^{(1,1)}_{2,1},\hat{\vect{H}}^{(1,1)}_{2,2},
	\right.\nonumber \\ &  &\left. \>
	\hat{\vect{H}}^{(2,1)}_{2,1},\hat{\vect{H}}^{(2,1)}_{2,2},\hat{\vect{H}}^{(2,1)}_{3,1},\hat{\vect{H}}^{(2,1)}_{3,2}, \cdots, 
	\right.\nonumber \\ &  &\left. \>
	\hat{\vect{H}}^{(6,6)}_{6,6},\hat{\vect{H}}^{(6,6)}_{6,7},\hat{\vect{H}}^{(6,6)}_{7,6},\hat{\vect{H}}^{(6,6)}_{7,7}]^\top.
\end{IEEEeqnarray}

\subsection{Principal Component Analysis}
The final step in the data processing is to reduce the dimensionality of the pattern $\hat{\mathbf{x}}$ in~\eqref{EqDefHOGPattern} from $\reals^{1296}$ to $\reals^{2}$, while ensuring that the important structure of the pattern is preserved. In this simulation, \textit{principal component analysis (PCA)} is used to project the high-dimensional data onto a lower-dimensional subspace. 
From $60{,}000$ images for training in the MNIST, the HOG of two handwritten numbers (in this simulation $6$ and $7$) are computed, as mentioned in Appendix~\ref{SubsectionHOG}. The resulting $12{,}183$ HOG vectors $\hat{\mathbf{x}}  \in \reals^{1296}$ are reduced to $\reals^{2}$ using PCA as follows:

$1.\big)$ To reduce the dimensionality, the first step in PCA is to compute the \textit{covariance matrix} of the data. This matrix captures the relationships between the different features (or dimensions) of the data. The covariance matrix is calculated as follows:

\begin{IEEEeqnarray}{rCl}
\label{EqCovarianceMatrix}
\mathbf{C} & = & \frac{1}{n-1} \sum_{i=1}^n (\hat{\mathbf{x}} _i - \muv)(\hat{\mathbf{x}} _i - \muv)^\top,
\end{IEEEeqnarray}

where $n = 12{,}183$, $\mathbf{C} \in \reals^{1296 \times 1296}$, and $\muv$ is the mean of all the training patterns given by
\begin{IEEEeqnarray}{rCl}
	\muv & = & \frac{1}{n} \sum_{i=1}^n \hat{\mathbf{x}} _i.
\end{IEEEeqnarray}

$2.\big)$ The next step in PCA is to perform an \textit{eigenvalue decomposition} of the covariance matrix $\mathbf{C}$ in~\eqref{EqCovarianceMatrix}. The decomposition can be written as:
\begin{IEEEeqnarray}{rCl}
\mathbf{C} & = & \mathbf{V} \mathbf{\Lambda} \mathbf{V}^\top,
\end{IEEEeqnarray}
where $\mathbf{V} \in \reals^{1296 \times 1296}$ is a matrix whose columns are the eigenvectors of $\mathbf{C}$ and $\mathbf{\Lambda} \in \reals^{1296 \times 1296}$ is a diagonal matrix containing the corresponding eigenvalues.
%

$3.\big)$ Following the computation of the eigenvectors, the dimensionality is reduced from $\reals^{1296}$ to $\reals^2$ by selecting the two eigenvectors associated with the largest eigenvalues. Denote these top two eigenvectors as $\mathbf{w}_1$ and $\mathbf{w}_2$. These eigenvectors constitute the columns of the projection matrix $\mathbf{W} \in \reals^{1296\times 2}$, defined as
\begin{IEEEeqnarray}{rCl}
\label{EqDefWPACsim}
\mathbf{W} & = & \left[ \mathbf{w}_1 \, \mathbf{w}_2 \right].
\end{IEEEeqnarray}
%

$4.\big)$ Once the projection matrix $\mathbf{W}$ is computed, each high-dimensional pattern $\hat{\mathbf{x}}  \in \reals^{1296}$ can be projected onto the new $\reals^2$ subspace. The projection is performed as follows:
\begin{IEEEeqnarray}{rCl}
\label{EqDefRlToR2}
\mathbf{x} & = & \mathbf{W}^\top \hat{\mathbf{x}} ,
\end{IEEEeqnarray}
with $\hat{\mathbf{x}}$ in~\eqref{EqDefHOGPattern}, $\vect{W}$ in~\eqref{EqDefWPACsim} and $\mathbf{x} \in \reals^2$ is the 2-dimensional coordinates of the original pattern $\hat{\mathbf{x}}$ in the reduced-dimensional space.

\subsection{Simulation Dataset}

In this simulation, a datapoint is a tuple $(\hat{\mathbf{x}},y) \in \reals^{1296}\times\{6,7\}$, with $\hat{\mathbf{x}}$ in~\eqref{EqDefHOGPattern} and $y$ being the label assigned by MNIST to the image $I$ in~\eqref{EqDefImgAsCell}. The label $y$ corresponds to the digit in the image $I$. Such an image produces the vector $\hat{\mathbf{x}}$, when its HOG features are computed.

\IEEEtriggeratref{49}
\bibliographystyle{IEEEtranlink}
\bibliography{iEEEtranBibStyle.bib}

\end{document}